\documentclass[10pt,journal,compsoc]{IEEEtran}
\usepackage{microtype}
\usepackage{graphicx}
\usepackage{subfigure}
\usepackage{booktabs}

\usepackage{amsmath}
\usepackage{amssymb}
\usepackage{tabularx}
\usepackage{color}
\usepackage{bm}
\usepackage{algorithm}
\usepackage{algorithmic}
\usepackage{multirow}
\usepackage{subfigure}
\usepackage{amsthm}
\usepackage{textcomp}
\newtheorem{thm}{Theorem}
\newtheorem{cor}{Corollary}
\newtheorem{prop}{Proposition}

\newtheorem{defi}{Definition}
\newtheorem{remark}{Remark}

\ifCLASSOPTIONcompsoc
  \usepackage[nocompress]{cite}
\else
  \usepackage{cite}
\fi
\ifCLASSINFOpdf
\else
\fi
\begin{document}
%
\title{On the Convergence of Learning-based Iterative  Methods for Nonconvex Inverse Problems}
%
%
%
%

\author{Risheng~Liu,~\IEEEmembership{Member,~IEEE,}
        Shichao~Cheng,
        Yi~He,
        Xin~Fan,~\IEEEmembership{Member,~IEEE,}
        Zhouchen Lin,~\IEEEmembership{Fellow,~IEEE,}
        and~Zhongxuan~Luo
\IEEEcompsocitemizethanks{\IEEEcompsocthanksitem R. Liu, Y. He, and X. Fan are with the DUT-RU International School of Information Science \& Engineering, Dalian University of Technology, and also with the Key Laboratory for Ubiquitous Network and Service Software of Liaoning Province, Dalian 116024, China.
E-mail: \{rsliu,xin.fan\}@dlut.edu.cn, heyiking@outlook.com.
\IEEEcompsocthanksitem S. Cheng is with the School of Mathematical Sciences, Dalian University of Technology, and also with the Key Laboratory for Ubiquitous Network and Service Software of Liaoning Province, Dalian 116024, China.
E-mail: shichao.cheng@outlook.com.
\IEEEcompsocthanksitem Z. Lin is with the Key Laboratory of Machine Perception (Ministry of Education), School of Electronics Engineering and Computer Science, Peking University, Beijing 100871, China, and also with the Cooperative Medianet Innovation Center, Shanghai Jiao Tong University, Shanghai 200240, China.
E-mail: zlin@pku.edu.cn.
\IEEEcompsocthanksitem Z. Luo is with the DUT-RU International School of Information Science \& Engineering, Dalian University of Technology, the Key Laboratory for Ubiquitous Networ and Service Software of Liaoning Province, and the School of Mathematical Sciences, Dalian University of Technology, Dalian 116024, China, and also with the Institute of Artificial Intelligence, Guilin University of Electronic Technology, Guilin 541004, China.
E-mail: zxluo@dlut.edu.cn}
\thanks{Manuscript received April 19, 2018; revised August 26, 2015.}}

%
%

\markboth{Journal of \LaTeX\ Class Files,~Vol.~14, No.~8, August~2015}%
{Shell \MakeLowercase{\textit{et al.}}: Bare Demo of IEEEtran.cls for Computer Society Journals}
%



\IEEEtitleabstractindextext{%
\begin{abstract}
Numerous tasks at the core of statistics, learning and vision areas are specific cases of ill-posed inverse problems. Recently, learning-based (e.g., deep) iterative  methods have been empirically shown to be useful for these problems. Nevertheless, integrating learnable structures into iterations  is still a laborious process, which can only be guided by intuitions or empirical insights. Moreover, there is a lack of rigorous analysis about the convergence behaviors of these reimplemented iterations, and thus the significance of such methods is a little bit vague. This paper moves beyond these limits and proposes Flexible Iterative Modularization Algorithm (FIMA), a generic and provable paradigm for nonconvex inverse problems. Our theoretical analysis reveals that FIMA allows us to generate globally convergent trajectories for learning-based iterative methods. Meanwhile, the devised scheduling policies on flexible modules should also be beneficial for classical numerical methods in the nonconvex scenario. Extensive experiments on real applications verify the superiority of FIMA.
\end{abstract}

\begin{IEEEkeywords}
Nonconvex optimization, Learning-based iteration, Global convergence , Computer vision.
\end{IEEEkeywords}}

\maketitle

\IEEEdisplaynontitleabstractindextext

%
\IEEEpeerreviewmaketitle

\IEEEraisesectionheading{\section{Introduction}\label{sec:introduction}}
\IEEEPARstart{I}{n} applications throughout statistics, machine learning and computer vision, one is often faced with the challenge of solving ill-posed inverse problems. In general, the basic inverse problem leads to a discrete linear system of the form $\mathcal{T}(\mathbf{x})=\mathbf{y} + \mathbf{n}$, where
$\mathbf{x}\in\mathbb{R}^{D}$ is the latent variable to be estimated, $\mathcal{T}$ denotes some given linear operations on $\mathbf{x}$, and $\mathbf{y},\mathbf{n}\in\mathbb{R}^D$ are the observation and an unknown error term, respectively. 
Typically, these inverse problems can be  addressed by solving the composite minimization model:
\begin{equation}
	\min\limits_{\mathbf{x}}\Psi(\mathbf{x}):=f(\mathbf{x};\mathcal{T},\mathbf{y})+g(\mathbf{x}),\label{eq:psi}
\end{equation}
where $f$ is the fidelity that captures the loss of data fitting,
and $g$ refers to the prior that promotes desired distribution on the solution. Recent studies illustrate that many problems (e.g., image deconvolution, matrix factorization and dictionary learning) naturally require to be solved in the nonconvex scenario. This trend motivates us to investigate Nonconvex Inverse Problems (NIPs) in the form of Eq.~\eqref{eq:psi} and with the practical configuration that $f$ is continuously differentiable, $g$ is nonsmooth, and both $f$ and $g$ are possibly nonconvex.

Over the past decades, a broad class of first-order methods have been developed to solve special instances of Eq.~\eqref{eq:psi}. For example, by integrating Nesterov's acceleration~\cite{nesterov1983method} into the fundamental Proximal Gradient (PG) scheme, Accelerated Proximal Gradient (APG, a.k.a. FISTA~\cite{beck2009a}) method is initially
proposed to solve convex models in the form of Eq.~\eqref{eq:psi} for different applications, such as image restoration \cite{beck2009a}, image deblurring \cite{xu2011image}, and sparse/low-rank learning \cite{candes2011robust}, etc. While these APGs generate a sequence of objectives that may oscillate~\cite{beck2009a}, \cite{beck2009fast} developed a variant of APG that guarantees the monotonicity of the sequence.
For nonconvex energies in Eq.~\eqref{eq:psi}, Li and Lin \cite{li2015accelerated} investigated a monotone APG (mAPG) and proved the convergence under the Kurdyka-{\L}ojasiewicz (K{\L}) constraint \cite{attouch2009convergence}. The work in \cite{yao2016more}  developed another variation of APG (APGnc)
for nonconvex problems, but their original analysis only characterized the fixed-point convergence. Recently,
Li et al. \cite{li2017convergence}  also proved the subsequence convergence of APGnc and estimated its convergence rates by further exploiting K{\L} property.

Unfortunately, even with some theoretically proved convergence properties, these classical numerical solvers may still fail in real-world scenarios. This is mainly
because that the abstractly designed and fixed updating schemes do not exploit the particular structure of the problem at hand nor the input data distribution \cite{moreau2016understanding}.  


In recent years, various learning-based strategies \cite{gregor2010learning,Liu2016Learning,chen2016trainable,Yang2017ADMM,wang2016proximal} have been proposed to address practical inverse problems in the form of Eq.~\eqref{eq:psi}. These methods first introduced hyperparameters into the classical numerical solvers and then performed discriminative learning on collected training data to obtain some data-specific (but possibly inconsistent) iteration schemes. Inspired by the success of deep learning in different application fields, some preliminary studies considered the handcrafted network architectures as the implicit priors (a.k.a. deep priors) for inverse problems. Following this perspective, various deep priors are designed and nested into numerical iterations \cite{diamond2017unrolled,Zhang2017Learning,ulyanov2017deep}. Alternately, the works in \cite{li2016learning} and \cite{andrychowicz2016learning} addressed the iteration learning issues from the perspectives of deep reinforcement and recurrent learning, respectively.

Nevertheless, existing hyperparameters learning approaches can only build iterations based on the specific energy forms (e.g., $\ell_1$-penalty and MRFs), so that they are inapplicable for more generic inverse problems. Meanwhile, due to severe inconstancy of parameters during iterations, rigorous analysis on the resulted trajectories is also missing. 
Deep iterative methods have been executed in many learning and vision problems in practice. However, due to the complex network structure, little or even to no results have been proposed for the convergence behaviors of these methods.
In summary, the lack of strict theoretical investigations is one of the most fundamental limits in prevalent learning-based iterative methods, especially in the challenging nonconvex scenario.



To break the limits of prevalent approaches, this paper explores Flexible Iterative Modularization Algorithm (FIMA), a \emph{generic} and \emph{convergent} algorithmic framework that combines together the learnable architecture (e.g., mainstream deep networks) with principled knowledges (formulated by mathematical models), to tackle challenging NIPs in  Eq.~\eqref{eq:psi}.
Specifically, derived from the fundamental forward-backward updating mechanism, FIMA replaces  specific calculations corresponding to the fidelity and priors in Eq.~\eqref{eq:psi} with two user-specified (learnable) computational modules. 
A series of theoretical investigations are established for FIMA. For example, we first prove the subsequence convergence of FIMA with explicit momentum policy (called eFIMA), which is as good as those mathematically designed nonconvex proximal methods with Nesterov's acceleration (e.g., various APGs in \cite{li2015accelerated,yao2016more,li2017convergence}). By introducing a carefully devised error-control policy (i.e., implicit momentum policy, called iFIMA), we further enhance the results and obtain a globally convergent Cauchy sequence for Eq.~\eqref{eq:psi}. We prove that this guarantee can also be preserved for FIMA with multiple blocks of unknown variables (called mFIMA). As a nontrivial byproduct, we finally show how to specify  modules in FIMA for challenging inverse problems in low-level vision area (e.g., non-blind and blind image deconvolution). 
Our primary contributions are summarized as follows:
\begin{enumerate}
	\item FIMA provides a generic framework that unifies almost all existing learning-based iterative methods, as well as a series of scheduling policies that make it possible to develop theoretically convergent learning-based iterations for challenging nonconvex inverse problems in the form of Eq.~\eqref{eq:psi}. 
	\item Even with highly flexible (learnable) iterations, the convergence guarantees obtained by FIMA is still as good as (eFIMA) or better (iFIMA) than prevalent mathematically designed nonconvex APGs. So it is worth noting that our devised scheduling policies together with the flexible algorithmic structures should also be beneficial for classical nonconvex algorithms.
	\item FIMA also provides us a practical and effective ensemble of domain knowledge and sophisticated learned data distributions for real applications. Thus we can bring the expressive power of knowledge-based and data-driven methodologies to yield state-of-the-art performance on challenging low-level vision tasks.
\end{enumerate}

\section{Related Work}


\subsection{Classical First-order Numerical Solvers}\label{sec:pg}

We first briefly review a group of classical first-order algorithms, which have been widely used to solve inverse problems. The gradient descent (GD) scheme on a differentiable function $f$ can be reformulated as minimizing the following quadratic approximation of $f$ at given point $\mathbf{v}$ with step size $\gamma>0$, i.e.,
$Q_{\gamma f}(\mathbf{x};\mathbf{v}):=f(\mathbf{v})+\langle \nabla f(\mathbf{v}),\mathbf{x}-\mathbf{v}\rangle + \frac{1}{2\gamma}\|\mathbf{x}-\mathbf{v}\|^2$.
As for the nonsmooth function $g$, its proximal mapping (PM) with parameter $\gamma>0$ can be defined as
$\mathtt{prox}_{\gamma g}(\mathbf{v})\in\arg\min\limits_{\mathbf{x}}g(\mathbf{x}) + \frac{1}{2\gamma}\|\mathbf{x}-\mathbf{v}\|^2$.
So it is natural to consider PG as cascade of GD (on $f$) and PM (on $g$), or equivalently optimizing the quadratic approximation of Eq.~\eqref{eq:psi}, i.e.,
$\mathbf{x}^{k+1}\in\arg\min_{\mathbf{x}}g(\mathbf{x})+Q_{\gamma^kf}(\mathbf{x};\mathbf{v}^k)$,
where $\mathbf{v}^k$ is some calculated variable at $k$-th iteration. Thus
most prevalent proximal schemes can be summarized as
$$
\begin{array}{lr}
\mathbf{v}^{k}=
\left\{\begin{array}{lr}
\mathbf{x}^k, & \mbox{(A-1)}\\
\mathbf{x}^k + \beta^k(\mathbf{x}^k-\mathbf{x}^{k-1}), & \mbox{(A-2)}
\end{array}\right.\\
\mathbf{x}^{k+1}=
\left\{\begin{array}{lr}
\mathtt{prox}_{{\gamma^k}g}\left(\mathbf{v}^{k}-\gamma^k\nabla f(\mathbf{v}^{k})\right), &\mbox{(B-1)}\\
\mathtt{prox}_{{\gamma^k}g}^{\varepsilon^k}\left(\mathbf{v}^{k}-\gamma^k\nabla  f\left(\mathbf{v}^{k}+\mathbf{e}^k\right)\right), &\mbox{(B-2)}
\end{array}\right.
\end{array}
$$
where $\varepsilon^k$ and $\mathbf{e}^k$ in (B-2) denote the errors in PM and GD calculations, respectively~\cite{gu2016inexact}.
Within this general scheme, we first obtain original PG by setting $\mathbf{v}^k=\mathbf{x}^k$ (i.e., (A-1)) and computing PM in (B-1)~\cite{beck2009a}. Using Nesterov's acceleration~\cite{nesterov1983method} (i.e., (A-2) with $\beta^k>0$), we have the well-known APG method~\cite{beck2009a,li2015accelerated,li2017convergence}.
Moreover, by introducing $\varepsilon^k$ and $\mathbf{e}^k$ to respectively capture the inexactness of PM and GD (i.e., (B-2)), we actually consider inexact PG and APG for both convex \cite{schmidt2011convergence} and nonconvex \cite{gu2016inexact} problems. Notice that in the nonconvex scenario, most classical APGs can only guarantee the subsequence convergence to the critical points~\cite{li2015accelerated,li2017convergence}.

\subsection{Learning-based Iterative Methods}\label{sec:learning-based}

In \cite{gregor2010learning}, a trained version of FISTA (called LISTA) is introduced to approximate the solution of LASSO. \cite{bronstein2012learning,moreau2016understanding} extended LISTA for more generic sparse coding tasks and provided an adaptive acceleration. Unfortunately, LISTA is built on convex $\ell_1$ regularization, thus may not be applicable for other complex nonconvex inverse problems (e.g., $\ell_0$ prior). 
By introducing hyperparameters in MRF and solving the resulted variational model with different iteration schemes, various learning-based iterative methods are proposed for inverse problems in image domain (e.g., denoising, super-resolution, and MRI imaging). For example, \cite{schmidt2014shrinkage,chen2016trainable,Yang2017ADMM,chan2017plug,wang2016proximal} have considered half-quadratic
splitting, gradient descent, Alternating Direction Method of Multiplier (ADMM) and primal-dual method, respectively. But their parameterizations are completely based on MRF priors. Even worse, the original convergence properties are lost in these resulted iterations.

To better model complex image degradations, \cite{diamond2017unrolled,Zhang2017Learning,ulyanov2017deep} considered Convolutional Neural Networks (CNNs) as implicit priors  for image restoration. Since these methods discard the regularization term in Eq.~\eqref{eq:psi}, we may not enforce principled constraints on their solutions. It is also unclear when and where these iterative trajectories should stop. Another group of very recent works \cite{li2016learning,andrychowicz2016learning} directly formulated the descent directions from reinforcement learning perspective or using recurrent networks. However, due to the high computational budgets, they can only be applied to relative simple tasks (e.g., linear regression). Besides, due to the complex topological network structure, it is extremely hard  to provide strict theoretical analysis for these methods.

\section{The Proposed Algorithms}\label{sec:fima}

This section develops Flexible Iterative Modularization Algorithm (FIMA) for nonconvex inverse problems in Eq.~\eqref{eq:psi}. The convergence behaviors are also investigated accordingly. 
Hereafter, some fairly loose assumptions are enforced on Eq.~\eqref{eq:psi}: $f$ is proper and Lipschitz smooth (with modulus $L$) on a bounded set, $g$ is proper, lower semi-continuous and proximable\footnote{The function $g$ is proximable if $\min_{\mathbf{x}}g(\mathbf{x}) + \frac{\gamma}{2}\|\mathbf{x}-\mathbf{y}\|^2$ can be easily solved by the given $\mathbf{y}$ and $\gamma>0$.} and $\Psi$ is coercive. Notice that the proofs and definitions are deferred until Supplementary Materials.


\subsection{Abstract Iterative Modularization}\label{sec:modular}

As summarized in Sec.~\ref{sec:pg}, a large amount of first-order methods can be summarized as forward-backward-type iterations. This motivates us to consider the following even more abstract updating principle:
\begin{equation}
	\mathbf{x}^{k+1}=\mathcal{A}_g\circ\mathcal{A}_f(\mathbf{x}^{k}),\label{eq:af_ag}
\end{equation}
where $\mathcal{A}_f$ and $\mathcal{A}_g$ respectively stand for the user-specified modules for $f$ and $g$, and $\circ$ denotes operation composition. Building upon this formulation, it is easy to see that designing a learning-based iterative method reduces to the problem of iteratively specifying and learning $\mathcal{A}_f$ and $\mathcal{A}_g$. 

It is straightforward that most prevalent approaches \cite{diamond2017unrolled,Zhang2017Learning,ulyanov2017deep,schmidt2014shrinkage,chen2016trainable,Yang2017ADMM,wang2016proximal} naturally fall into this general formulation. Nevertheless, currently it is still impossible to provide any strict theoretical results for practical trajectories of Eq.~\eqref{eq:af_ag}. This is mainly due to the lack of efficient mechanisms to control the propagations generated by these handcrafted operations. Fortunately, in the following, we will introduce different scheduling policies to automatically guide the iterations in Eq.~\eqref{eq:af_ag}, resulting in a series of \emph{theoretically convergent learning-based iterative methods}. 




\subsection{Explicit Momentum: A Straightforward Strategy}\label{sec:momentum}

The momentum of objective values is one of the most important properties for numerical iterations. This property is also  necessary for analyzing the convergence of some classical algorithms. Inspired by these points, we present an explicit momentum FIMA (eFIMA) (i.e., Alg.~\ref{alg:eFIMA}), in which we explicitly compare $\Psi(\mathbf{u}^k)$ and $\Psi(\mathbf{x}^k)$ and choose the variable with less objective value as our monitor (denoted as $\mathbf{v}^k$). 
Finally, a proximal refinement is performed to adjust the learning-based updating at each stage.

\begin{algorithm}
	\caption{Explicit Momentum FIMA (eFIMA)}\label{alg:eFIMA}
	\begin{algorithmic}[1]
		\REQUIRE $\mathbf{x}^0$, $\mathcal{A}=\{\mathcal{A}_g, \mathcal{A}_f\}$, and $\{0< \gamma^{k} < 1/L \}$.
		\WHILE{not converged}
		\STATE $\mathbf{u}^{k}=\mathcal{A}_{g}\circ\mathcal{A}_f(\mathbf{x}^{k})$.\label{step:u}
		\IF{$\Psi(\mathbf{u}^k)\leq\Psi(\mathbf{x}^k)$}\label{step:ms}
		\STATE $\mathbf{v}^{k}=\mathbf{u}^{k}$.
		\ELSE
		\STATE $\mathbf{v}^{k}=\mathbf{x}^k$.
		\ENDIF \label{step:me}
		\STATE $\mathbf{x}^{k+1}=\mathtt{prox}_{{\gamma^k}g}\left(\mathbf{v}^{k}-\gamma^k\nabla f(\mathbf{v}^{k})\right)$.\label{step:x}
		\ENDWHILE	
	\end{algorithmic}
\end{algorithm}

The following theorem first verifies the sufficient descent of $\{\Psi(\mathbf{x}^k)\}_{k\in\mathbb{N}}$ and then proves the subsequence convergence of eFIMA. It is nice to observe that these results are not based on any specific choices of $\mathcal{A}_f$ and $\mathcal{A}_g$.

\begin{thm}\label{thm:eFIMA}Let $\{\mathbf{x}^k\}_{k\in\mathbb{N}}$ be the sequence generated by eFIMA. Then at the $k$-th iteration, there exists a sequence $\{\alpha^k | \alpha^k> 0\}_{k\in\mathbb{N}}$, such that
	\begin{equation}
	\Psi\left(\mathbf{x}^{k+1}\right)\leq\Psi\left(\mathbf{v}^k\right)-\alpha^k\|\mathbf{x}^{k+1}-\mathbf{v}^k\|^2,
	\label{eq:func_increase}
	\end{equation}
	where $\mathbf{v}^k$ is the monitor in Alg.~\ref{alg:eFIMA}.
	Furthermore, $\{\mathbf{x}^k\}_{k\in\mathbb{N}}$ is bounded and any of its accumulation points are the critical points of $\Psi(\mathbf{x})$ in Eq.~\eqref{eq:psi}.
\end{thm}


Based on Theorem~\ref{thm:eFIMA} and considering $\Psi$ as a semi-algebraic function\footnote{Indeed, a variety of functions (e.g., the indicator function of polyhedral set, $\ell_0$ and rational $\ell_p$ penalties) satisfy the semi-algebraic property~\cite{bolte2014proximal}. }, the convergence rate of eFIMA can be straightforwardly estimated as follows.
\begin{cor}\label{cor:eFIMA}
	Let $\phi(s)=\frac{t}{\theta}s^{\theta}$ be a desingularizing function with a constant $t>0$ and a parameter $\theta\in(0,1]$  \cite{chouzenoux2016block}. Then $\{\mathbf{x}^k\}_{k\in\mathbb{N}}$ generated by eFIMA converges after finite iterations if $\theta=1$. The linear and sub-linear rates can be obtained if choosing $\theta\in[1/2,1)$ and $\theta\in(0,1/2)$, respectively.
\end{cor}

\begin{remark}
	Theorem~\ref{thm:eFIMA} and Corollary~\ref{cor:eFIMA} actually provide us a unified methodology to analyze the convergence issues for not only learning-based methods, but also classical nonconvex solvers. That is, on the one hand, within eFIMA, we can provide an easily-implemented and strictly convergent way to extend almost all the learning-based methods reviewed in Sec.~\ref{sec:learning-based}.
	On the other hand, by respectively specifying $\mathcal{A}_g$ and $\mathcal{A}_f$ as proximal operation and Nesterov's acceleration, eFIMA will reduce to the classical nonconvex APG, thus we can also obtain the same convergence results for a variety of prevalent APG methods  \cite{li2015accelerated,yao2016more,li2017convergence}. 
\end{remark}

\subsection{Implicit Momentum via Error Control}\label{sec:error-control}

Indeed, even with the explicit momentum schedule, we may still not obtain a globally convergent iteration. This is mainly because that there is no policy to efficiently control the inexactness of the user-specified modules (i.e., $\mathcal{A}$). In this subsection, we show how address this issue by controlling the first-order optimality error during iterations.

Specifically, we consider the auxiliary of $\Psi$ at $\mathbf{x}^k$ (denoted as $\Psi^k$) and denote its sub-differential (denoted as $\mathbf{d}_{\Psi^k}^{\mathbf{x}}$)\footnote{Strictly speaking, $\partial\Psi^k(\mathbf{x})$ is the so-called limiting Frech\'et sub-differential. We state its formal definition and propose a practical computation scheme for $\mathbf{d}_{\Psi^k}^{\tilde{\mathbf{u}}}$ in Supplemental Materials.} as
\begin{equation}
	\begin{array}{l}
		\Psi^k(\mathbf{x})=f(\mathbf{x}) + g(\mathbf{x}) +\frac{\mu^k}{2}\|\mathbf{x}-\mathbf{x}^k\|^2,\\
		\mathbf{d}_{\Psi^k}^{\mathbf{x}}=\mathbf{d}_{g}^{\mathbf{x}} + \nabla f\left(\mathbf{x}\right) + \mu^{k}(\mathbf{x} - \mathbf{x}^{k})\in\partial\Psi^k(\mathbf{x}),
	\end{array}\label{eq:error}
\end{equation}
where $\mu^k>0$ is the penalty parameter and $\mathbf{d}_{g}^{\mathbf{x}}\in\partial g(\mathbf{x})$. 

As shown in Alg.~\ref{alg:iFIMA}, at stage $k$, a variable $\tilde{\mathbf{u}}^k$ is obtained by proximally minimizing 
$\Psi^k$ at $\mathbf{u}^k$ (i.e., Step~3 of Alg.~\ref{alg:iFIMA}). Roughly, this new variable is just an ensemble of the last updated $\mathbf{x}^k$ and the output $\mathbf{u}^k$ of user-specified $\mathcal{A}$ following the specific proximal structure in Eq.~\eqref{eq:psi}. Then the monitor is obtained by checking the boundedness of $\mathbf{d}_{\Psi^k}^{\tilde{\mathbf{u}}}$. Notice that the constant $C^k$ actually reveals our tolerance to the inexactness of $\mathcal{A}$ at $k$-th iteration. 

\begin{algorithm}
	\caption{Implicit Momentum FIMA (iFIMA)}\label{alg:iFIMA}
	\begin{algorithmic}[1]
		\REQUIRE $\mathbf{x}^0$, $\mathcal{A}=\{\mathcal{A}_g, \mathcal{A}_f\}$, $\{0< 2C^{k} < \mu^{k}<\infty\}$, and $\{0< \gamma^{k} < 1/L \}$.
		\WHILE{not converged}
		\STATE $\mathbf{u}^{k}=\mathcal{A}_{g}\circ\mathcal{A}_f(\mathbf{x}^{k})$.
		\STATE $\tilde{\mathbf{u}}^{k}=\mathtt{prox}_{{\gamma^k}g}\left(\mathbf{u}^{k}-\gamma^{k}\left(\nabla f(\mathbf{u}^{k}) + \mu^{k}(\mathbf{u}^{k} - \mathbf{x}^{k}) \right) \right)$.\label{step:ss}
		\IF{$\|\mathbf{d}_{\Psi^k}^{\tilde{\mathbf{u}}^{k}} \|\leq C^k\|\tilde{\mathbf{u}}^{k}-\mathbf{x}^k\|$}
		\STATE $\mathbf{v}^{k}=\tilde{\mathbf{u}}^{k}$.
		\ELSE
		\STATE $\mathbf{v}^{k}=\mathbf{x}^k$.
		\ENDIF\label{step:se}
		\STATE $\mathbf{x}^{k+1}=\mathtt{prox}_{{\gamma^k}g}\left(\mathbf{v}^{k}-\gamma^k\nabla f(\mathbf{v}^{k})\right)$.
		\ENDWHILE	
	\end{algorithmic}
\end{algorithm}

\begin{prop}\label{prop:x-u}
	Let $\{\mathbf{x}^k,\tilde{\mathbf{u}}^k,\mathbf{v}^k\}_{k\in\mathbb{N}}$ be the sequences generated by Alg.~\ref{alg:iFIMA}. Then there exist two sequences $\{\alpha^k | \alpha^k>0\}_{k\in\mathbb{N}}$ and  $\{\beta^k |\beta^k>0\}_{k\in\mathbb{N}}$, such that the inequality~\eqref{eq:func_increase} in Theorem~\ref{thm:eFIMA} and 
	$\Psi(\tilde{\mathbf{u}}^k) \leq \Psi(\mathbf{x}^{k}) -  \beta^k\|\tilde{\mathbf{u}}^k-\mathbf{x}^{k}\|^2$ are respectively satisfied.
\end{prop}

Equipped with Proposition~\ref{prop:x-u}, it will be straightforward to guarantee that the objective values  generated by Alg.~\ref{alg:iFIMA} (i.e., $\{\Psi(\mathbf{x}^k)\}_{k\in\mathbb{N}}$) also has sufficient descent. So we call this version of FIMA as implicit momentum FIMA (iFIMA). Then the global convergence of iFIMA is proved as follows.
\begin{thm}\label{thm:iFIMA}
	Let $\{\mathbf{x}^k\}_{k\in\mathbb{N}}$  be the sequence generated by iFIMA. Then
	$\{\mathbf{x}^k\}_{k\in\mathbb{N}}$ is bounded and any of its accumulation points are the critical points of $\Psi$. If $\Psi$ is semi-algebraic, we further have that $\{\mathbf{x}^k\}_{k\in\mathbb{N}}$
	is a Cauchy sequence, thus globally converges to a critical point of $\Psi(\mathbf{x})$ in Eq.~\eqref{eq:psi}. 
\end{thm}
Indeed, based on Theorem~\ref{thm:iFIMA}, it is also easy to obtain the same convergence rate as that in Corollary~\ref{cor:eFIMA} for iFIMA.
\begin{remark}
	The results in Theorem~\ref{thm:iFIMA} is even better than that for prevalent nonconvex APGs. This actually suggests that our devised error-control policy together with the flexible algorithmic structures should also be beneficial for classical nonconvex algorithms.
\end{remark}
\begin{remark}Theorems~\ref{thm:eFIMA} and \ref{thm:iFIMA} indicate that the convergence of FIMA does not depend on the particular choices of $\mathcal{A}_f$ and $\mathcal{A}_g$ in general. This allows us to utilize different types of iterative modules, such as classical numerical schemes, off-the-shelf methods, and deep networks. 
\end{remark}
\begin{remark}
	However, it will be shown in Sec.~\ref{sec:exp} that the choices of $\mathcal{A}_f$ and $\mathcal{A}_g$ do affect our speed and accuracy in practice. This is because in FIMA, the scheduling of learnable  and numerical modules are automatically and adaptively adjusted, so that improper $\mathcal{A}_f$ or $\mathcal{A}_g$ will directly result in too many expensive refinements. 
\end{remark}

\subsubsection{Practical Calculation of $\mathbf{d}^{\tilde{\mathbf{u}}^{k}}_{\Psi^k}$ in iFIMA}
Here we propose a practical calculation scheme for  $\mathbf{d}^{\tilde{\mathbf{u}}^{k}}_{\Psi^k}\in\partial\Psi^k(\tilde{\mathbf{u}}^k)$ defined in Eq.~\eqref{eq:error} and used in Alg.~\ref{alg:iFIMA}. In fact, it is challenging to directly calculate $\mathbf{d}^{\tilde{\mathbf{u}}^{k}}_{\Psi^{k}}$ since the sub-differential $\mathbf{d}^{\tilde{\mathbf{u}}^{k}}_{g}$ is often intractable in the non-convex scenario.
Fortunately, our following analysis provides an efficient practical calculation scheme for $\mathbf{d}^{\tilde{\mathbf{u}}^{k}}_{\Psi^k}$ within FIMA framework.
Specifically, from Alg.~\ref{alg:iFIMA}, we have
\begin{equation}
\begin{array}{l}
\tilde{\mathbf{u}}^{k}\in\mathtt{prox}_{{\gamma^k}g}\left(\mathbf{u}^{k}-\gamma^{k}\left(\nabla f(\mathbf{u}^{k}) + \mu^{k}(\mathbf{u}^{k} - \mathbf{x}^{k}) \right) \right).
\end{array}\label{eq:d-u}
\end{equation}
On the other hand, from definition in Eq.~\eqref{eq:error}, we have
\begin{equation}
\begin{array}{l}
\quad \mathbf{d}^{\tilde{\mathbf{u}}^{k}}_{\Psi^k} = \mathbf{d}^{\tilde{\mathbf{u}}^{k}}_{g} + \nabla f(\tilde{\mathbf{u}}^{k}) + \mu^{k}(\tilde{\mathbf{u}}^{k} - \mathbf{x}^{k}) \\
\Rightarrow \mathbf{d}^{\tilde{\mathbf{u}}^{k}}_{g} = \mathbf{d}^{\tilde{\mathbf{u}}^{k}}_{\Psi^k} - \nabla f(\tilde{\mathbf{u}}^{k}) - \mu^{k}(\tilde{\mathbf{u}}^{k} - \mathbf{x}^{k}) \in \partial g(\tilde{\mathbf{u}}^{k}).
\end{array}
\end{equation}
By the property of proximal operation, we have
\begin{equation}
\begin{array}{l}
0 \in \gamma^{k} (\partial g(\tilde{\mathbf{u}}^{k}) - \mathbf{d}^{\tilde{\mathbf{u}}^{k}}_{g}) 
=  \gamma^{k} \partial g(\tilde{\mathbf{u}}^{k}) +\tilde{\mathbf{u}}^{k}-( \tilde{\mathbf{u}}^{k} + \gamma^{k}\mathbf{d}^{\tilde{\mathbf{u}}^{k}}_{g} )\\
\Leftrightarrow \tilde{\mathbf{u}}^{k} \in \mathtt{prox}_{\gamma^kg}\left( \tilde{\mathbf{u}}^{k} + \gamma^{k}\mathbf{d}^{\tilde{\mathbf{u}}^{k}}_{g} \right) \\
\Leftrightarrow \tilde{\mathbf{u}}^{k} \in  \mathtt{prox}_{\gamma^kg}\left( \tilde{\mathbf{u}}^{k}
-\gamma^{k} \left( \nabla f(\tilde{\mathbf{u}}^{k}) + \mu^{k}(\tilde{\mathbf{u}}^{k} - \mathbf{x}^{k})\right) \right.\\
\qquad \qquad \qquad  \qquad + \gamma^{k}\mathbf{d}^{\tilde{\mathbf{u}}^{k}}_{\Psi^{k}} ).
\end{array} \label{eq:error-com}
\end{equation}
Therefore, by comparing Eqs.~\eqref{eq:d-u} and \eqref{eq:error-com}, we actually have the following practically calculation scheme for $\mathbf{d}^{\tilde{\mathbf{u}}^{k}}_{\Psi^k}$:
$$\mathbf{d}^{\tilde{\mathbf{u}}^{k}}_{\Psi^k}=\left(\mu^{k} - 1/ \gamma^k \right)\left( \tilde{\mathbf{u}}^{k} - \mathbf{u}^{k}\right)-\left(\nabla f\left(\mathbf{u}^{k}\right)-\nabla f\left(\tilde{\mathbf{u}}^{k}\right)\right). $$


\subsection{Multi-block Extension}

In order to tackle the inverse problems with blocks of unknown variables (e.g., blind deconvolution and dictionary learning), we now discuss how to extend FIMA for multi-block NIPs, which is formulated as
$\mathcal{T}(\mathfrak{X})=\mathbf{y} + \mathbf{n}$, where
$\mathfrak{X}=\{\mathbf{x}_n\}_{n=1}^N\in\mathbb{R}^{D_1}\times\cdots\times\mathbb{R}^{D_N}$ is a set of $N \geq 2$ unknown variables to be estimated. Notice that here $\mathcal{T}$ should be some given linear operations on $\mathfrak{X}$. The inference of such problem can be addressed by solving
\begin{equation}
	\min\limits_{\mathfrak{X}}\Psi(\mathfrak{X}):=f(\mathfrak{X};\mathcal{T},\mathbf{y})+\sum_{n=1}^Ng_n(\mathbf{x}_n),\label{eq:psi-m}
\end{equation}
where $f(\mathfrak{X}):\mathbb{R}^{D_1}\times\cdots\times\mathbb{R}^{D_N}\to (-\infty,+\infty]$ is still differentiable and each $g_n(\mathbf{x}_n):\mathbb{R}^{D_n}\to(-\infty,+\infty]$ may also nonsmooth and possibly nonconvex. 
Here both $f$ and block-wise $g_n$ ($\mathbf{x}_n$) follow the same assumptions as that in Eq.~\eqref{eq:psi} and $f$ should also satisfy the generalized Lipschitz smooth property on bounded subsets of $\mathbb{R}^{D_1}\times\cdots\times\mathbb{R}^{D_N}$.
For ease of presentation, we denote $\mathfrak{X}_{[<n]}=\{\mathbf{x}_i\}_{i=1}^{n-1}$, $\mathfrak{X}_{[\leq n]}=\{\mathbf{x}_i\}_{i=1}^{n}$ and
the subscripts ${[>n]}$ and ${[\geq n]}$ are defined in the same manner.
Then we summarize the main iterations of multivariable FIMA (mFIMA) as follows\footnote{Due to space limit, the details of mFIMA are presented in Supplemental Material.}:
$$
\begin{array}{l}
\mathbf{u}^{k}_{n}=\mathcal{A}_{g_{n}}\circ\mathcal{A}_f\left(\mathfrak{X}^{k+1}_{[< n]},\mathfrak{X}^{k}_{[\geq n]}\right),\\
\mathbf{x}^{k+1}_{n}=\mathtt{prox}_{{\gamma^k}g_{n}}\left(\mathbf{v}^{k}_{n}-\gamma^k\nabla_{n} f\left(\mathfrak{X}^{k+1}_{[< n]},\mathbf{v}^{k}_{n},\mathfrak{X}^{k}_{[> n]}\right)\right).
\end{array}
$$
Here $\mathbf{v}^k_n$ is the monitor of $\mathbf{x}_n^k$, obtained by the same error control strategy as that in iFIMA. Then we summarize our multi-block FIMA in Alg.~\ref{alg:MFIMA} and prove the convergence of mFIMA in Corollary~\ref{cor:MFIMA-ims}. 

\begin{cor}\label{cor:MFIMA-ims}
	Let $\{\mathfrak{X}^k\}_{k\in\mathbb{N}}$ be the sequence
	generated by mFIMA. Then we have the same convergence properties as that in Theorem~\ref{alg:iFIMA} and Corollary~\ref{cor:eFIMA} for $\{\mathfrak{X}^k\}_{k\in\mathbb{N}}$.
\end{cor}

Then we summarize our multi-block FIMA in Alg.~\ref{alg:MFIMA}. Notice that here we adopt the error-control policy in iFIMA to guide the iterations of mFIMA.

\begin{algorithm}
	\caption{Multi-block FIMA}\label{alg:MFIMA}
	\begin{algorithmic}[1]
		\REQUIRE $\mathfrak{X}^0$, $\mathcal{A} = \{\mathcal{A}_{g_{1}}, \cdots, \mathcal{A}_{g_N}, \mathcal{A}_f\}$, $\{0< 2C_n^{k} < \mu_n^{k}<\infty\}$, and $\{0< \gamma_n^{k} < 1/L_n \}$.
		\WHILE{not converged}
		\FOR{$n=1:N$}
		\STATE $\mathbf{u}^{k}_{n}=\mathcal{A}_{g_{n}}\circ\mathcal{A}_f\left(\mathfrak{X}^{k+1}_{[< n]},\mathfrak{X}^{k}_{[\geq n]} \right)$.
		\STATE 
		$\tilde{\mathbf{u}}_{n}^{k}\in\mathtt{prox}_{{\gamma_n^k}g_{n}}(\mathbf{u}_{n}^{k}-\gamma_n^{k}(\nabla_{n} f(\mathfrak{X}_{[<n]}^{k+1},\mathbf{u}_n^k,\mathfrak{X}_{[>n]}^{k}) $
		$\qquad +  \mu_{n}^{k}(\mathbf{u}_{n}^{k} - \mathbf{x}_{n}^{k}) ) ).$
		\IF{$\|\mathbf{d}_{\Psi_{n}^k}^{\tilde{\mathbf{u}}_{n}^{k}} \|\leq C_{n}^k\|\tilde{\mathbf{u}}_{n}^{k}-\mathbf{x}_{n}^k\|$}
		\STATE $\mathbf{v}_{n}^{k}=\tilde{\mathbf{u}}_{n}^{k}$.
		\ELSE
		\STATE $\mathbf{v}_{n}^{k}=\mathbf{x}_{n}^k$.
		\ENDIF
		\STATE $\mathbf{x}^{k+1}_{n} \in\mathtt{prox}_{{\gamma_n^k}g_{n}}\left(\mathbf{v}^{k}_{n}-\gamma_n^k\nabla_{n} f\left(\mathfrak{X}^{k+1}_{[< n]},\mathbf{v}^{k}_{n},\mathfrak{X}^{k}_{[> n]}\right)\right)$. \label{step:Mprox}
		\ENDFOR
		\ENDWHILE	
	\end{algorithmic}
\end{algorithm}

\section{Applications}\label{sec:app}

As a nontrivial byproduct, this section illustrates how to apply FIMA to tackle practical inverse problems in low-level vision area, such as image deconvolution in the standard non-blind and even more challenging blind scenarios.

%
%

\textbf{Non-blind Deconvolution (Uni-block)} aims to restore the latent image $\mathbf{z}$ from corrupted observation $\mathbf{y}$ with known blur kernel $\mathbf{b}$. In this part, we utilize the well-known sparse coding formulation~\cite{beck2009a}: $\mathbf{y}=\mathbf{D}\mathbf{x}+\mathbf{n}$, where $\mathbf{x}$, $\mathbf{D}$ and $\mathbf{n}$ are
the sparse code, given dictionary and unknown noises, respectively. Indeed, the form of $\mathbf{D}$ is given as $\mathbf{D}=\mathbf{B}\mathbf{W}^{\top}$, where $\mathbf{B}$ is the matrix form of $\mathbf{b}$, $\mathbf{W}^{\top}$ denotes the inverse of the wavelet transform $\mathbf{W}$ (i.e., $\mathbf{x}=\mathbf{W}\mathbf{z}$ and $\mathbf{z}=\mathbf{W}^{\top}\mathbf{x}$).
So by defining $f(\mathbf{x};\mathbf{D},\mathbf{y})=\|\mathbf{y} - \mathbf{D}\mathbf{x}\|^2$ and $g(\mathbf{x})=\lambda\|\mathbf{x}\|_{p}$ ($0\leq p<1$), we obtain a special case of Eq.~\eqref{eq:psi} as follows
\begin{equation}
	\min\limits_{\mathbf{x}} f(\mathbf{x};\mathbf{D},\mathbf{y}) + g(\mathbf{x}).
	\label{eq:task-sparse-coding}
\end{equation}

Now we are ready to design iterative modules (i.e., $\mathcal{A}_f$ and $\mathcal{A}_g$) to optimize the SC model in Eq.~\eqref{eq:task-sparse-coding}. With the well-known imaging formulation $\mathbf{y}=\mathbf{b}\otimes\mathbf{z}+\mathbf{n}$ ($\otimes$ denotes the convolution operator), we actually update $\mathbf{z}$ by solving
$\mathcal{A}_f(\mathbf{z}^k):=\arg\min_{\mathbf{z}} \|\mathbf{y} - \mathbf{b}\otimes\mathbf{z}\|^2 +\tau\|\mathbf{z}- \mathbf{z}^{k}\|^2$ to aggregate principles of the task and information from last updated variable,
where $\mathbf{z}^{k}=\mathbf{W}^{\top}\mathbf{x}^{k}$ and $\tau$ is a positive constant. Then $\mathcal{A}_f$ on $\mathbf{x}$ can be defined as
$\mathcal{A}_f(\mathbf{x}^k)=\mathbf{W}\mathcal{A}_f(\mathbf{z}^k)$, i.e.,
\begin{equation}
	\mathcal{A}_f(\mathbf{x}^k)=\mathbf{W}(\mathbf{B}^T\mathbf{B}+\tau\mathbf{I})^{-1}\left(\mathbf{B}^T\mathbf{y}+\tau\mathbf{W}^{\top}\mathbf{x}^{k}\right),\label{eq:af-s}
\end{equation}
where  $\mathbf{I}$ is the identity matrix.
It is easy to check that $\mathcal{A}_f$ can be efficiently calculated by FFT~\cite{schmidt2014shrinkage}. 


\textbf{Blind Deconvolution (Multi-block)} involves the joint estimation of both the latent image $\mathbf{z}$ and blur kernel $\mathbf{b}$, given only an observed $\mathbf{y}$. 
Here we formulate this problem on image gradient domain and solve the following special case of Eq.~\eqref{eq:psi-m} with two unknown variables $(\mathbf{x},\mathbf{b})$\footnote{Notice that in this section, $\mathbf{x}$ is defined with different meanings, i.e., image gradient in Eq.~\eqref{eq:task-map}, while sparse code in Eq.~\eqref{eq:task-sparse-coding}.}:
\begin{equation}
	\min\limits_{\mathbf{x},\mathbf{b}} f(\mathbf{x},\mathbf{b};\nabla\mathbf{y}) + g_{\mathbf{x}}(\mathbf{x}) + g_{\mathbf{b}}(\mathbf{b}),\label{eq:task-map}
\end{equation}
where $f(\mathbf{x},\mathbf{b};\nabla\mathbf{y})=\|\nabla\mathbf{y}-\mathbf{b}\otimes\mathbf{x}\|^2$ , $g_{\mathbf{x}}(\mathbf{x})=\lambda_{\mathbf{x}}\|\mathbf{x}\|_{0}$, and
$g_{\mathbf{b}}(\mathbf{b})=\chi_{\Omega_{\mathbf{b}}}(\mathbf{b})$. Here $\chi_{\Omega_{\mathbf{b}}}$ is the indicator function of the set
$\Omega_{\mathbf{b}}:=\{\mathbf{b}\in\mathbb{R}^{D_{\mathbf{b}}}: [\mathbf{b}]_i\geq 0, \sum_{i=1}^{D_{\mathbf{b}}}[\mathbf{b}]_i=1\}$,
where $[\cdot]_i$ denotes the $i$-th element. So the proximal updating in mFIMA corresponding to $g_{\mathbf{x}}$ and $g_{\mathbf{b}}$ can be respectively calculated by hard-thresholding \cite{xu2011image} and simplex projection~\cite{duchi2008efficient}.
Here we need to specify three  modules (i.e., $\mathcal{A}_f$, $\mathcal{A}_{g_{\mathbf{x}}}$ and $\mathcal{A}_{g_{\mathbf{b}}}$) for miFPG. We first follow similar idea in the non-blind case to define $\mathcal{A}_f(\mathbf{x}^{k},\mathbf{b}^k)$ using the aggregated deconvolution energy
\begin{equation}
	\begin{array}{c}
		\mathcal{A}_f(\mathbf{x}^{k},\mathbf{b}^k):=\arg\min\limits_{\mathbf{x},\mathbf{b}}\|\nabla\mathbf{y}-\mathbf{b}\otimes\mathbf{x}\|^2 \\
		+ \tau_{\mathbf{x}}\|\mathbf{x}-\mathbf{x}^{k}\|^2 + \tau_{\mathbf{b}}\|\mathbf{b}-\mathbf{b}^{k}\|^2,\label{eq:af-model-m}
	\end{array}
\end{equation}
where $\tau_{\mathbf{b}}$ and $\tau_{\mathbf{x}}$ are positive constants. We then train CNNs on image gradient domain and solve $\min_{\mathbf{b}}\|\nabla\mathbf{y}-\mathbf{b}\otimes\mathbf{x}\|^2 + \lambda_{\mathbf{b}}\|\mathbf{b}\|^2$ using conjugate gradient method \cite{cho2011handling}
to formulate $\mathcal{A}_{g_{\mathbf{x}}}$ and $\mathcal{A}_{g_{\mathbf{b}}}$, respectively.

\section{Experimental Results}\label{sec:exp}

This section conducts experiments to verify our theoretical results and compares the performance of FIMA with other state-of-the-art learning-based iterative methods on real-world inverse problems. All experiments are performed on a PC with Intel Core i7 CPU at 3.4 GHz, 32 GB RAM and a NVIDIA GeForce GTX 1050 Ti GPU. More results can also be found
in Supplemental Materials.

\subsection{Non-blind Image Deconvolution }\label{sec:imresto}
We first evaluate FIMA on solving Eq.~\eqref{eq:task-sparse-coding} for image restoration. The test images are collected by \cite{schmidt2014shrinkage,Lai2016A} and different levels of Gaussian noise are further added to generate our corrupted observations.

\begin{figure}[t]
	\begin{tabular}{c@{\extracolsep{0.2em}}c@{\extracolsep{0.2em}}c}
		\includegraphics[width=0.145\textwidth]{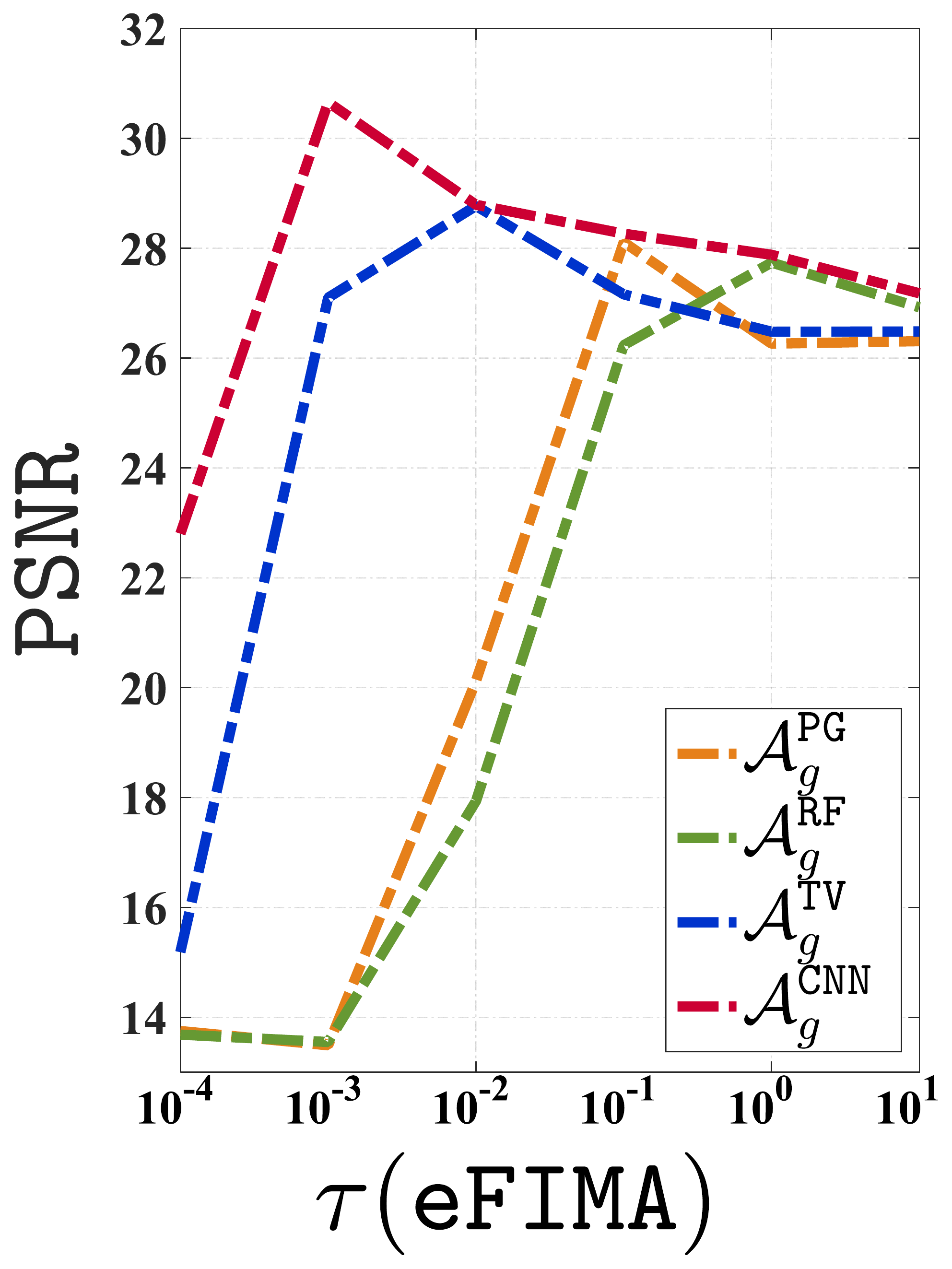}~
		&\includegraphics[width=0.145\textwidth]{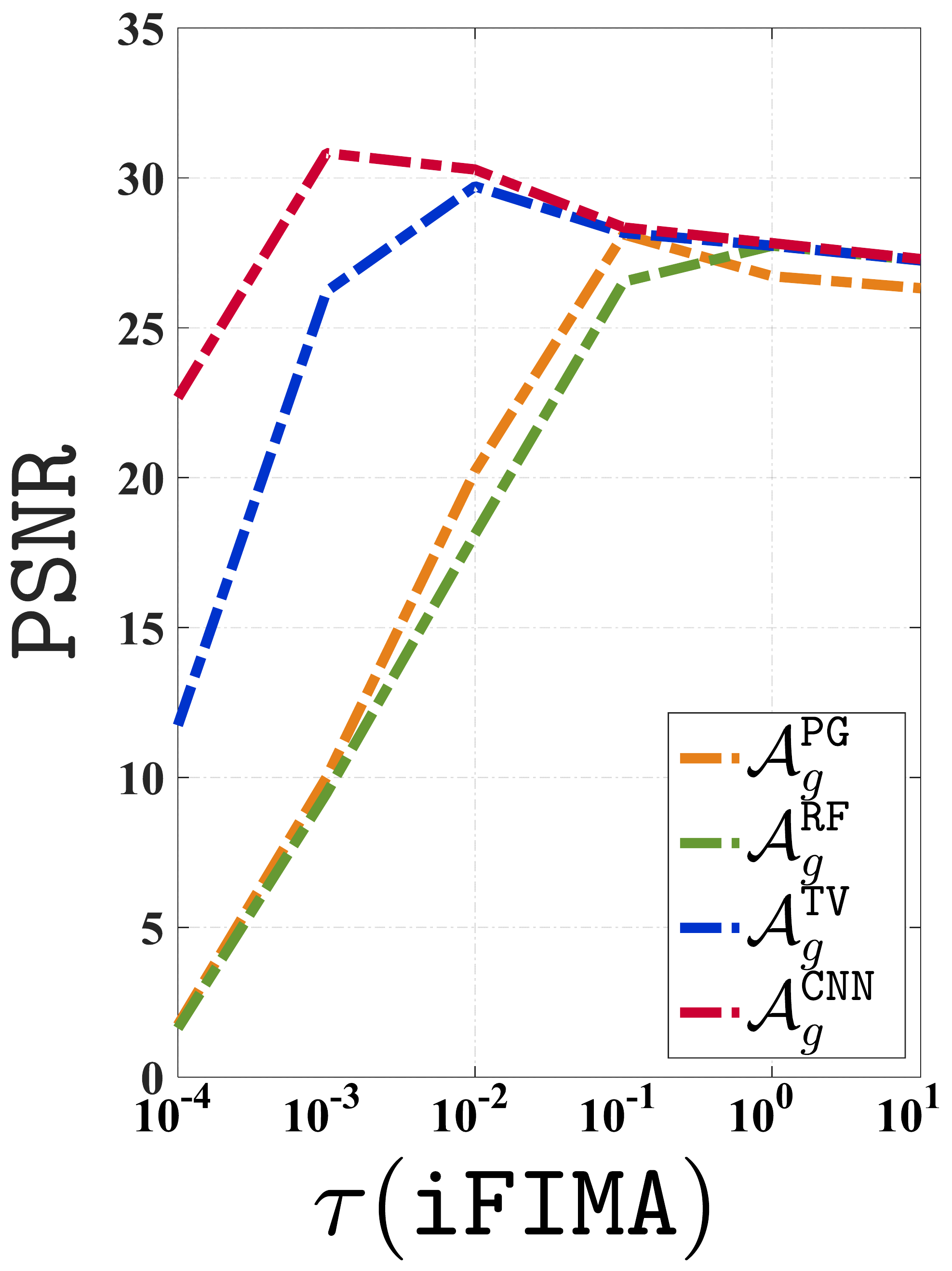}~
		&\includegraphics[width=0.155\textwidth]{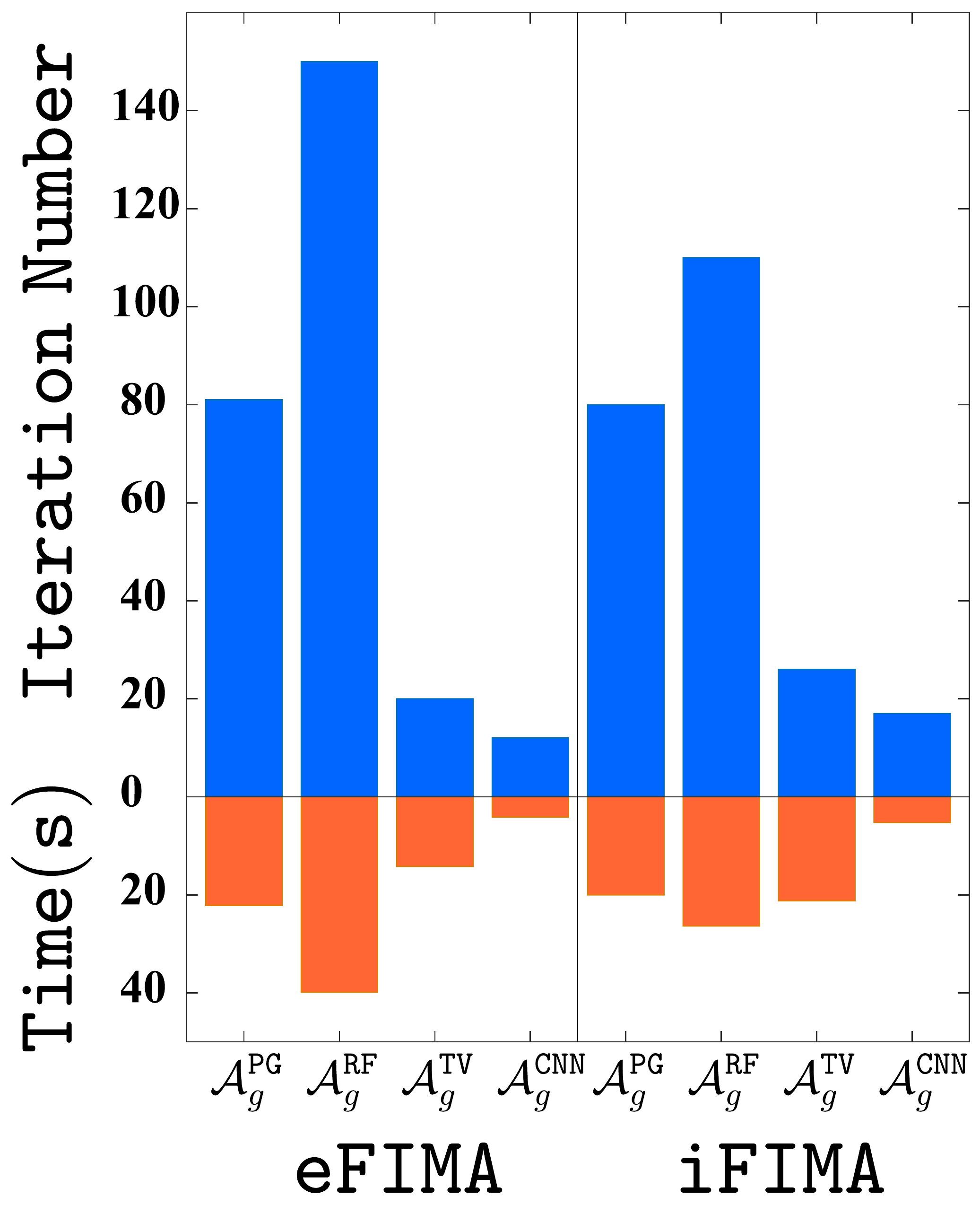}\\
		(a) & (b) & (c)\\
	\end{tabular}
	\caption{Comparisons of FIMA with different $\mathcal{A}_f^{\tau}$ ($\tau\in[10^{-4},10^1]$) and $\mathcal{A}_g\in\{\mathcal{A}_g^{\mathtt{PG}}, \mathcal{A}_g^{\mathtt{RF}}, \mathcal{A}_g^{\mathtt{TV}}, \mathcal{A}_g^{\mathtt{CNN}}\}$. The bar charts in the rightmost subfigure compares the overall iteration number and running time  (in seconds, ``Time(s)'' for short).}
	\label{fig:lambda-vary}
\end{figure}

\textbf{Modules Evaluation:}
Firstly, the influences of different choices of $\mathcal{A}$ in FIMA is studied. Following Eq.~\eqref{eq:af-s}, we adopt $\mathcal{A}_f^{\tau}$ with varying $\tau$. As for $\mathcal{A}_g$, different choices are also considered: classical PG ($\mathcal{A}_g^{\mathtt{PG}}$),
Recursive Filter \cite{unser1991recursive} ($\mathcal{A}_g^{\mathtt{RF}}$), Total Variation~\cite{wang2008a} ($\mathcal{A}_g^{\mathtt{TV}}$) and CNNs ($\mathcal{A}_g^{\mathtt{CNN}}$). For $\mathcal{A}_g^{\mathtt{CNN}}$, we introduce a residual structure $\mathbf{x}=\mathbf{x}+\mathcal{R}(\mathbf{x})$ \cite{he2016deep} and define $\mathcal{R}$ as a cascade of $7$ dilated convolution layers (with filter size $3\times 3$). ReLUs are added between each two linear layers and batch normalizations are used for the $2$-nd to $6$-th linear layers. We collect 800 images, in which 400 have been used in \cite{schmidt2014shrinkage} and the other 400 are randomly sampled from ImageNet \cite{5206848}. Here we just adopt similar strategies in \cite{Zhang2017Learning} to train $\mathcal{A}_g^{\mathtt{CNN}}$ with different noise levels. Fig.~\ref{fig:lambda-vary} analyzes the contributions of $\mathcal{A}_f^{\tau}$  ($\tau\in[10^{-4},10^1]$) and $\mathcal{A}_g\in\{\mathcal{A}_g^{\mathtt{PG}}, \mathcal{A}_g^{\mathtt{RF}}, \mathcal{A}_g^{\mathtt{TV}}, \mathcal{A}_g^{\mathtt{CNN}}\}$. We observe that $\mathcal{A}_g^{\mathtt{TV}}$ is relatively better than $\mathcal{A}_g^{\mathtt{PG}}$ and $\mathcal{A}_g^{\mathtt{RF}}$, while $\mathcal{A}_g^{\mathtt{CNN}}$ performs consistently better and faster than other strategies. So hereafter we always utilize 
$\mathcal{A}_g^{\mathtt{CNN}}$ in eFIMA and iFIMA. We also observe that even with different $\mathcal{A}_g$, relatively large $\tau$ in $\mathcal{A}_f^{\tau}$ will result in analogous quantitative results. Thus we experimentally set $\tau=10^{-3}$ for $\mathcal{A}_f^{\tau}$ in eFIMA and iFIMA for all the experiments.

\begin{figure}[t]
	\begin{tabular}{c@{\extracolsep{0.2em}}c}
		\includegraphics[width=0.22\textwidth]{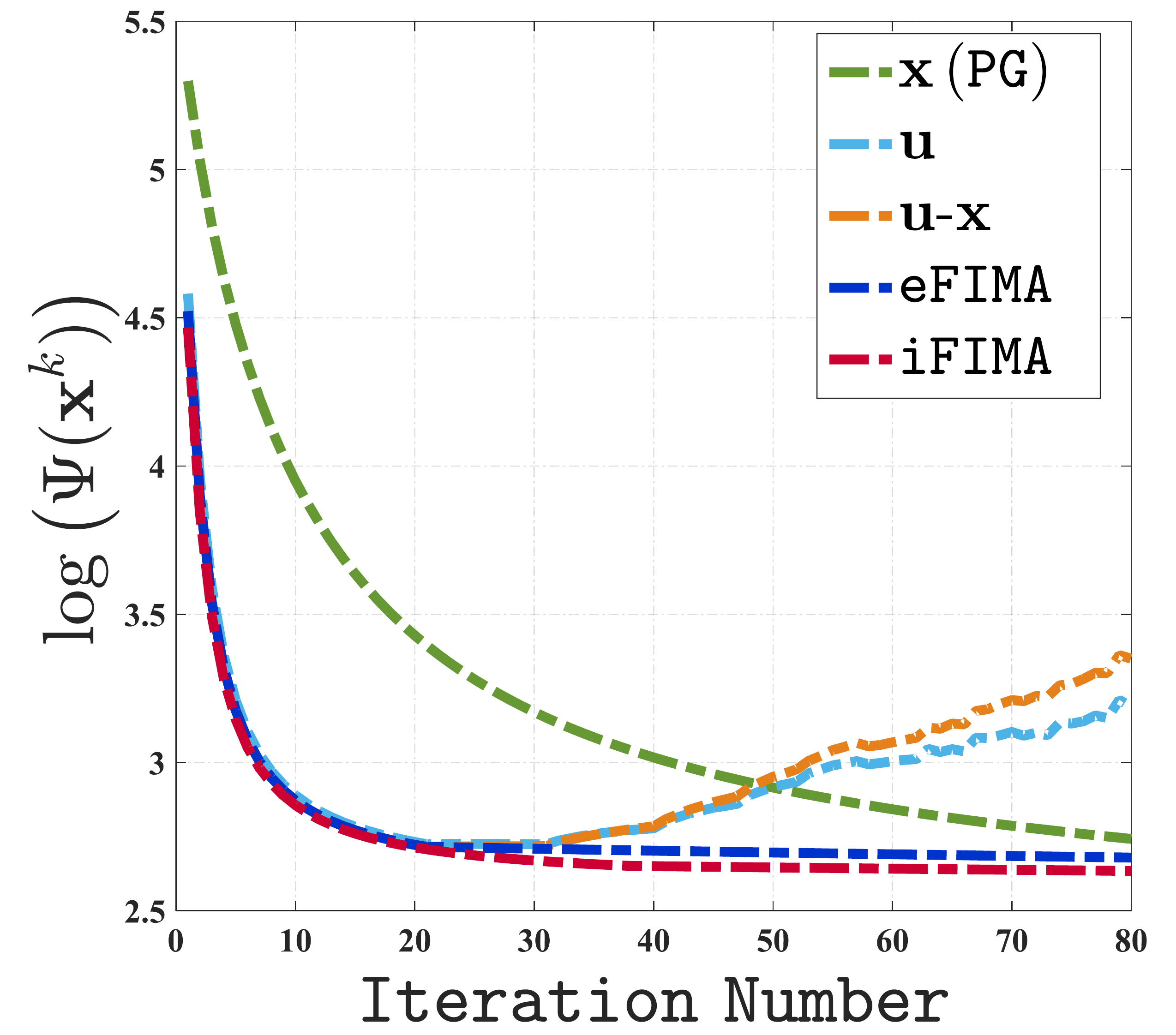}~~~
		&\includegraphics[width=0.22\textwidth]{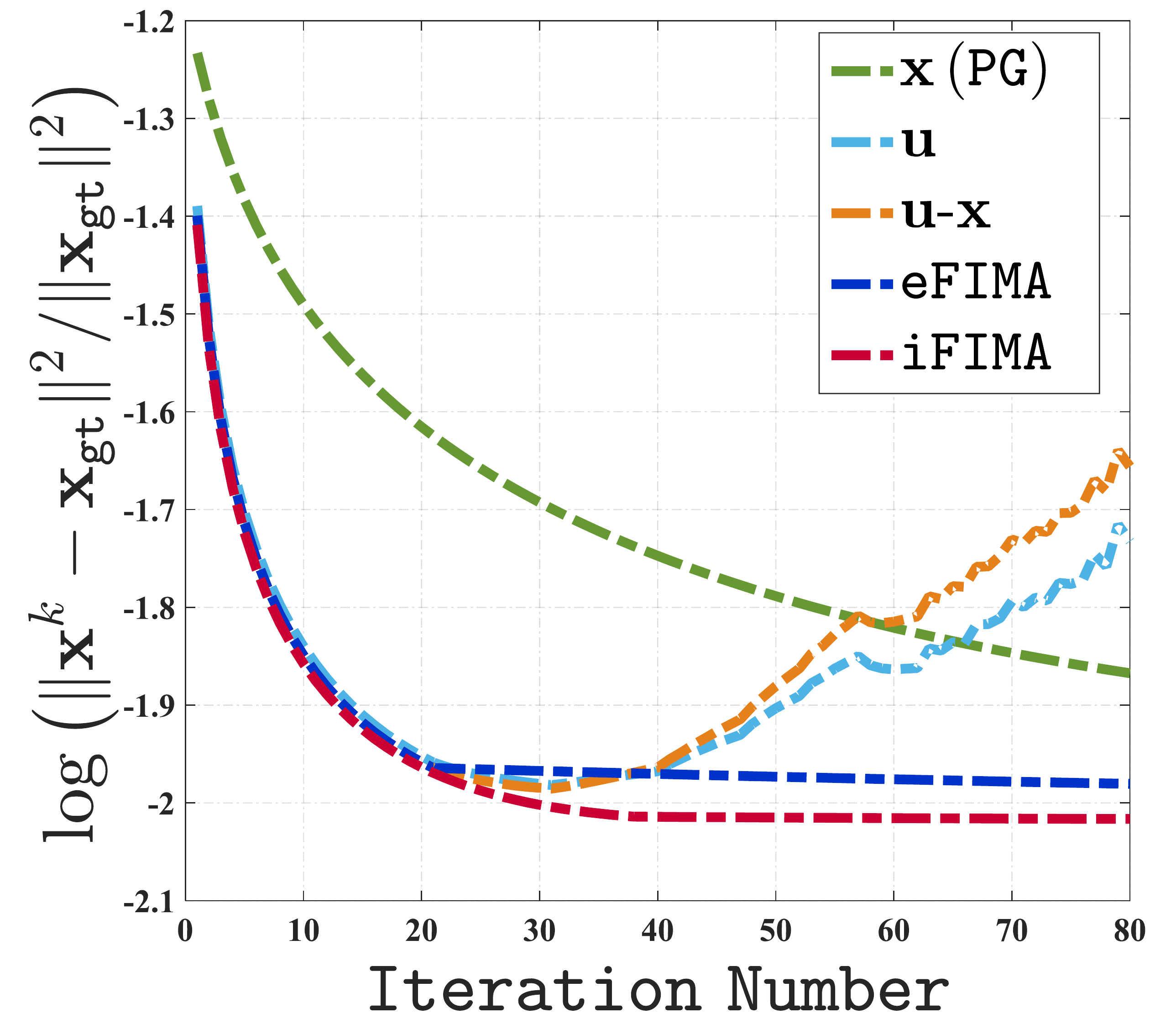} \\
		(a) & (b)\\
		\includegraphics[width=0.22\textwidth]{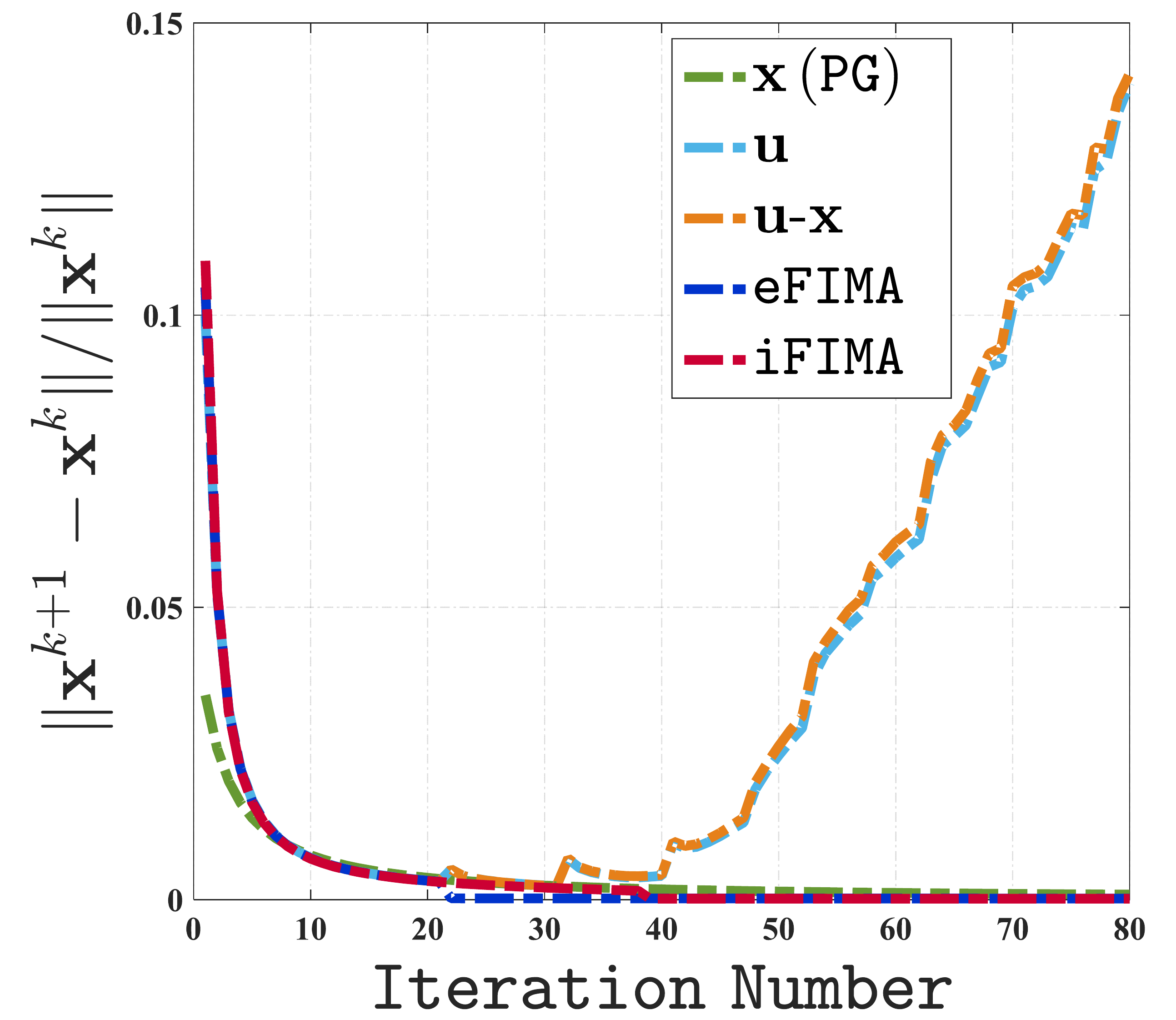}~~~
		&\includegraphics[width=0.22\textwidth]{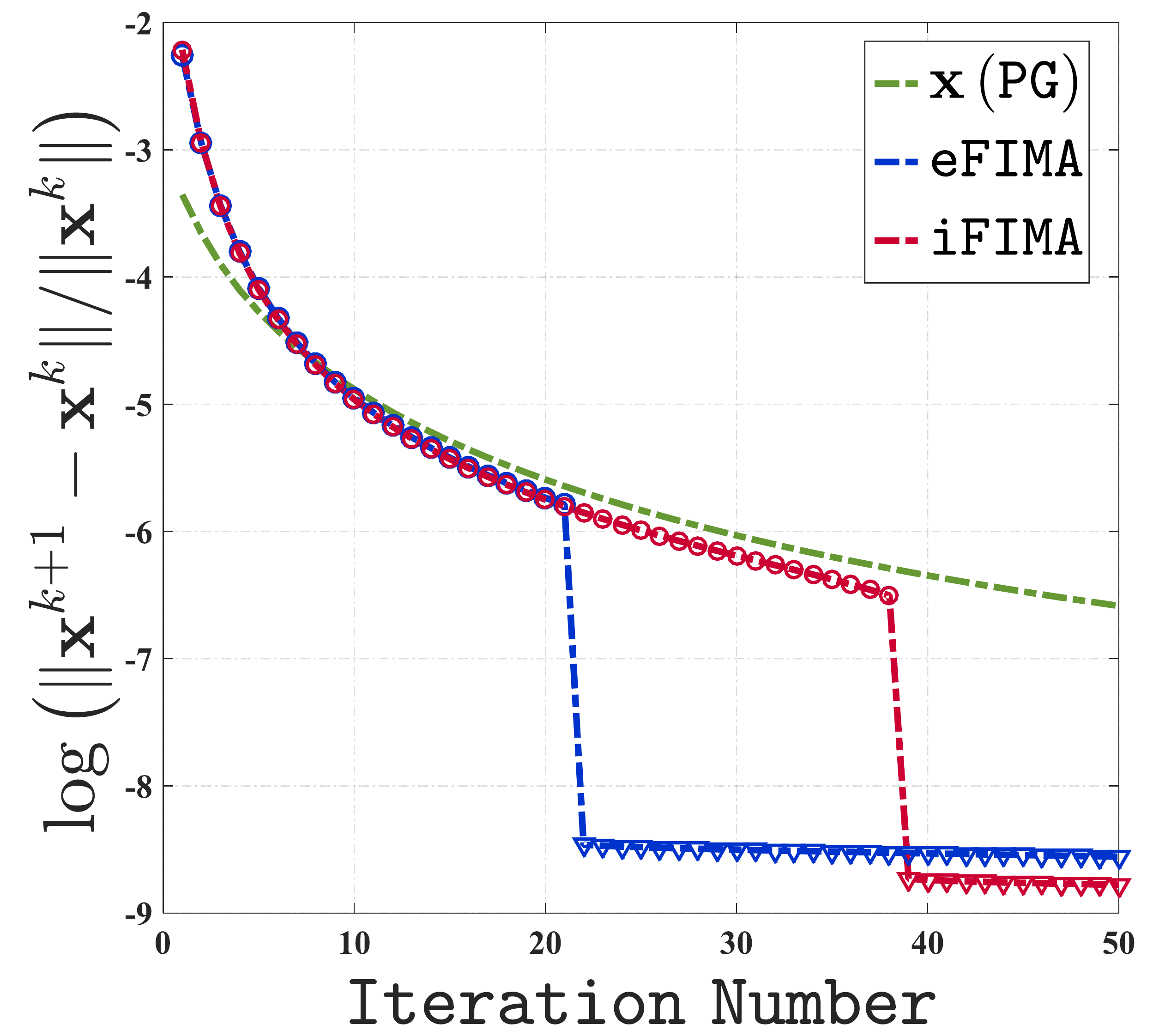} \\
		(c) & (d)\\
	\end{tabular}
	\caption{The iteration curves of FIMA with different settings. The first three subfigures express the function values, constructive errors, and iteration errors, respectively. Subfigure (d) only plots the first 50 iterations for illustrate the scheduling policies of FIMA.
}
	\label{fig:self-two-com}
\end{figure}

\begin{figure*}[ht]
	\centering
	\begin{tabular}{c@{\extracolsep{0.2em}}c@{\extracolsep{0.2em}}c@{\extracolsep{0.2em}}c@{\extracolsep{0.2em}}c}
		\includegraphics[width=0.18\textwidth]{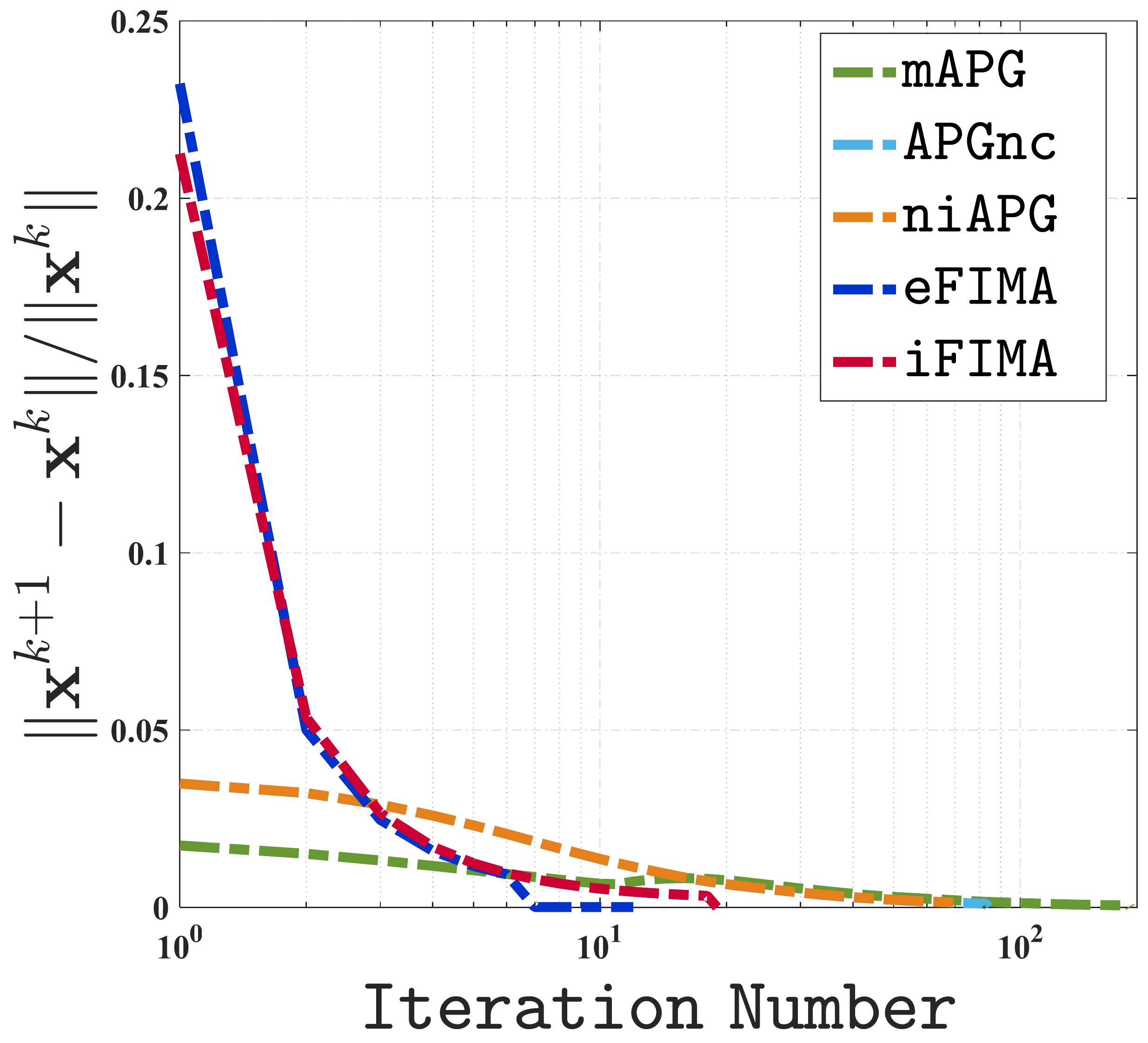}~
		&\includegraphics[width=0.175\textwidth]{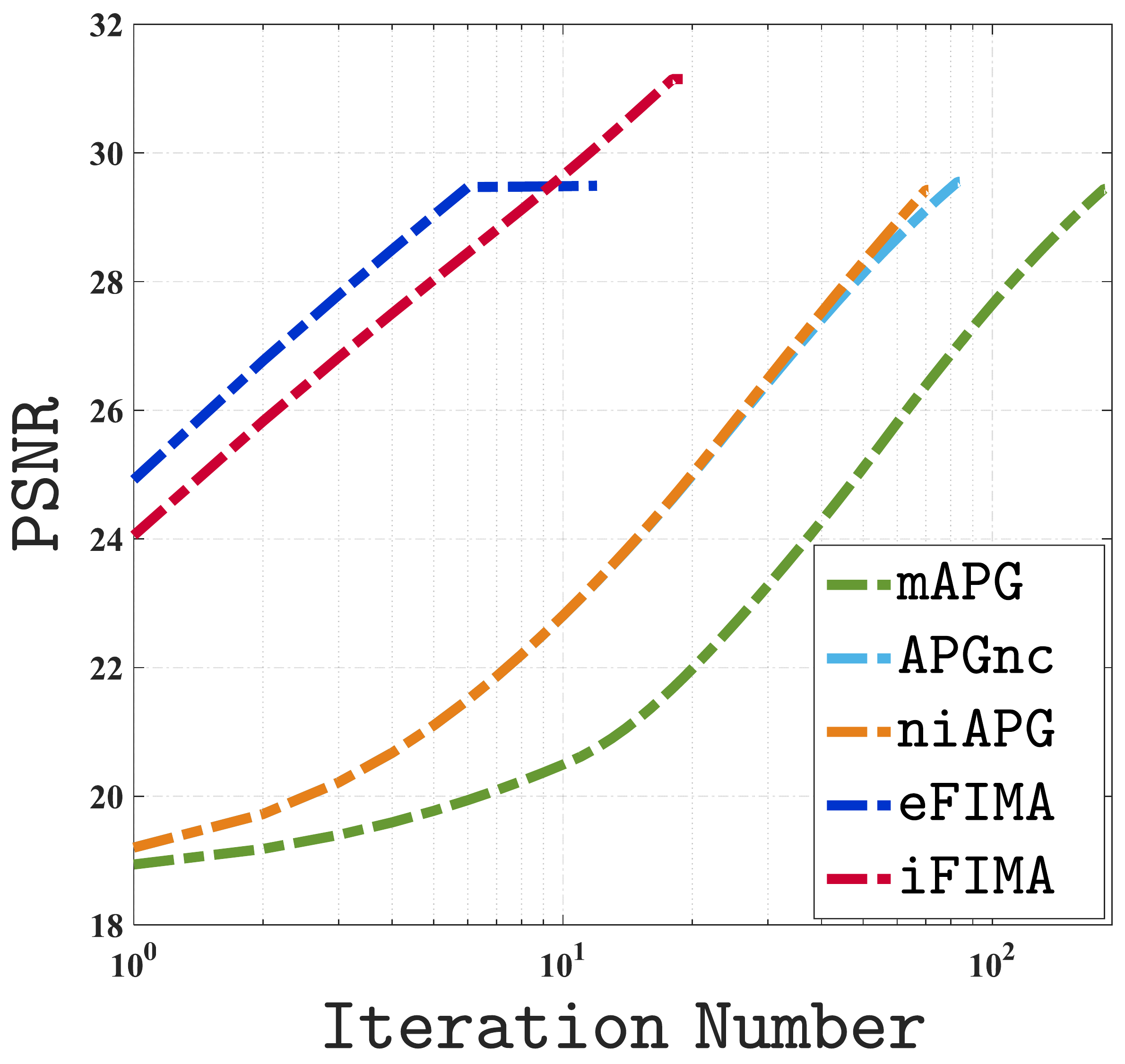}~
		&\includegraphics[width=0.18\textwidth]{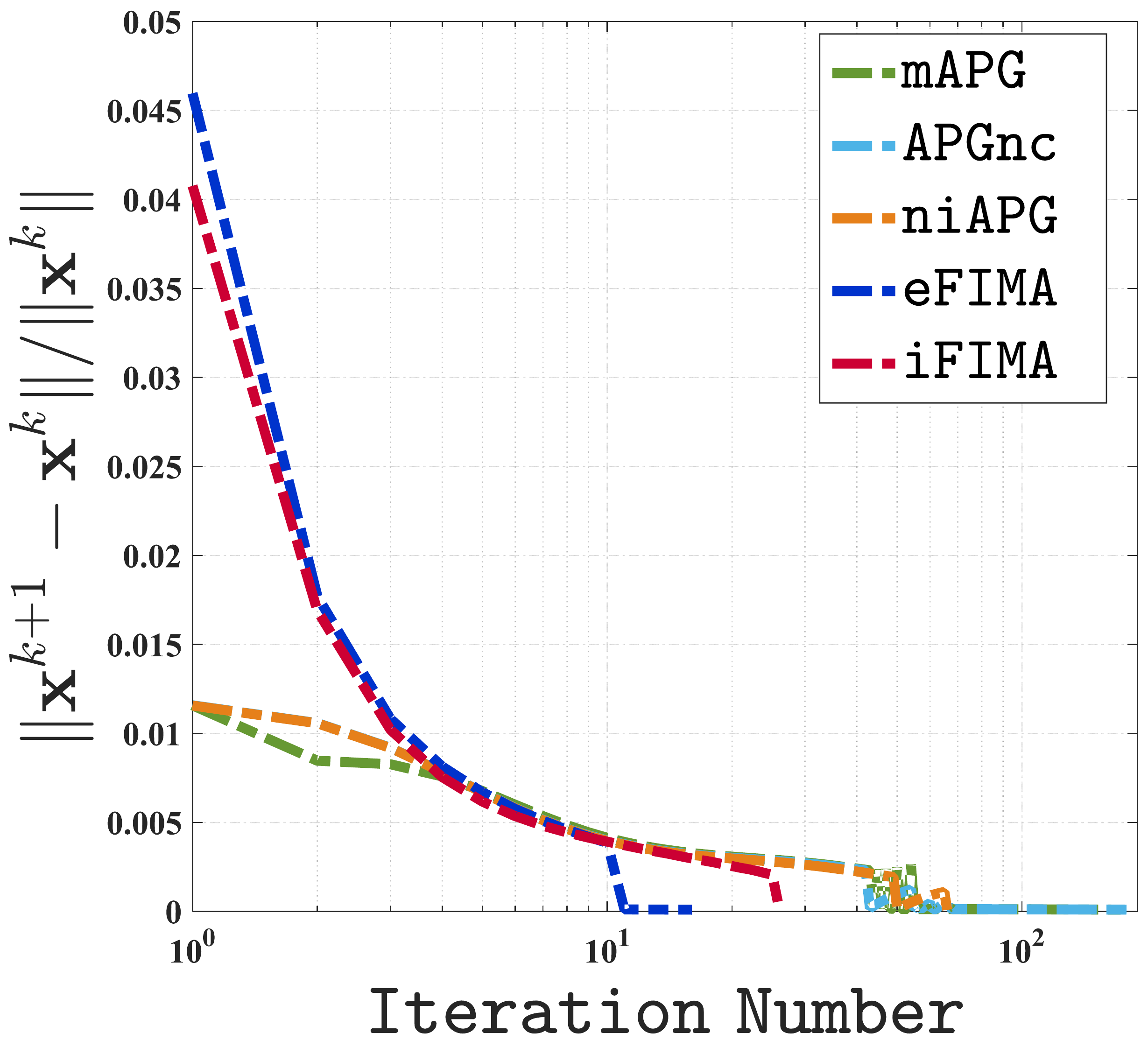}~
		&\includegraphics[width=0.18\textwidth]{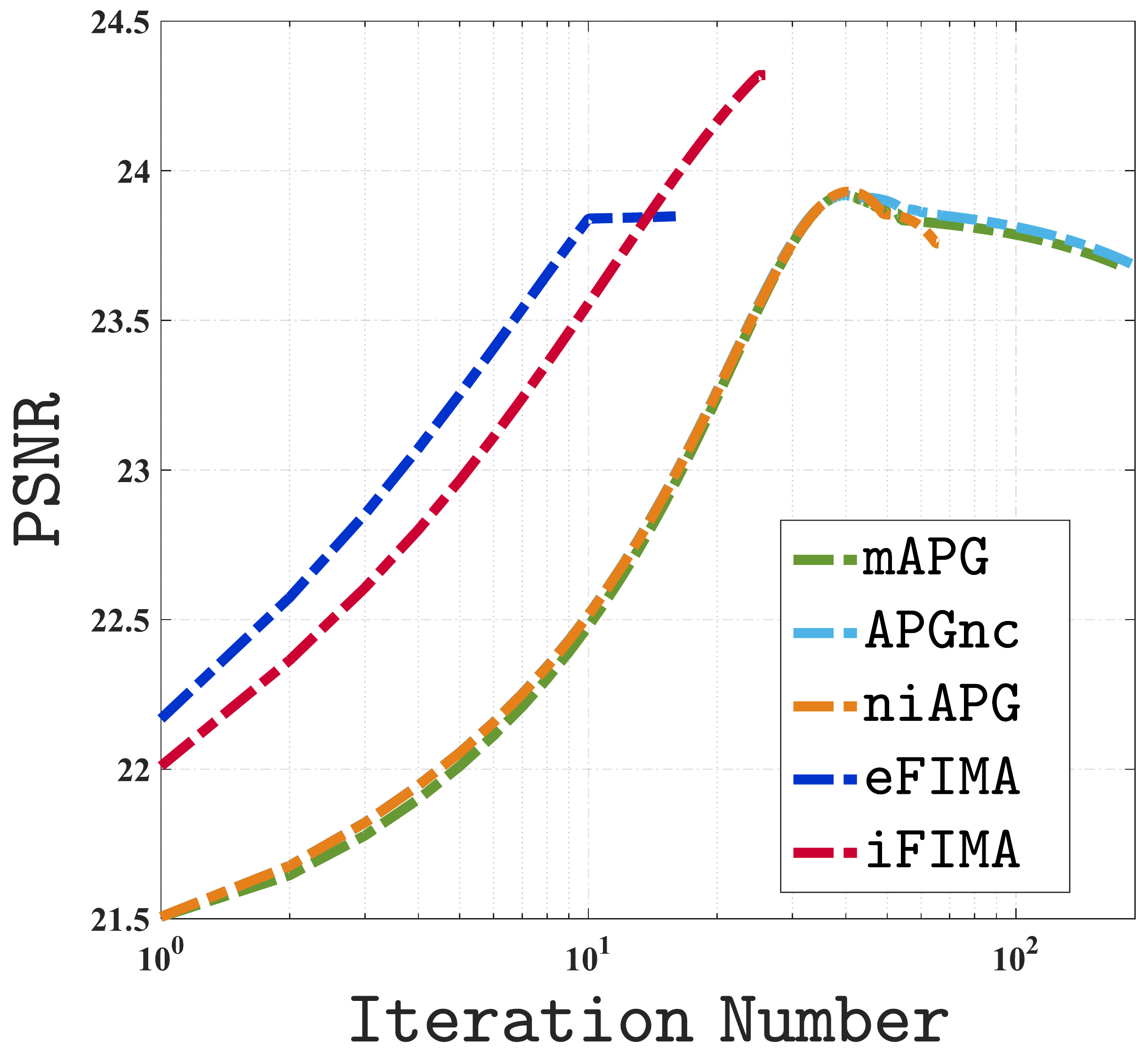}~
		&\includegraphics[width=0.195\textwidth]{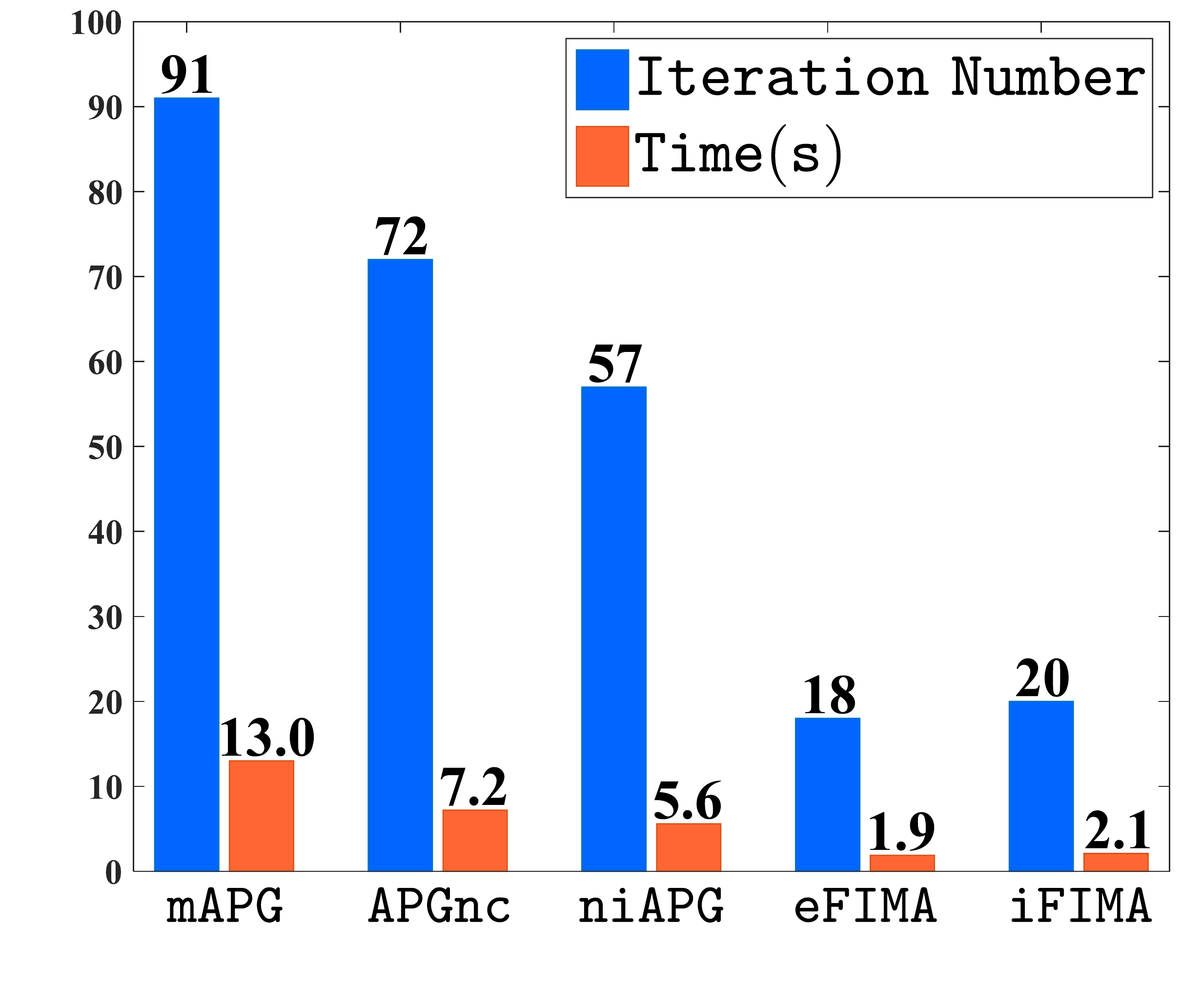}\\
		(a) $\sigma =1\text{\textperthousand}$ &(b)  $\sigma =1\text{\textperthousand}$ &(c) $\sigma = 1\text{\%}$&(d) $\sigma = 1\text{\%}$&(e)\\
	\end{tabular}
	\caption{Comparing iteration behaviors of FIMA to classical nonconvex APGs, including exact ones (mAPG, and APGnc) and inexact niAPG. The left four subfigures compare curves of iteration errors and PSNRs with different noise level (1\textperthousand~and 1\%), respectively. The rightmost subfigure plot bar charts of the averaged iteration number and ``Time(s)'' on the  dataset \cite{schmidt2014shrinkage}.}
	\label{fig:other-pgs}
\end{figure*}

\textbf{Convergence Behaviors:}
We then verify the convergence properties of FIMA. The convergence behaviors of both each module in our algorithms and other nonconvex APGs are considered. To be fair and comprehensive, we adopt specific iteration numbers ($K = 80$) and  iteration errors ($\|\mathbf{x}^{k+1}-\mathbf{x}^{k}\| / \|\mathbf{x}^{k}\| \leq 10^{-4}$) as stopping criterion in Figs.~\ref{fig:self-two-com} and \ref{fig:other-pgs}, respectively.

In Fig.~\ref{fig:self-two-com}(a), (b), and (c), we plot the curves of objective values ($\log\left(\Psi(\mathbf{x}^{k})\right)$), reconstruction errors ($\log\left(\|\mathbf{x}^{k+1}- \mathbf{x}^{k}\|^{2}/\|\mathbf{x}^{k}\|^{2}\right)$) and iteration errors for FIMA with different settings. The legends ``$\mathbf{x}$'', ``$\mathbf{u}$'', and ``$\mathbf{u}$-$\mathbf{x}$'' respectively denote that at each iteration, we only perform classical PG (i.e., only the last step in Algs.~\ref{alg:eFIMA} and \ref{alg:iFIMA}), task-driven modules $\mathcal{A}$ (i.e., only perform Eq.~\eqref{eq:af_ag}), and their naive combination (without any scheduling policies). It can be seen that the function values and reconstruction errors of PG decrease slower than our FIMA strategies, while both ``$\mathbf{u}$''-curve (i.e., naive $\mathcal{A}_g\circ\mathcal{A}_f$) and ``$\mathbf{u}$-$\mathbf{x}$"-curve (i.e., $\mathcal{A}$ with PG refinement but no ``explicit momentum'' or ``error-control'' policy) have oscillations and could not converge after only 30 iterations. Moreover, we observe that adding PG to ``$\mathbf{u}$'' (i.e., ``$\mathbf{u}$-$\mathbf{x}$") make the curve worse rather than correct it to the descent direction. It illustrates that the pure adding strategies indeed break the convergence guarantee. In contrast, since of the choice mechanism in our algorithms, both eFIMA and iFIMA can provide a reliable variable ($\mathbf{v}^{k}$) in the current iteration to satisfy the convergence condition. We further explore the choice mechanism of FIMA in Fig.~\ref{fig:self-two-com}(d). The ``circles" in each curve represent the ``explicit momentum'' or ``error-control'' policy is satisfied, while the ``triangles" denote only perform PG in the current stage. It can be seen that the eFIMA strategy is more strict than iFIMA, the judgment policy fails only $20$ iterations in eFIMA while remains almost $40$ iterations in iFIMA. Both eFIMA and iFIMA have better performance than other compared schemes, thus verifies the efficiency of our proposed scheduling policies in Sec.~\ref{sec:fima}.

We also compare the iteration behaviors of FIMA to classical nonconvex APGs, including mAPG~\cite{li2015accelerated}, APGnc~\cite{li2017convergence}) and inexact niAPG~\cite{yao2016more} on the dataset collected by \cite{schmidt2014shrinkage}, which consists of 68 images corrupted by different blur kernels of the size ranging from 17$\times$17 to 37$\times$37. We add 1\textperthousand and 1\% Gaussian noise to generate our corrupted observations, respectively. In Fig.~\ref{fig:other-pgs}, the left four subfigures compare curves of iteration errors and PSNR on an example image and the rightmost one illustrate the averaged iteration numbers and run time on the whole dataset.
It can be seen that our eFIMA and iFIMA are faster and better than these abstractly designed classical solvers under the same iteration error ($\leq 1e-4$). Moreover, we observe that the performance of these nonconvex APGs is not satisfied when the noise level is bigger. The PSNRs of them (Fig.~\ref{fig:other-pgs}(d)) descent after dozens of steps, while our FIMA remains higher PSNR and fewer iterations. It illustrates that our strategy is more stable than traditional nonconvex APGs in image restoration because of the flexible modules and effective choice mechanisms.

\begin{figure*}[ht]
	\centering	\begin{tabular}{c@{\extracolsep{0.2em}}c@{\extracolsep{0.2em}}c@{\extracolsep{0.2em}}c@{\extracolsep{0.2em}}c}
		\includegraphics[width=0.18\textwidth]{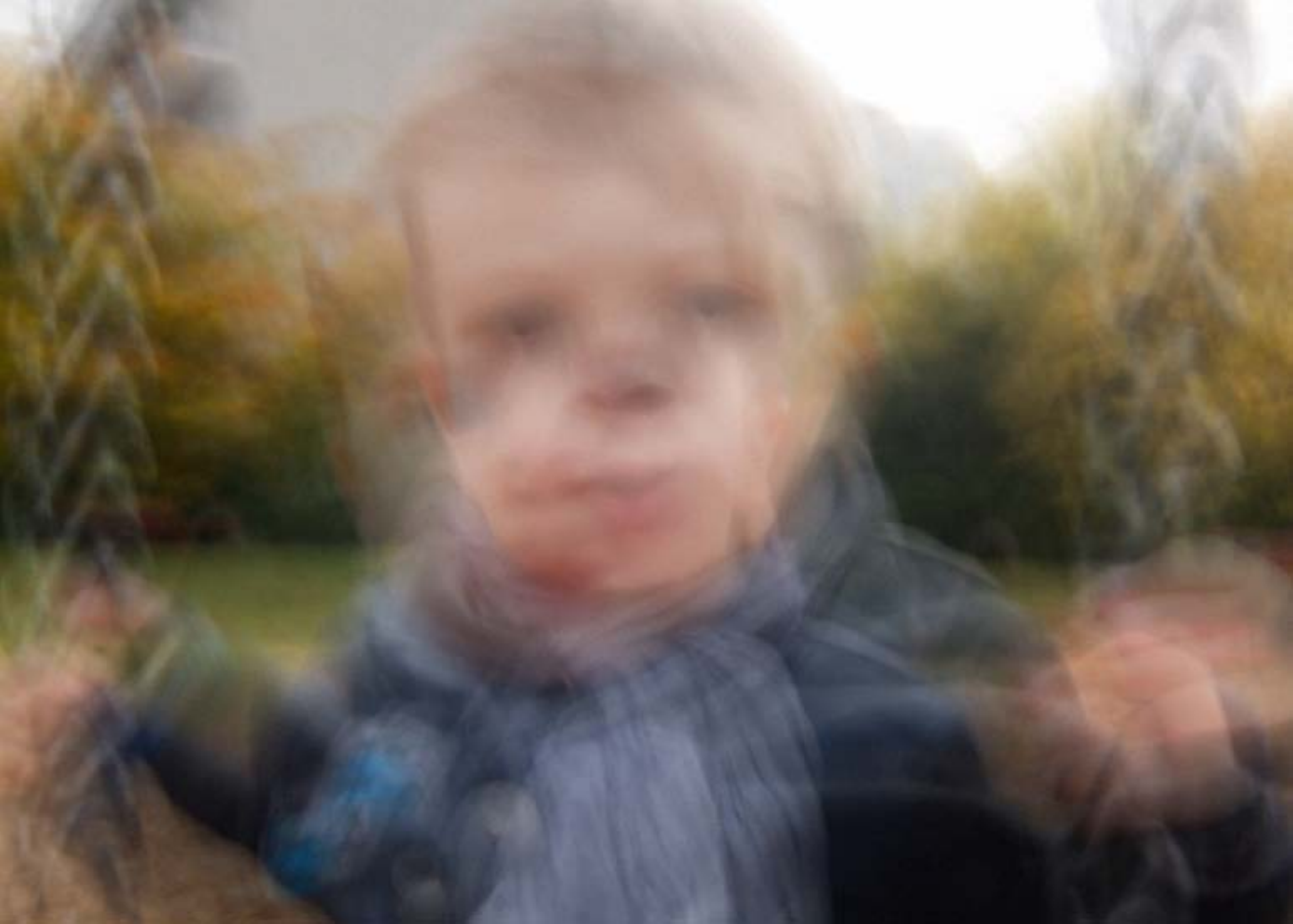}~
		&\includegraphics[width=0.18\textwidth]{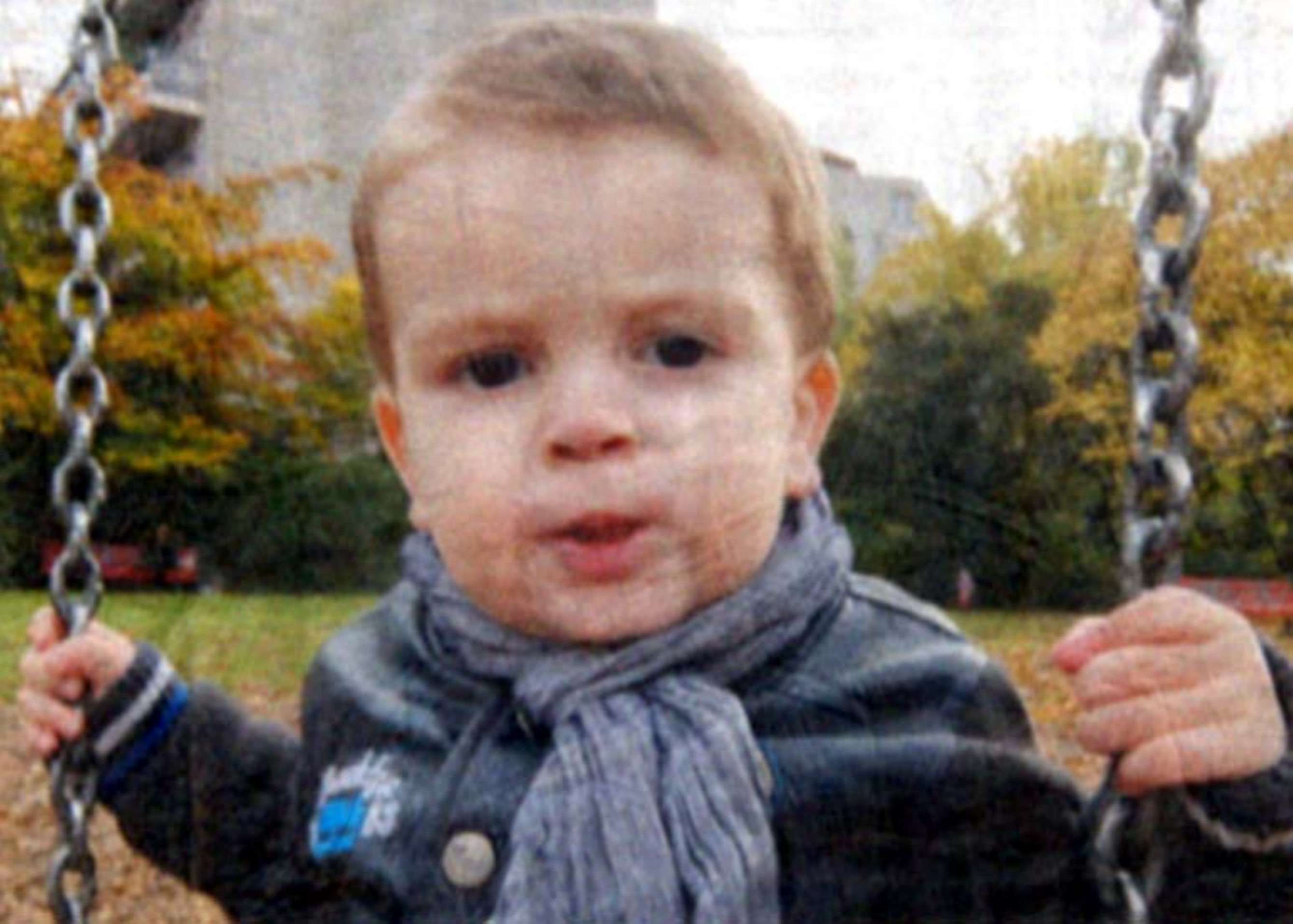}~
		&\includegraphics[width=0.18\textwidth]{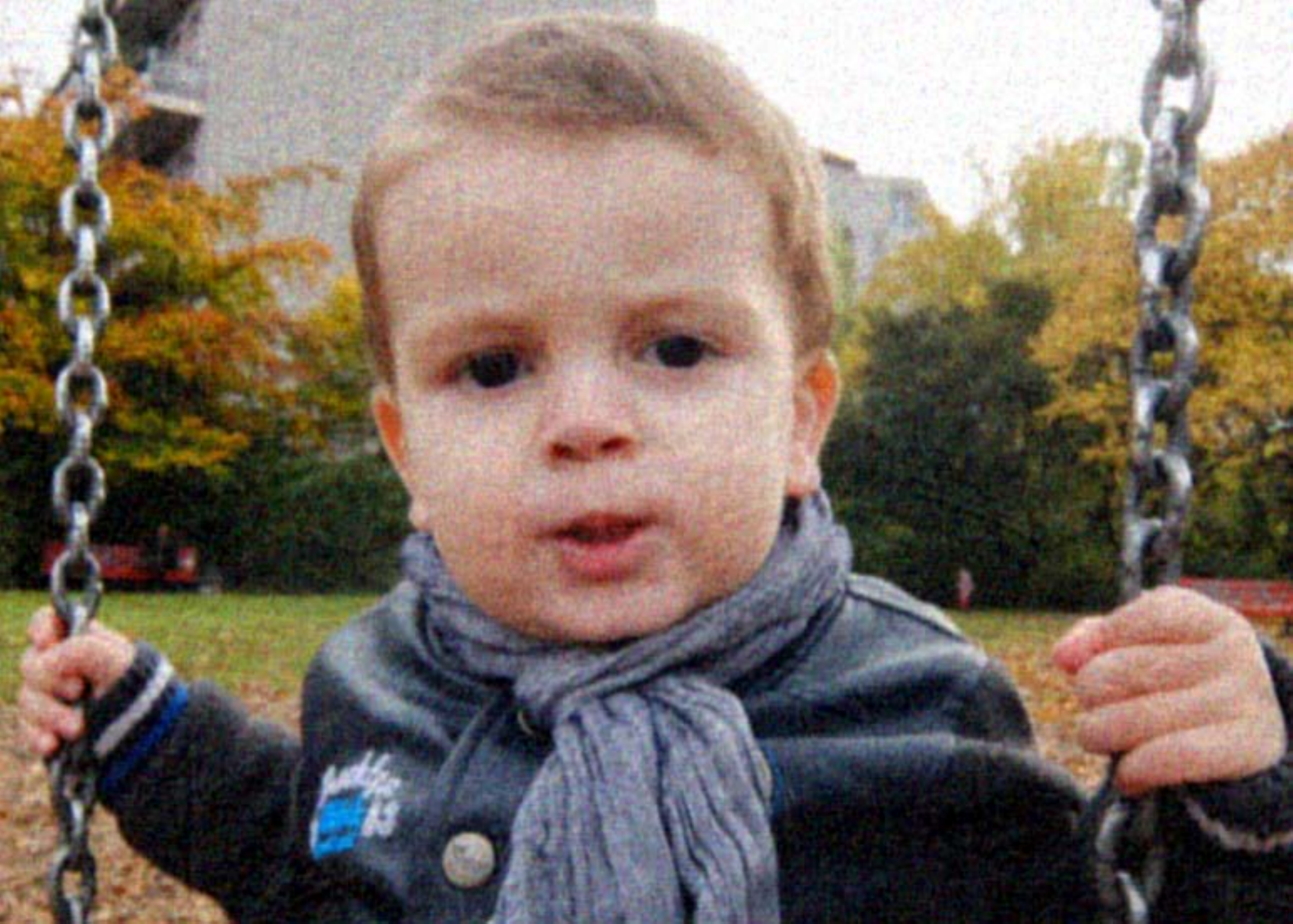}~
		&\includegraphics[width=0.18\textwidth]{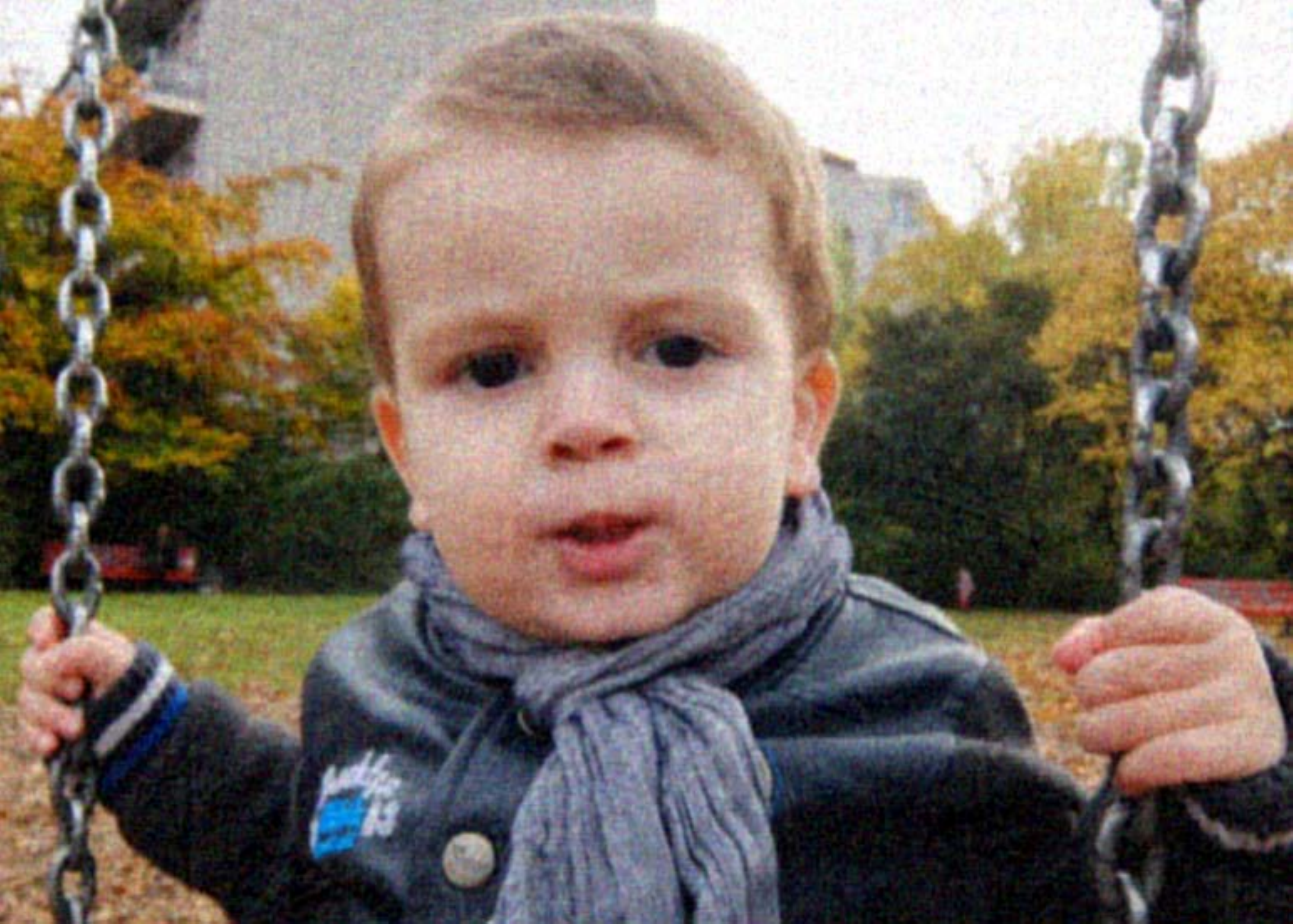}~
		&\includegraphics[width=0.18\textwidth]{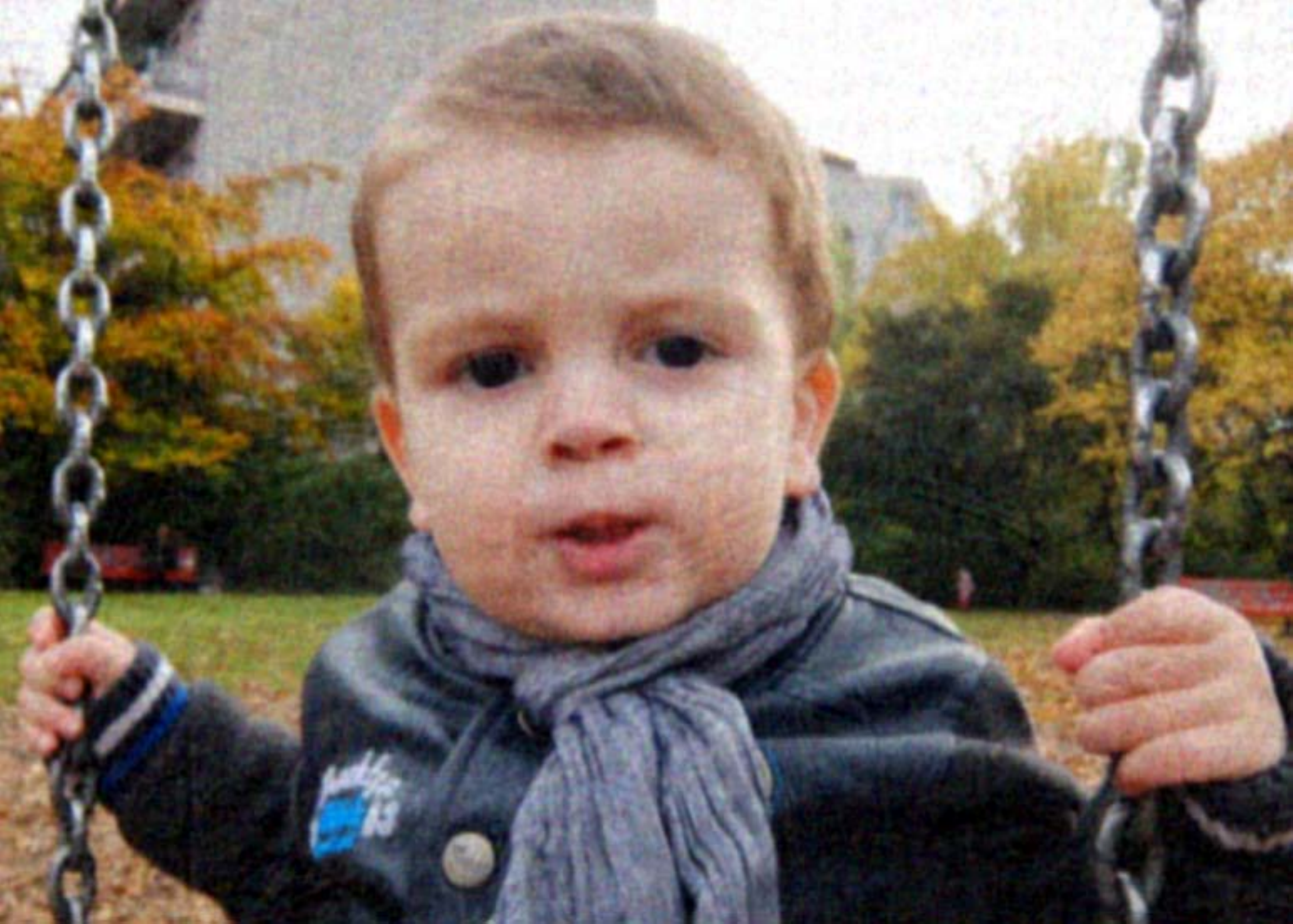}\\
		Input & PG & mAPG  & APGnc  & niAPG \\
		-&(24.97/0.79)&(25.67/0.73)&(25.68/0.73)&(26.17/0.78)\\
		\includegraphics[width=0.18\textwidth]{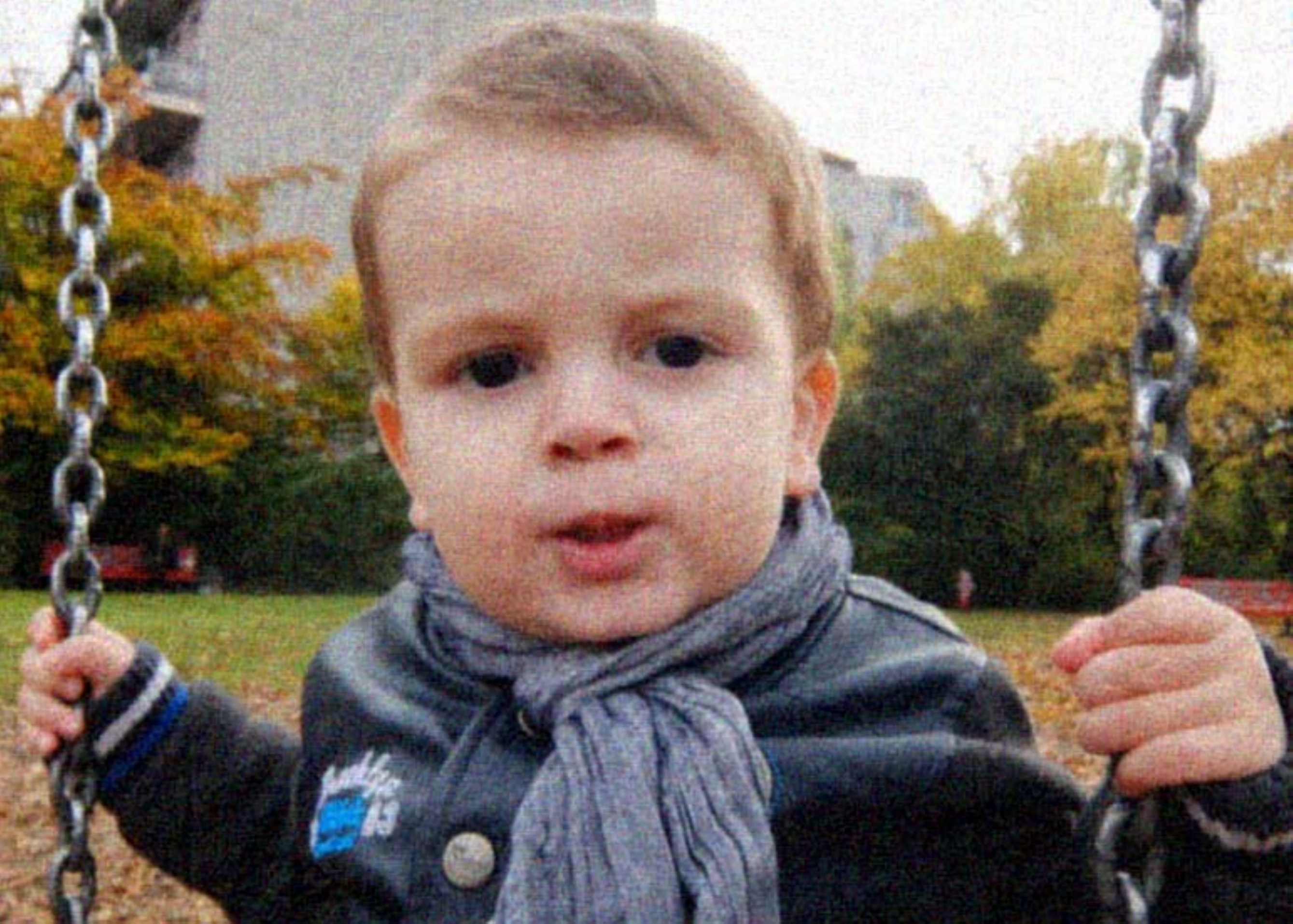}~
		&\includegraphics[width=0.18\textwidth]{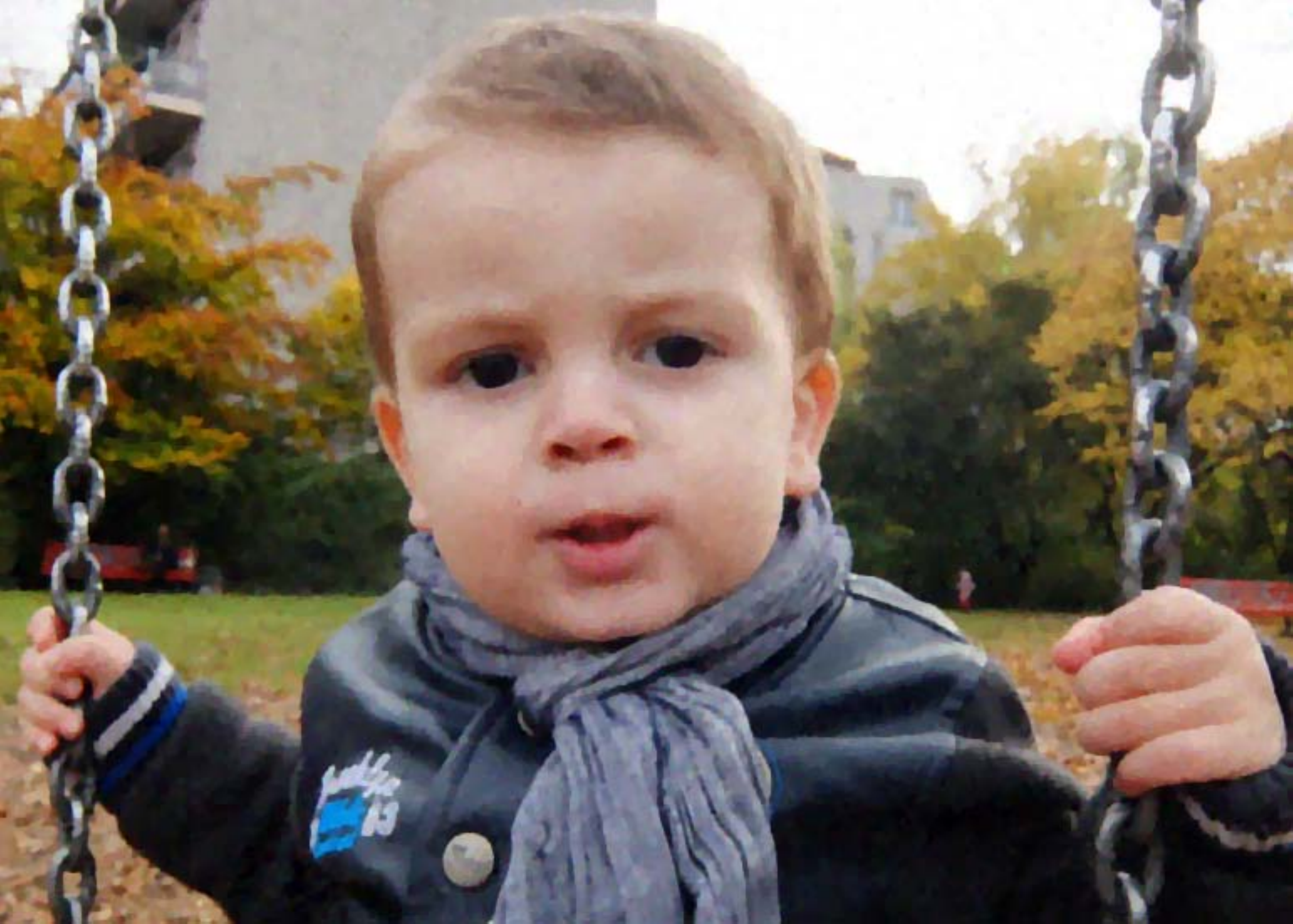}~
		&\includegraphics[width=0.18\textwidth]{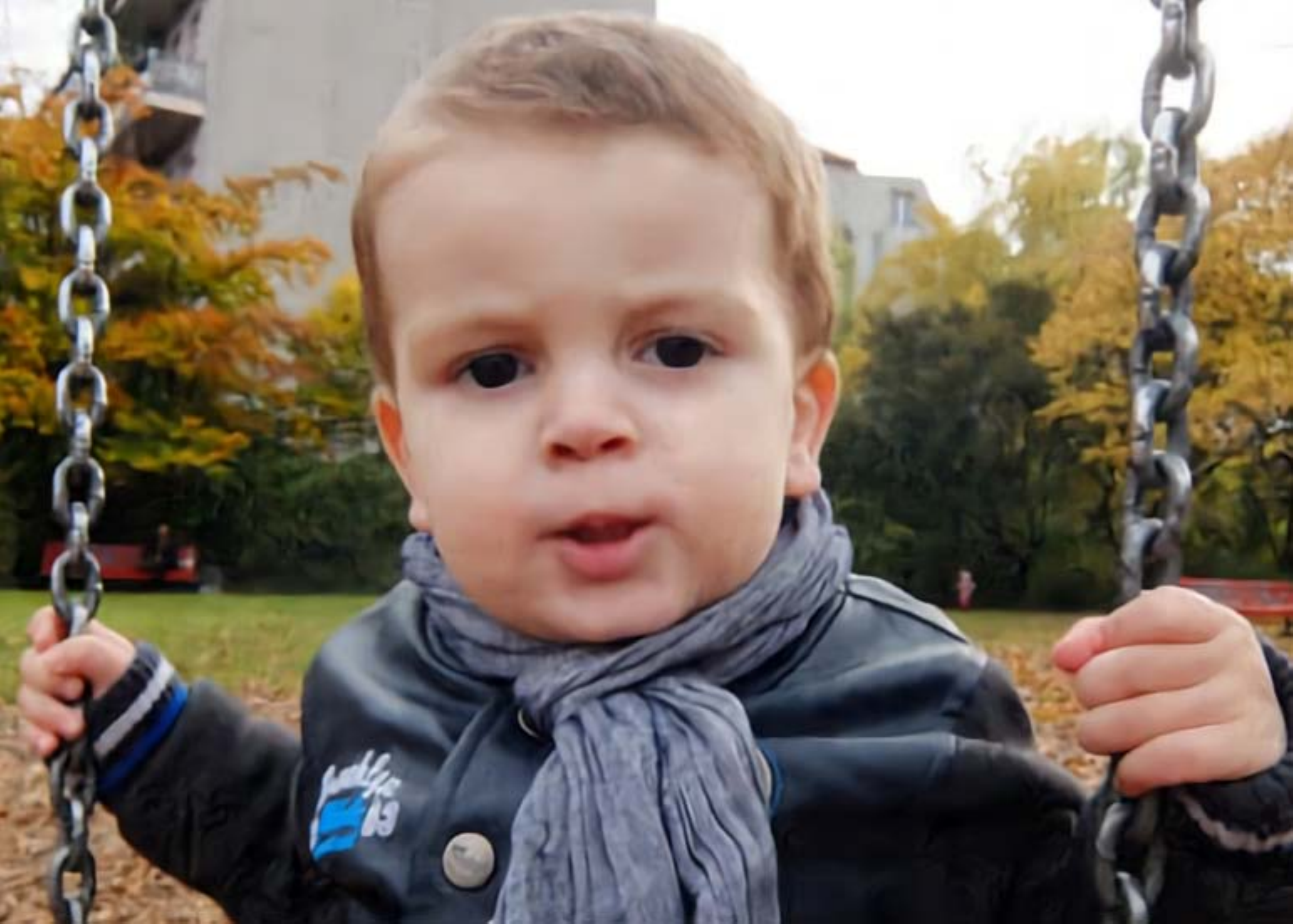}~
		&\includegraphics[width=0.18\textwidth]{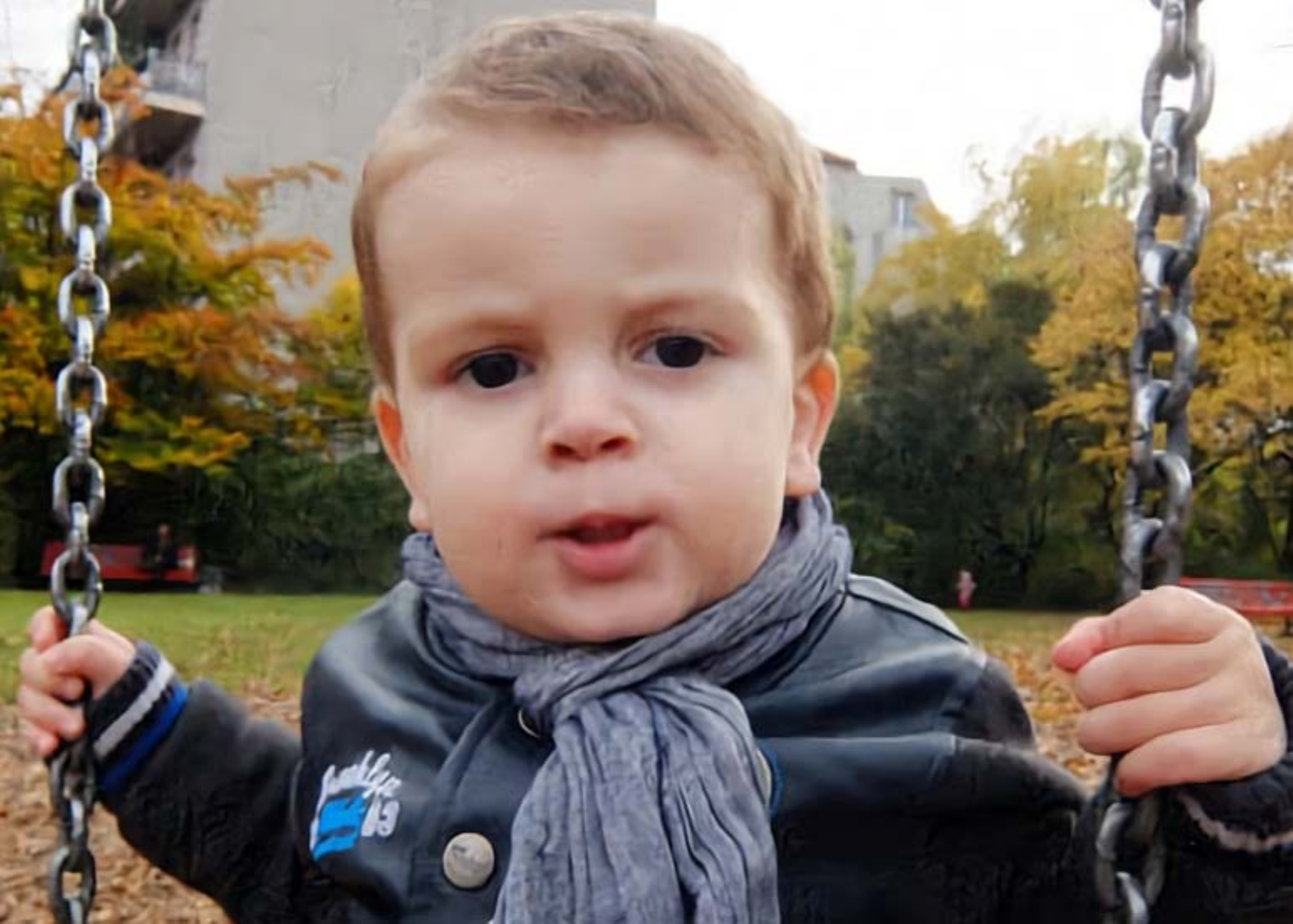}~
		&\includegraphics[width=0.18\textwidth]{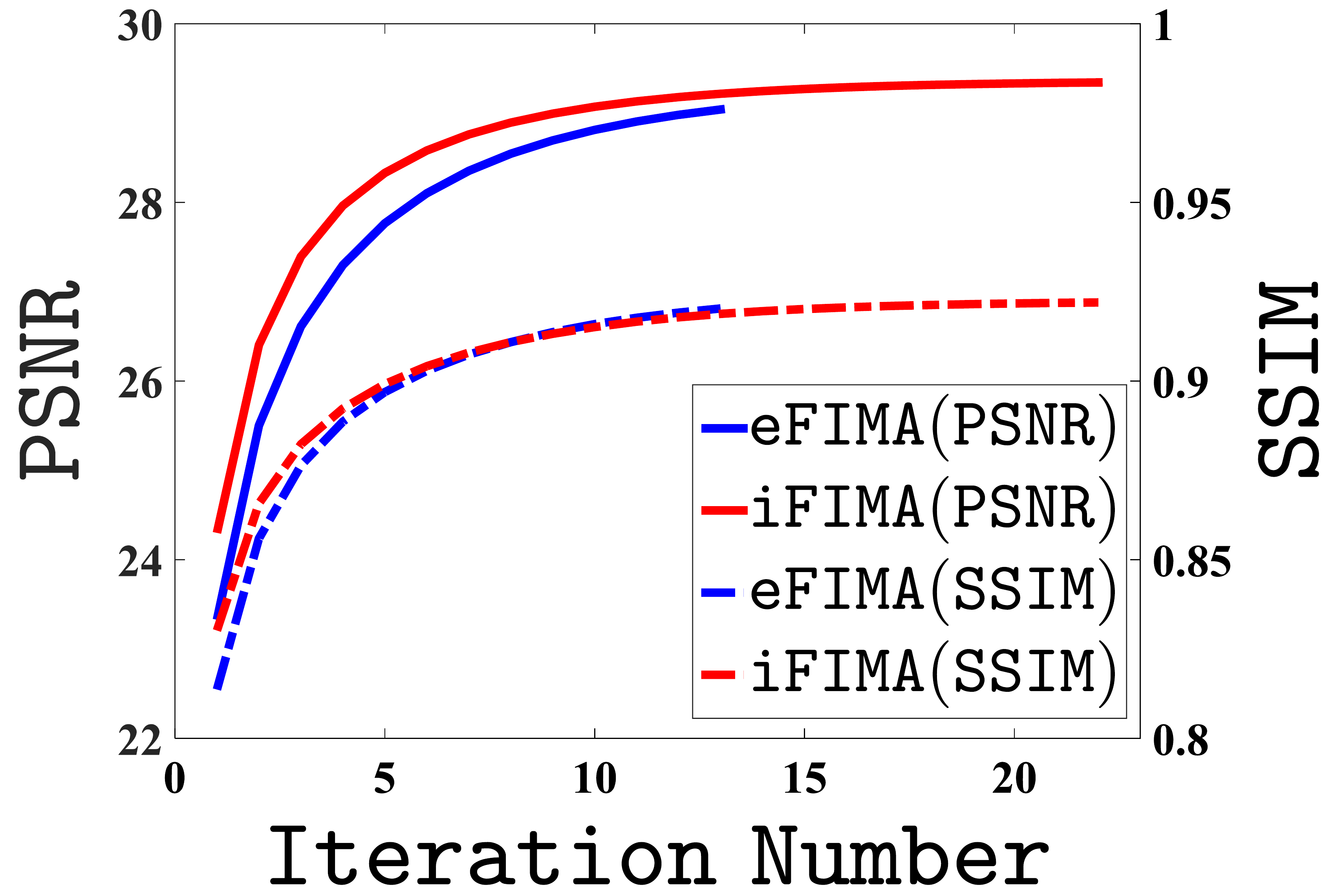}\\
		FISTA & FTVd &eFIMA&iFIMA & Curves of scores\\
		(25.03/0.68)&(27.75/0.88)&(29.04/\textbf{0.92}) &(\textbf{29.34}/\textbf{0.92})&\\
	\end{tabular}
	\caption{The non-blind deconvolution performances (1\% noise level) of eFIMA and iFIMA with comparisons to convex optimization based algorithms (i.e., FISTA and FTVd), and non-convex solvers (i.e., APGnc, mAPG, and niAPG). The quantitative scores (PSNR/SSIM) are reported below each image. The rightmost subfigure on the bottom row plots the curves of PSNR and SSIM of our methods. }
	\label{fig:pg-com}
\end{figure*}

In Fig.~\ref{fig:pg-com}, we illustrate the visual results of eFIMA and iFIMA with comparisons to both convex image restoration approaches, including FISTA \cite{beck2009a} (APG) and FTVd \cite{yang2009a}) (ADMM),
and nonconvex mAPG, APGnc, and niAPG on an example image with 1\% noise level but large kernel size (i.e, 75$\times$75)~\cite{Lai2016A}. Here FISTA and FTVd solve their original convex models, while mAPG, APGnc, and niAPG are based on the nonconvex model in Eq.~\eqref{eq:task-sparse-coding}. 
We have that APGs outperformed the original PG. The inexact niAPG is better than exact mAPG and APGnc. Since FTVd is specifically designed for this task, it is the best among all classical solvers, but worse than our FIMA. Overall, iFIMA obtain higher PSNR than eFIMA since the error-control mechanism actually tend to perform more accurate refinements.
\begin{figure*}[tb]
	\centering
	\begin{tabular}{c@{\extracolsep{0.2em}}c@{\extracolsep{0.2em}}c@{\extracolsep{0.2em}}c@{\extracolsep{0.2em}}c}
		\includegraphics[width=0.18\textwidth]{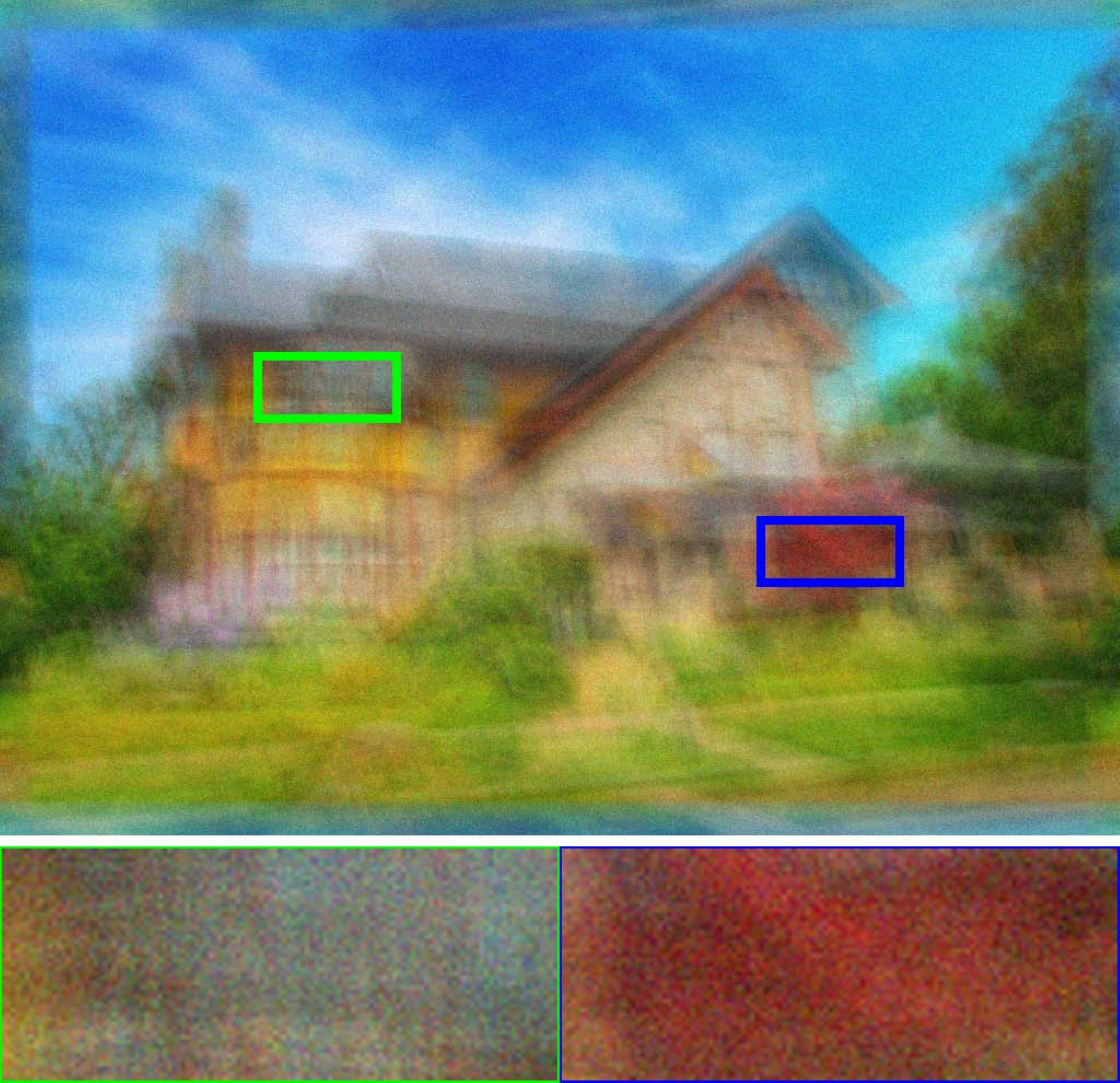}~
		&\includegraphics[width=0.18\textwidth]{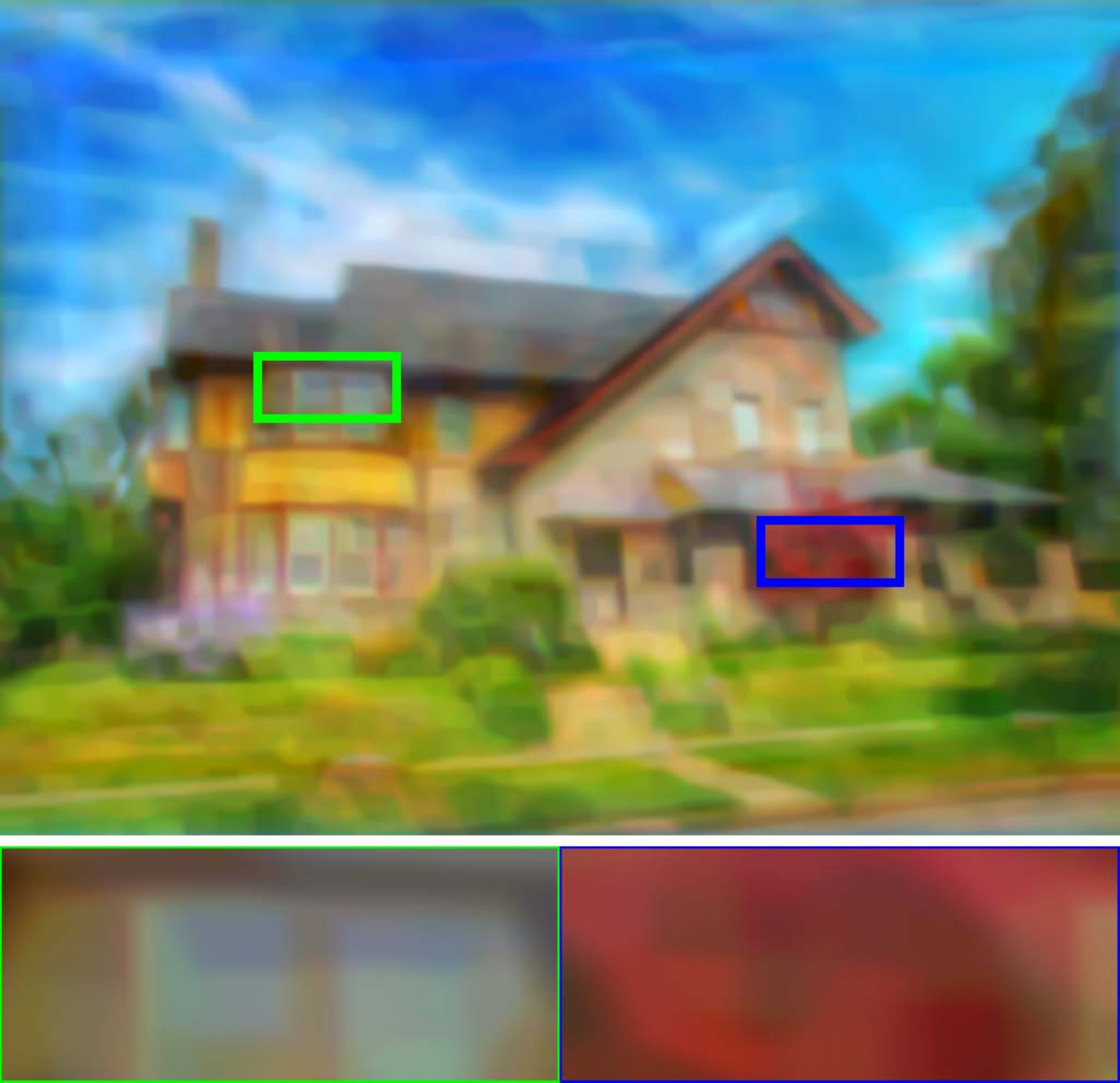}~
		&\includegraphics[width=0.18\textwidth]{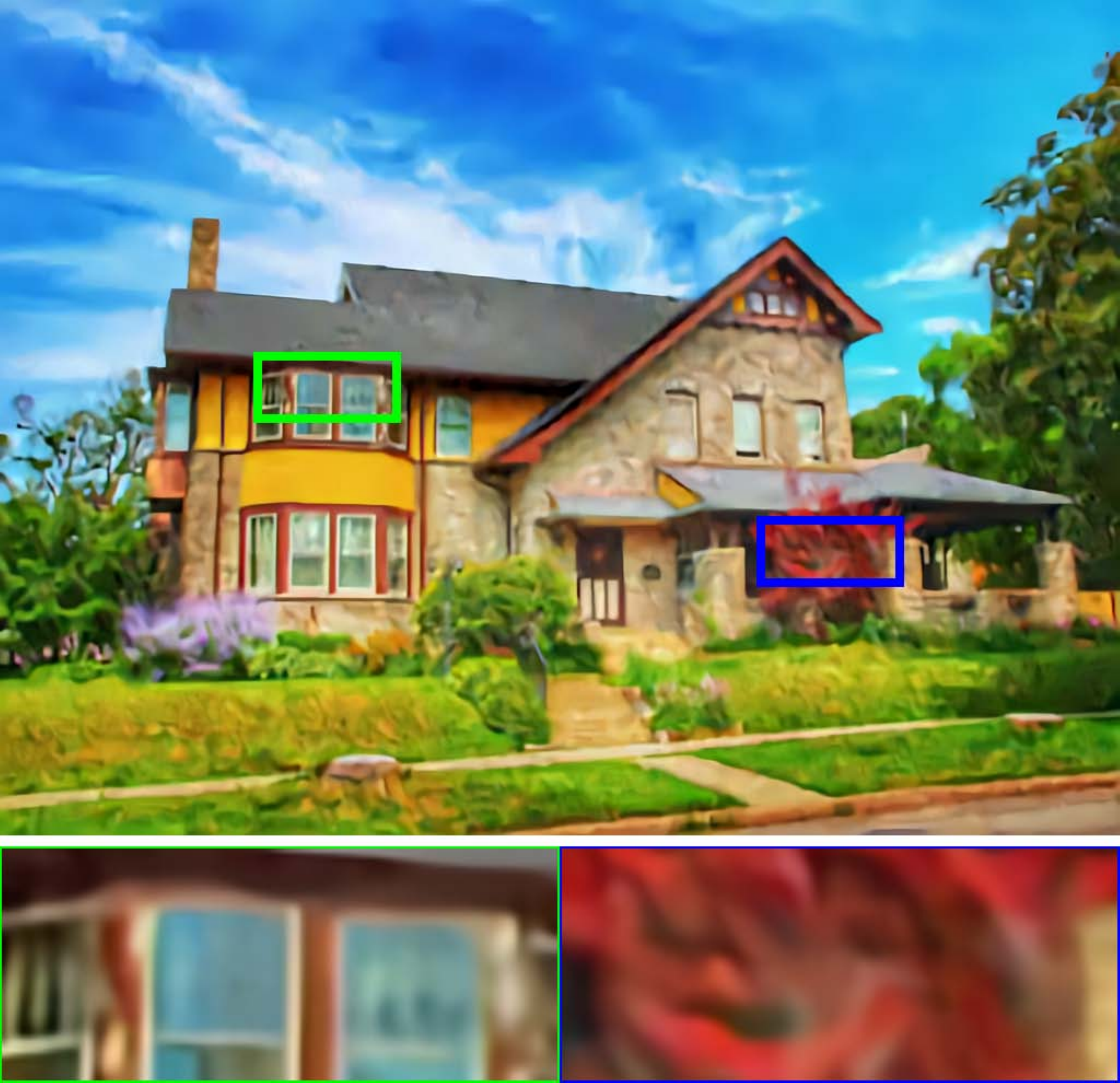}~
		&\includegraphics[width=0.18\textwidth]{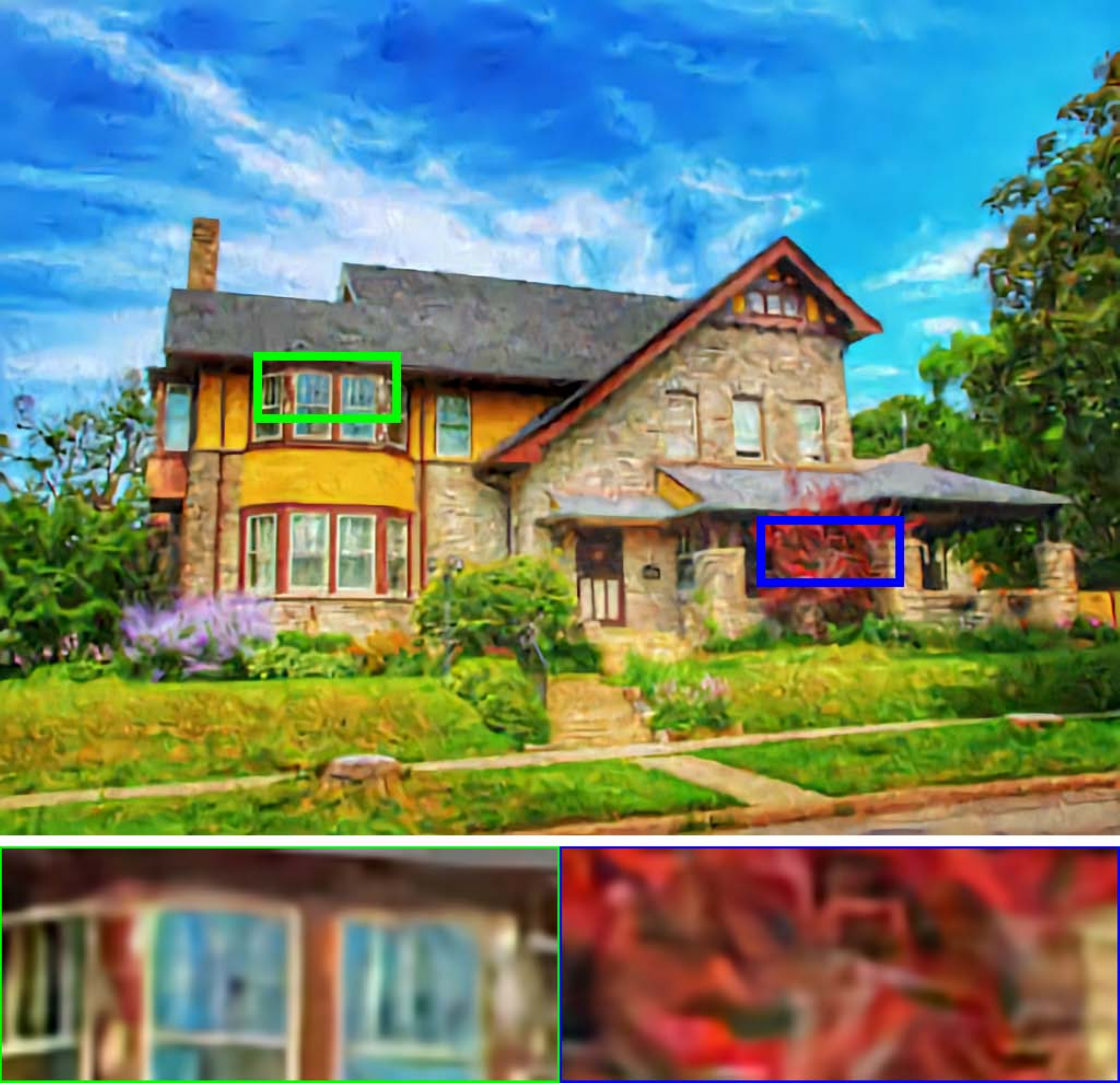}~
		&\includegraphics[width=0.18\textwidth]{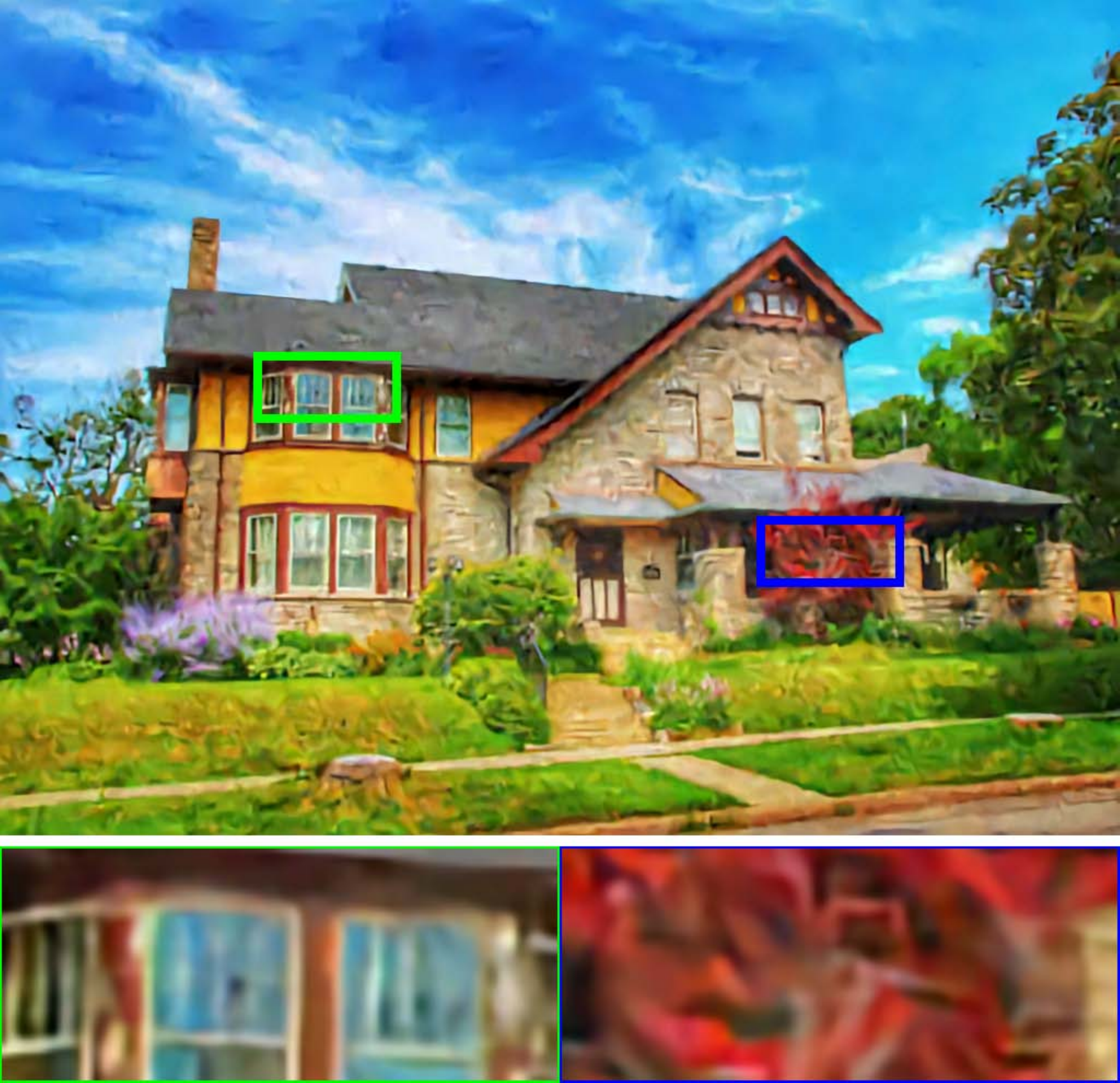}\\
		Input&PPADMM &IRCNN & eFIMA &iFIMA \\
		& (17.6 / 0.72)  & (20.96 / 0.82)  & (21.18 / \textbf{0.83})  & (\textbf{21.23} / \textbf{0.83})\\
	\end{tabular}
	\caption{The non-blind image deconvolution performance (5\% noise level) of FIMA with comparisons to existing plug-and-play type methods (i.e., PPADMM and IRCNN). The quantitative scores (PSNR/SSIM) are reported below each image.}
	\label{fig:nonblind-com}
\end{figure*}


\textbf{State-of-the-art Comparisons:}
We compare FIMA with state-of-the-art image restoration approaches, such as IDDBM3D~\cite{danielyan2012bm3d}, EPLL~\cite{zoran2011learning}, PPADMM \cite{chan2017plug}, RTF~\cite{schmidt2016cascades} and IRCNN~\cite{Zhang2017Learning}. Fig.~\ref{fig:nonblind-com} first compares our FIMA with two prevalent learning-based iterative approaches (i.e., PPADMM and IRCNN) on an example image with 5\% noise. Tab.~\ref{table:non-blind-dataSetResult} then reports the averaged quantitative results of all the compared methods on the image set (collected by \cite{schmidt2014shrinkage}) with different levels of Gaussian noise (i.e., 1\%, 2\%, 3\% and 4\%). We have that eFIMA and iFIMA not only outperform classical numerical solvers by a large margin in terms of speed and accuracy, but also achieve better performance than other state-of-the-art approaches. Within FIMA, it can be seen that the speed of eFIMA is faster, while PSNR and SSIM of iFIMA are relatively higher. This is mainly because the ``error control'' strategy tends to perform more refinements than the ``explicit momentum'' rule during iterations.
\begin{table*}[!htb]
	\centering
	\caption{Averaged PSNR, SSIM and Time(s) on the benchmark image set \cite{schmidt2014shrinkage}. Here $\sigma$ denotes the noise levels.}
	\vspace{0.5em}
	\small
	\begin{tabular}{|p{0.6cm}<{\centering}|p{0.9cm}<{\centering}|p{1.35cm}<{\centering}|p{0.75cm}<{\centering}|p{1.25cm}<{\centering}|p{0.75cm}<{\centering}|p{0.9cm}<{\centering}|p{0.7cm}<{\centering}|p{0.8cm}<{\centering}|p{0.8cm}<{\centering}|p{0.8cm}<{\centering}|p{0.8cm}<{\centering}|p{0.8cm}<{\centering}|}
		\hline
		\multirow{2}{*}{$\sigma$} &\multirow{2}{*}{Metric} &\multicolumn{5}{c|}{State-of-the-art Image Restoration Methods}& \multicolumn{4}{c|}{Classical Nonconvex  Methods}&\multicolumn{2}{c|}{Ours}\\ \cline{3-13}
		&  &  IDDBM3D &EPLL &PPADMM & RTF &IRCNN& PG &mAPG &APGnc &niAPG &eFIMA&iFIMA \\\hline\hline
		\multirow{3}{*}{1\%} &PSNR & 28.83 & 28.67 &28.01 &  29.12 & 29.78 & 27.32 &26.68&26.69 &27.24&29.81&\textbf{29.85}\\\cline{2-13}
		&SSIM    &0.81 & 0.81 & 0.78 &0.83&0.84&0.71&0.67&0.67&0.73&\textbf{0.85}&\textbf{0.85}\\\cline{2-13}
		&Time(s)    & 193.13 & 112.03 & 293.99 & 249.83 &2.67&20.36&13.02&7.16&5.29&\textbf{1.89}&2.06\\\hline\hline
		\multirow{3}{*}{2\%} &PSNR    & 27.60 & 26.79 & 26.54 & 25.58 & 27.90 &25.61 & 25.20  &25.28   & 25.63 &28.02&\textbf{28.06}\\\cline{2-13}
		&SSIM    & 0.76 & 0.74 & 0.72  &0.66&0.78&0.63&0.60&0.61& 0.64 &\textbf{0.79}&\textbf{0.79}\\\cline{2-13}
		&Time(s)    &  198.66 & 100.52 & 270.45 &254.26&2.68&15.43&7.70&4.66&3.30&\textbf{1.90}&2.07\\\hline\hline
		\multirow{3}{*}{3\%} &PSNR    & 26.72 & 25.68  & 25.78 & 21.18 & 26.81   & 24.63 &24.39&24.48  &24.76 &27.05&\textbf{27.07}\\\cline{2-13}
		&SSIM    & 0.72 & 0.69 & 0.68 &0.42&0.73&0.57&0.55&0.56&0.61&0.74&\textbf{0.75}\\\cline{2-13}
		&Time(s)    & 191.25   & 96.32 & 257.94 &252.47&2.68&13.89&6.44&5.37&2.63&\textbf{1.89}&2.07\\\hline\hline
		\multirow{3}{*}{4\%} &PSNR    &  26.06  &  24.88  & 25.27 & 17.95 &26.10&24.05    &  23.88  &23.95    &24.14 &26.20&\textbf{26.37}\\\cline{2-13}
		&SSIM    &0.69 & 0.65  &  0.66  &0.28&0.70&0.54&0.53&0.53&0.59&0.70&\textbf{0.72}\\\cline{2-13}
		&Time(s)    &  183.44 & 93.82 & 258.45 &255.84&2.67&11.99&6.01&7.82&2.35&\textbf{1.89}&2.07\\\hline		
	\end{tabular}
	\label{table:non-blind-dataSetResult}
\end{table*}
\subsection{Blind Image Deconvolution}

Blind deconvolution is known as one of the most challenging low-level vision tasks. Here we evaluate miFIAM on solving Eq.~\eqref{eq:task-map} to address this fundamentally ill-posed
multi-variables inverse problem. We adopt the same CNN module $\mathcal{A}_g^{\mathtt{CNN}}$ as that in Sec.~\ref{sec:imresto} but train it on image gradient domain to enhance its ability for sharp edge detection.

\begin{figure}[t]
	\centering
	
	\begin{minipage}{0.15\textwidth}
		\subfigure{
			\begin{minipage}{1\textwidth}
				\includegraphics[width=1\textwidth]{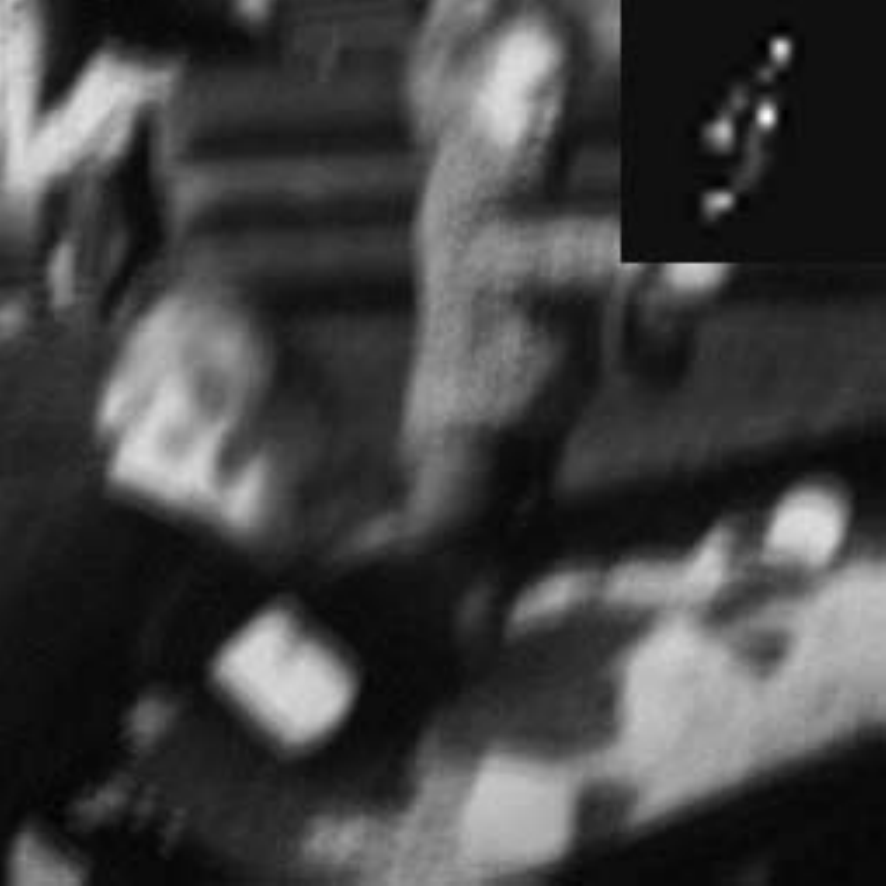}
				\centering  \footnotesize Input \vspace{-0.5ex}\\
			\end{minipage}
		}
	\end{minipage}~
	\begin{minipage}{0.15\textwidth}
		\subfigure{
			\begin{minipage}{1\textwidth}
				\includegraphics[width=1\textwidth]{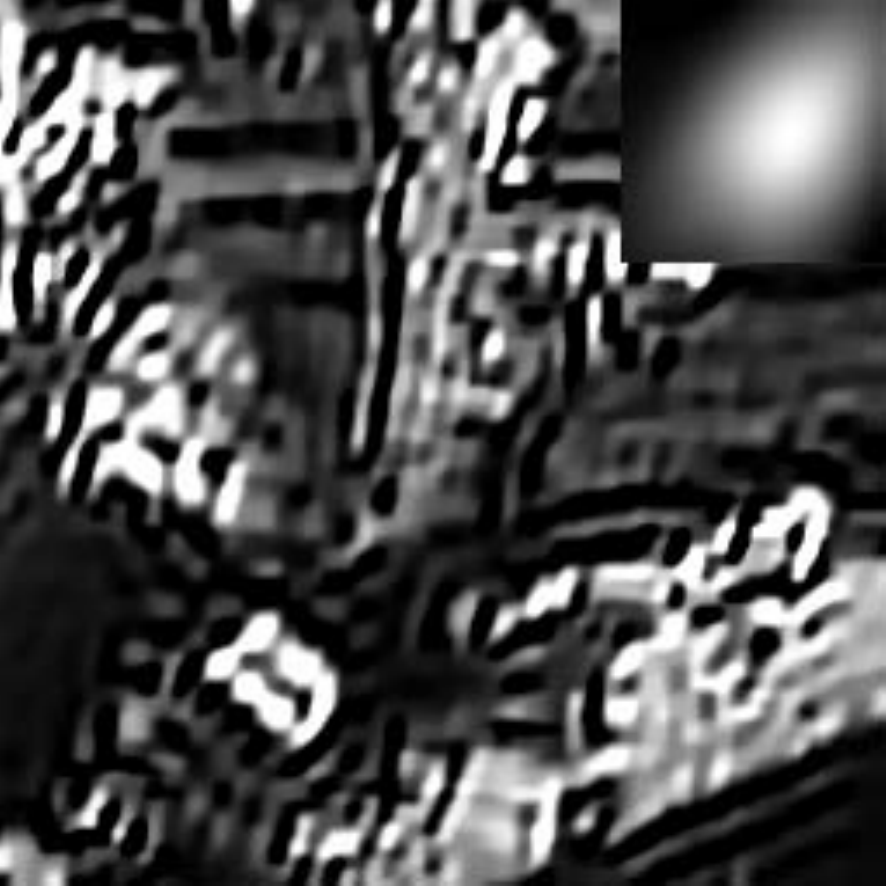}\\
				\centering  \footnotesize mFIMA without~$\mathcal{A}$\\
			\end{minipage}
		}
	\end{minipage}~
	\begin{minipage}{0.15\textwidth}
		\subfigure{
			\begin{minipage}{1\textwidth}
				\includegraphics[width=1\textwidth]{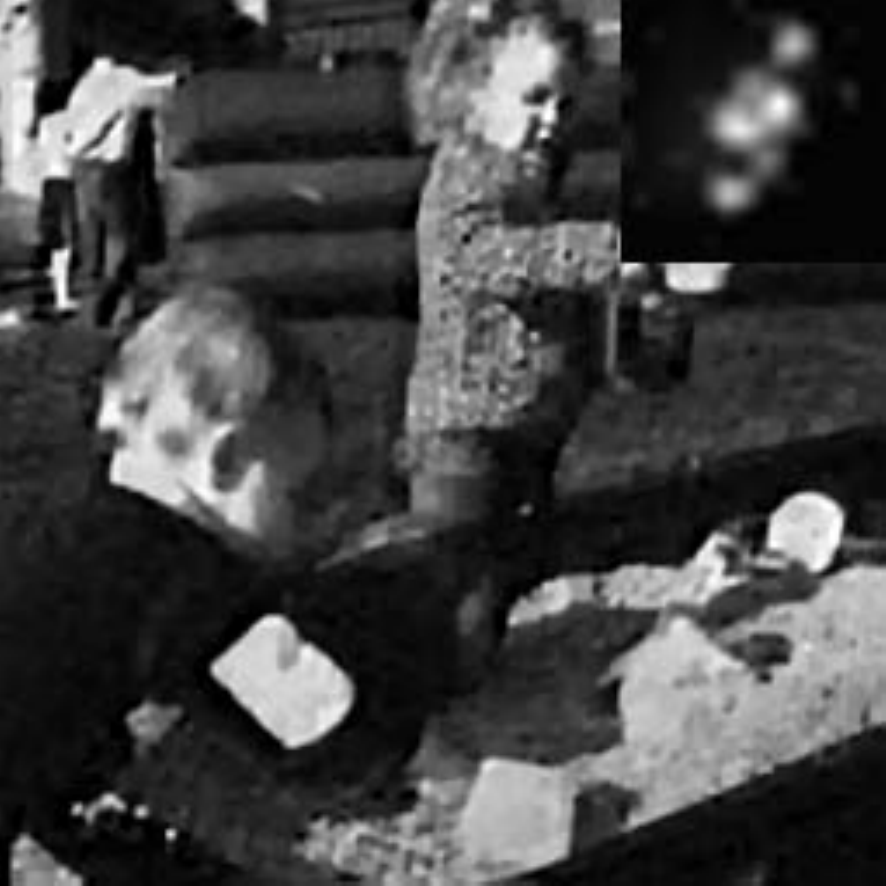}
				\centering \footnotesize  mFIMA with~$\mathcal{A}$\\
			\end{minipage}
		}
	\end{minipage} 
	\begin{minipage}{0.22\textwidth}
		\subfigure{
			\begin{minipage}{1\textwidth}
				\includegraphics[width=1\textwidth]{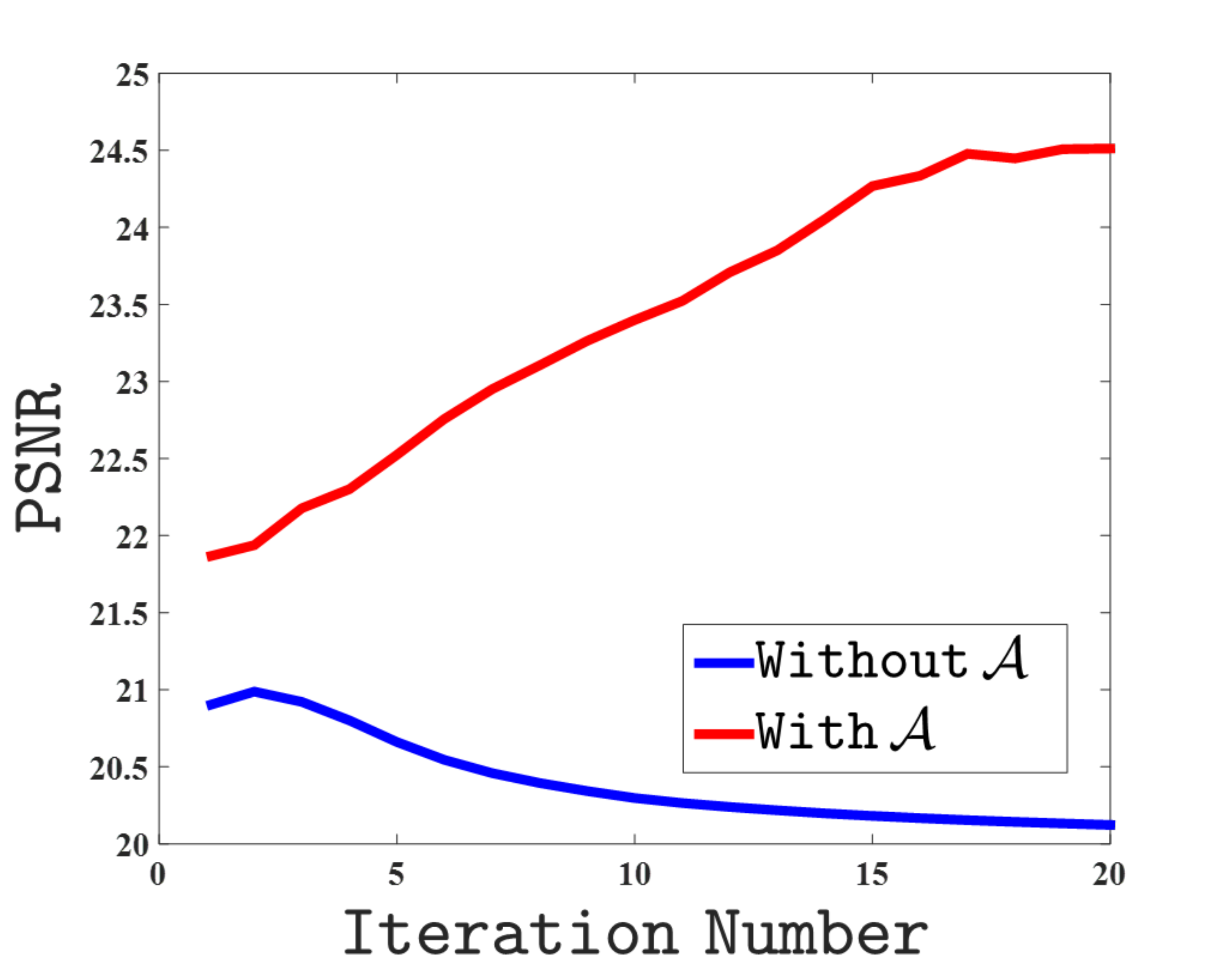}\\ 
			\end{minipage}
		}
	\end{minipage}
	\begin{minipage}{0.22\textwidth}
		\subfigure{
			\begin{minipage}{1\textwidth}
				\includegraphics[width=1\textwidth]{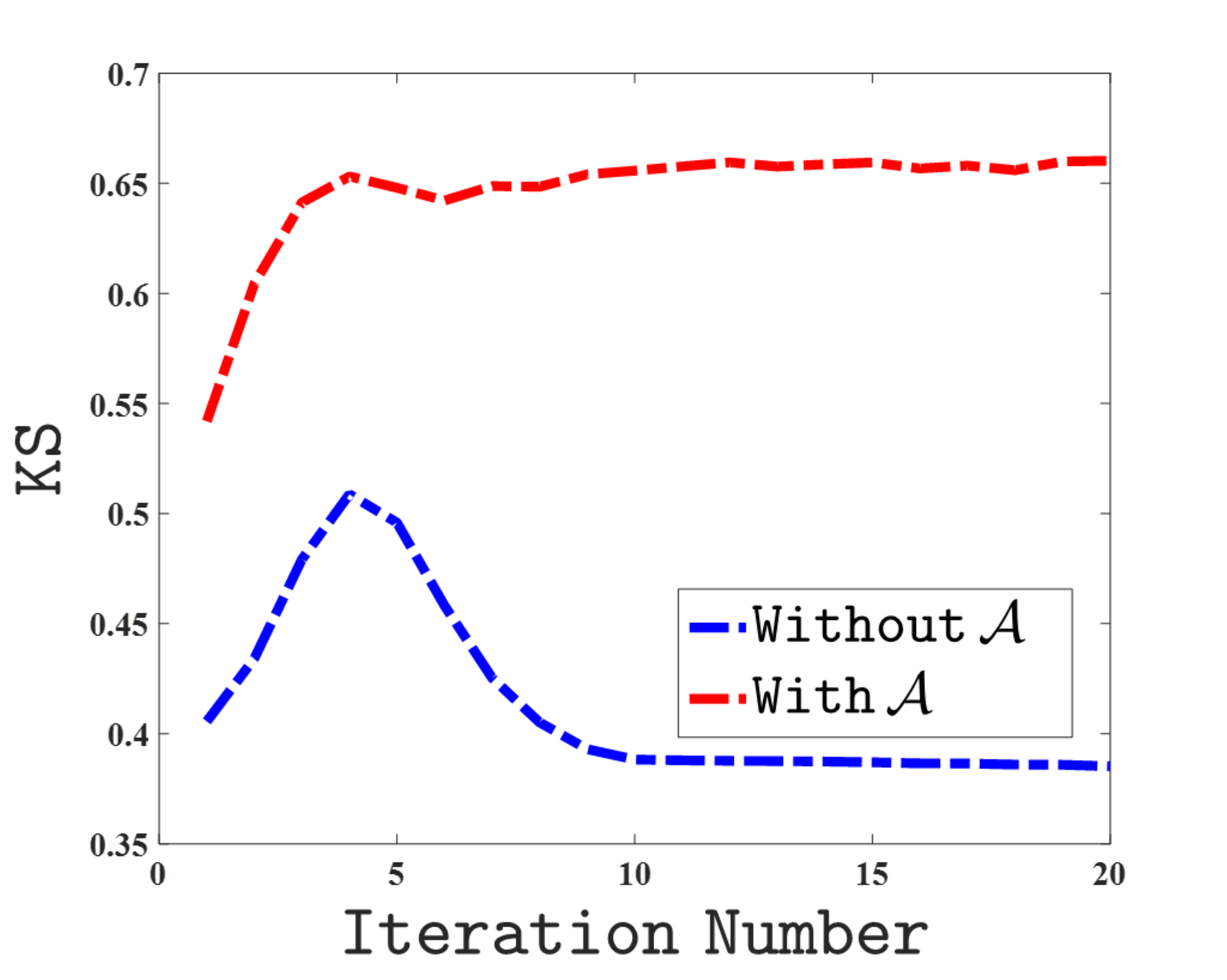}\\
			\end{minipage}
		}
	\end{minipage}
	\caption{The comparisons of mFIMA with and without the module $\mathcal{A}$. The top row compares the visual results of these different strategies. The bottom row plots the curves of PSNR and KS scores during iterations.}
	\label{fig:blind-loss}
\end{figure}

In Fig.~\ref{fig:blind-loss}, we show the visual performances of mFIMA in different settings (i.e., with and without $\mathcal{A}$) on an example blurry image from \cite{levin2009understanding}. 
We observe that mFIMA without $\mathcal{A}$ almost failed on this experiment. This is not surprising since \cite{levin2009understanding,patchdeblur_iccp2013} have proved that standard optimization strategy is likely to lead to degenerate global solutions like the delta kernel (frequently called the no-blur solution), or many suboptimal local minima. In contrast, the CNN-based modules successful avoid trivial results and significantly improve the deconvolution performance. We also plot the curves of quantitative scores (i.e., PSNR for the latent image and Kernel Similarity (KS) for the blur kernel) on the bottom row for these two strategies on the bottom row. As these scores are stable after 20 iterations, here we only plot curves of the first 20 iterations.
\begin{table}[t]
	\centering
	\caption{Averaged quantitative scores on Levin et al's benchmark.}
	\vspace{0.5em}
	\small
	\begin{tabular}{|c|c|c|c|c|c|}
		\hline
		Method  & PSNR &SSIM &ER& KS &Time(s)   \\\hline\hline
		Perrone et al.     &  29.27  &0.88 &1.35&0.80&113.70\\ \hline
		Levin et al.  & 29.03  & 0.89 &1.40   &0.81&41.77\\ \hline
		Sun et al.     &29.71&   0.90  &1.32 &0.82&209.47\\ \hline
		Zhang et al.   &28.01 &   0.86 &1.25  &0.58&37.45\\ \hline
		Pan et al.     &29.78  & 0.89& 1.33&0.80&102.60 \\ \hline
		Ours&          \textbf{30.37}&\textbf{0.91}& \textbf{1.20}& \textbf{0.83} &\textbf{5.65}\\ \hline
	\end{tabular}
	\label{table:blind-dataSetResult}
\end{table}

\begin{figure*}[tb]
	\centering
	\begin{tabular}{c@{\extracolsep{0.2em}}c@{\extracolsep{0.2em}}c@{\extracolsep{0.2em}}c@{\extracolsep{0.2em}}c}
		\includegraphics[width=0.18\textwidth]{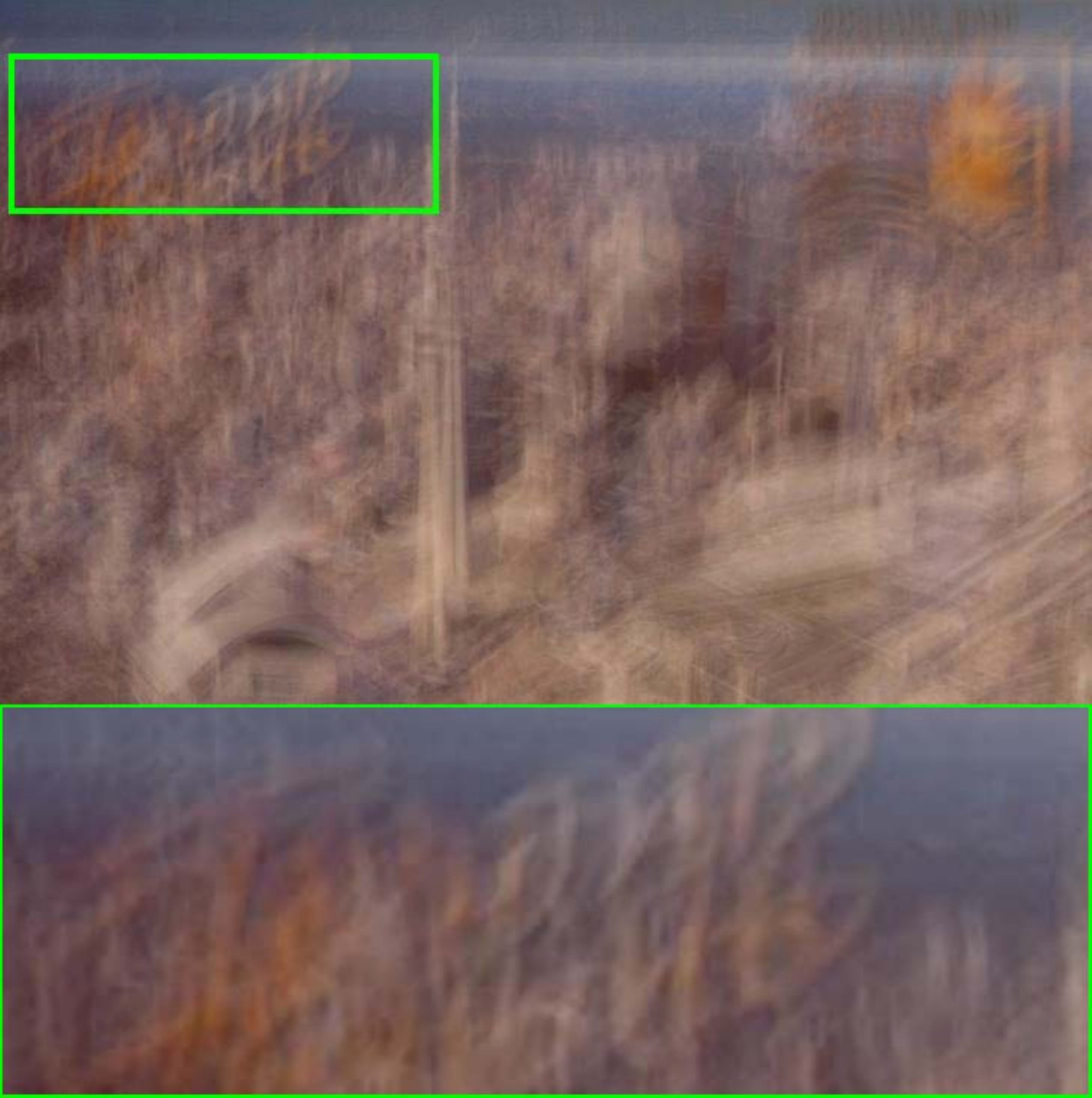}~
		&\includegraphics[width=0.18\textwidth]{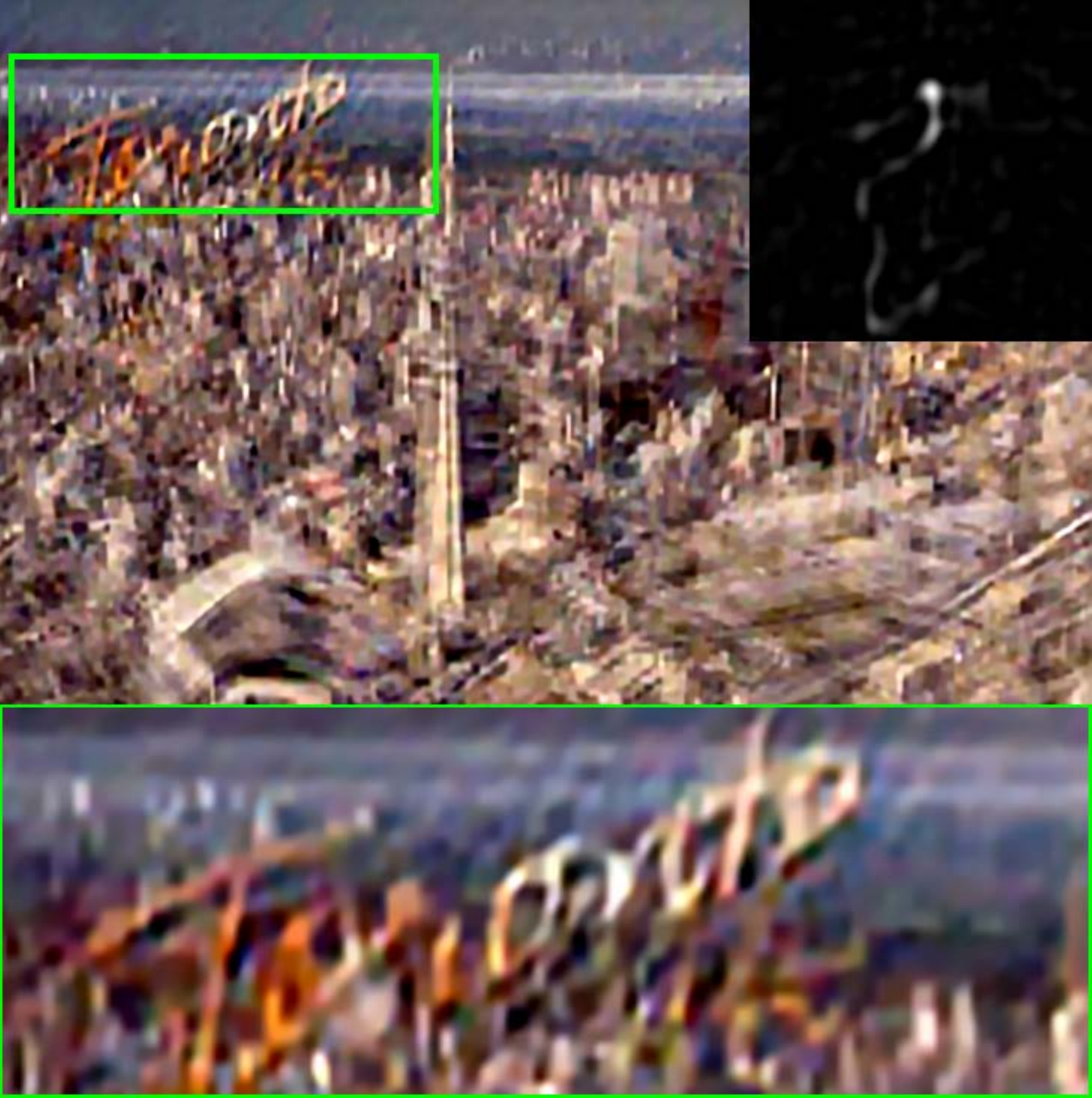}~
		&\includegraphics[width=0.18\textwidth]{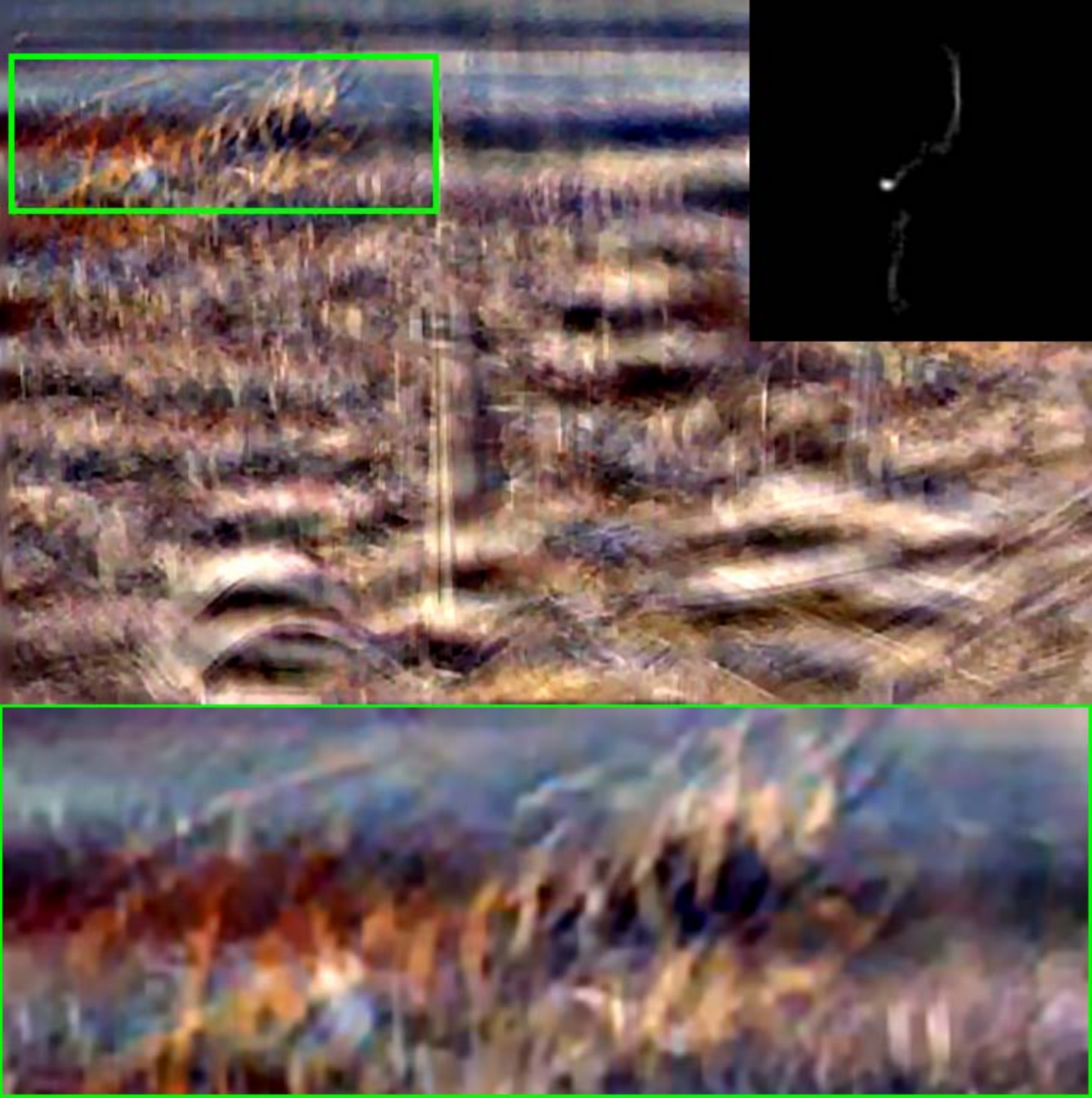}~
		&\includegraphics[width=0.18\textwidth]{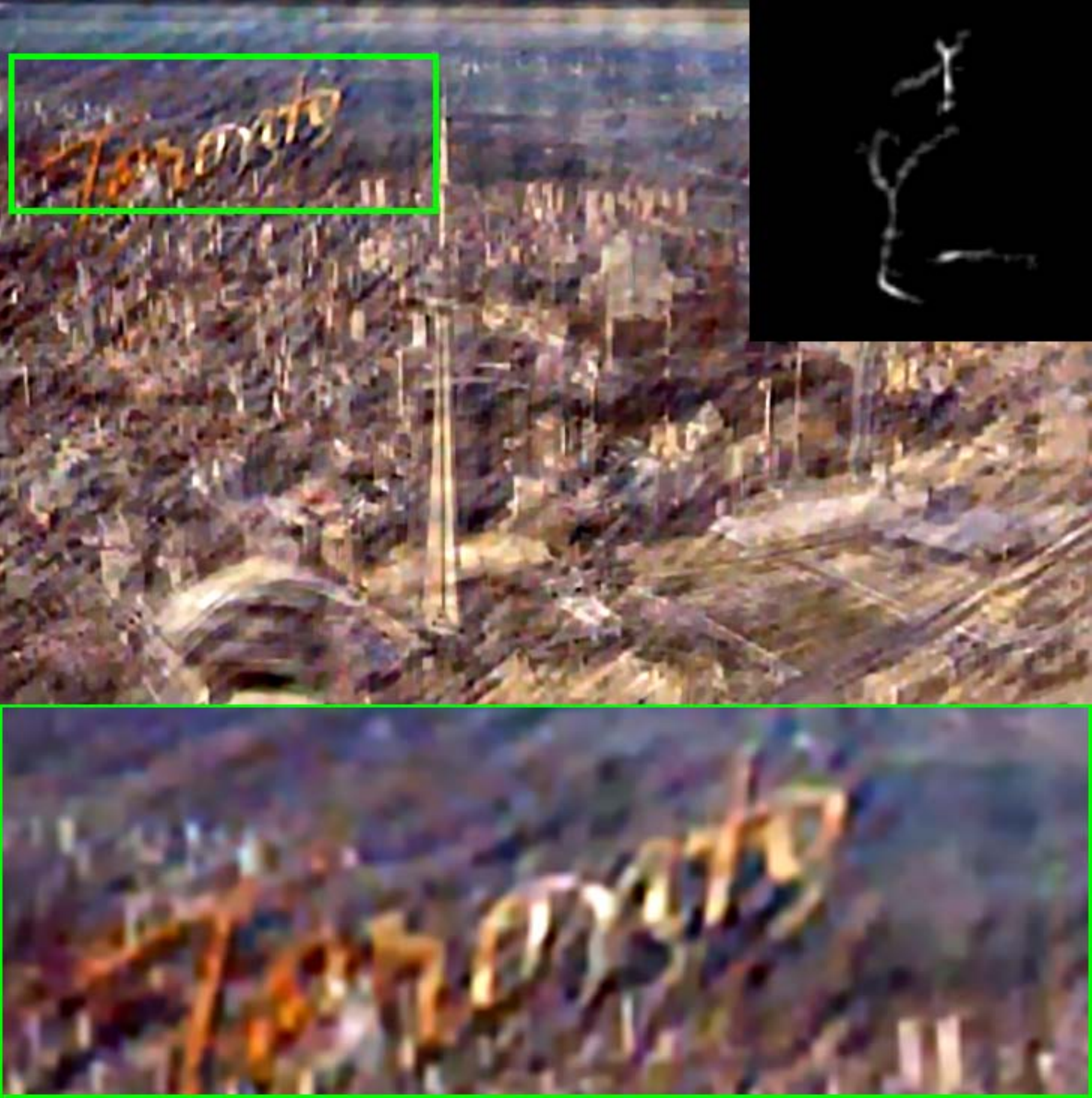}~
		&\includegraphics[width=0.18\textwidth]{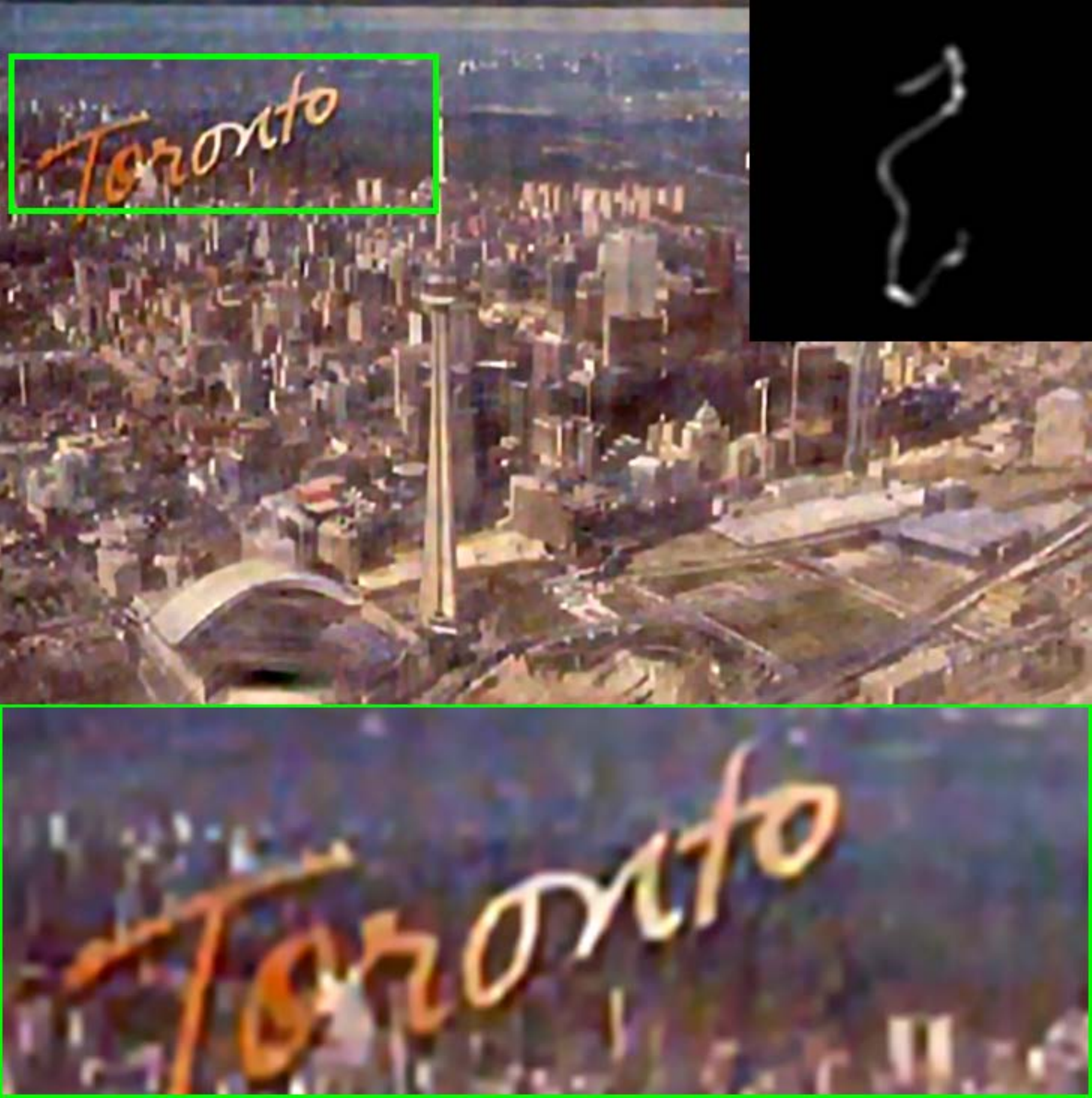}\\
		Input&Perrone et al.& Sun et al. & Pan et al. & Ours\\
	\end{tabular}
	\caption{Visual comparisons between mFIMA and other competitive methods (top 3 in Tab.~\ref{table:blind-dataSetResult}) on a real blurry image.}
	\label{fig:blind-com}
\end{figure*}

We then compare mFIMA with state-of-the-art deblurring methods\footnote{In this and the following experiments, the widely used multi-scale techniques are adopted for all the compared methods.}, such as Perrone et al.~\cite{perrone2014total}, Levin et al.~\cite{levin2009understanding}, Sun et al.~\cite{patchdeblur_iccp2013}, Zhang et al.~\cite{zhang2013multi} and Pan et al.~\cite{Pan2017Deblurring} on the most widely-used Levin et al's benchmark~\cite{levin2009understanding}, which consists of 32 blurred images generated by 4 clean images and 8 blur kernels.  Tab.~\ref{table:blind-dataSetResult} reports the averaged quantitative scores, including PSNR, SSIM, and Error Rate (ER) for the latent image, Kernel Similarity (KS) for the blur kernel and the overall run time.
Fig.~\ref{fig:blind-com} further compares the visual performance of mFIMA to Perrone et al., Sun et al. and Pan et al. (i.e., top 3 in Tab.~\ref{table:blind-dataSetResult}) on a real-world challenging blurry image collected by \cite{Lai2016A}.
It can be seen that mFIMA consistently outperforms all the compared methods both quantitatively and qualitatively, which verifies the efficiency of our proposed learning-based iteration methodology. 

\begin{figure*}[tb]
	\centering
	\begin{tabular}{c@{\extracolsep{0.2em}}c@{\extracolsep{0.2em}}c@{\extracolsep{0.2em}}c@{\extracolsep{0.2em}}c}
		\includegraphics[width=0.18\textwidth]{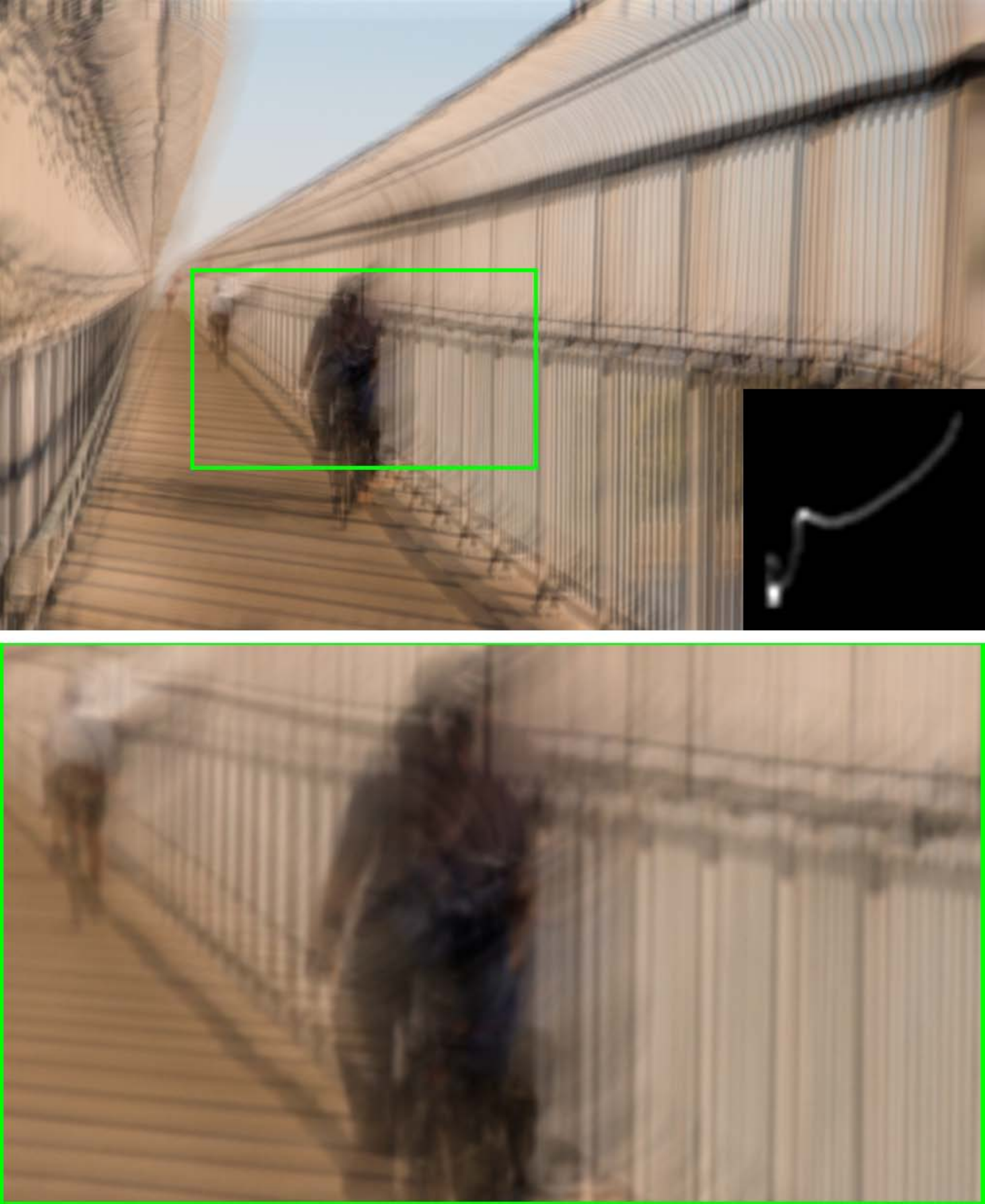}~
		&\includegraphics[width=0.18\textwidth]{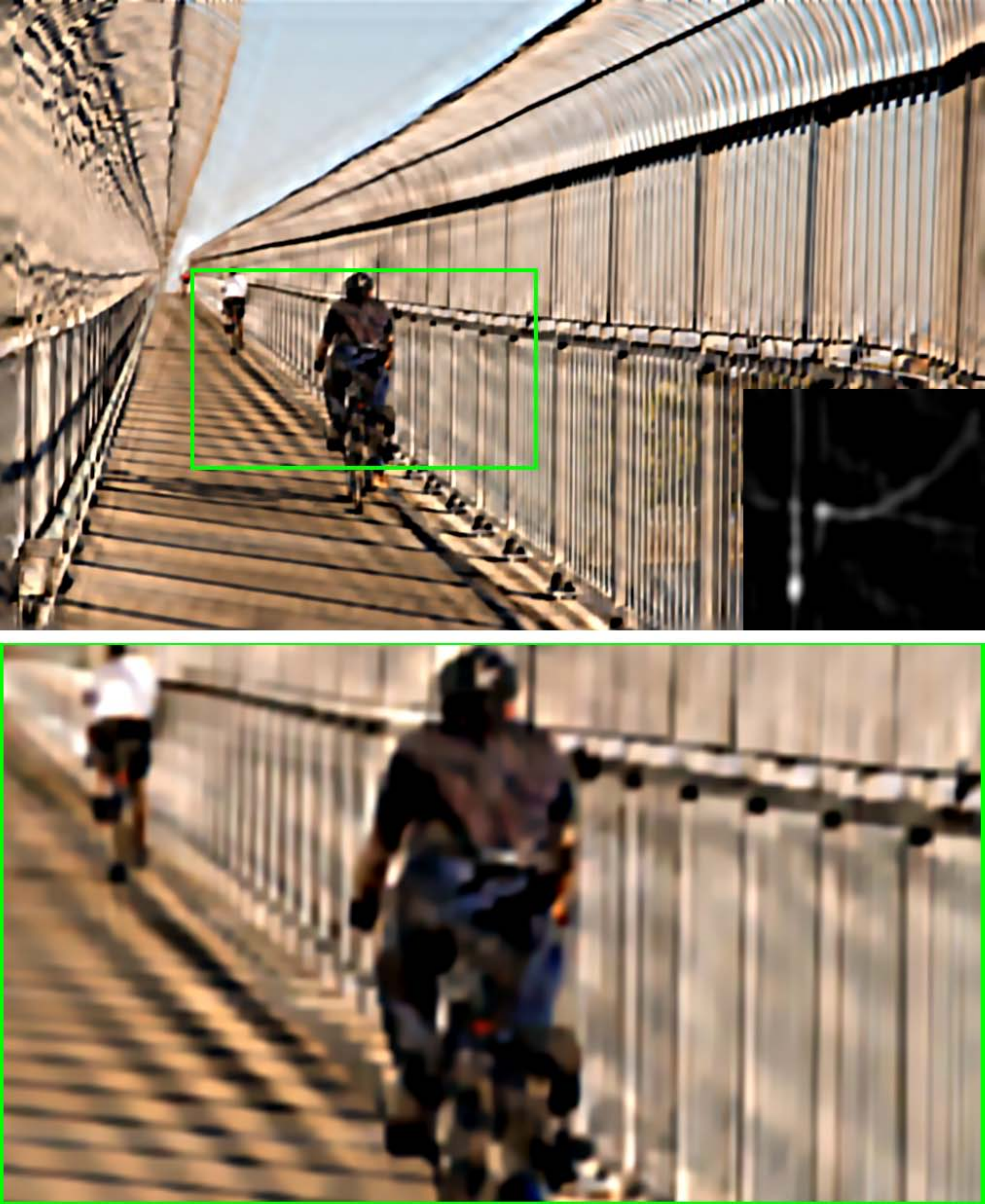}~
		&\includegraphics[width=0.18\textwidth]{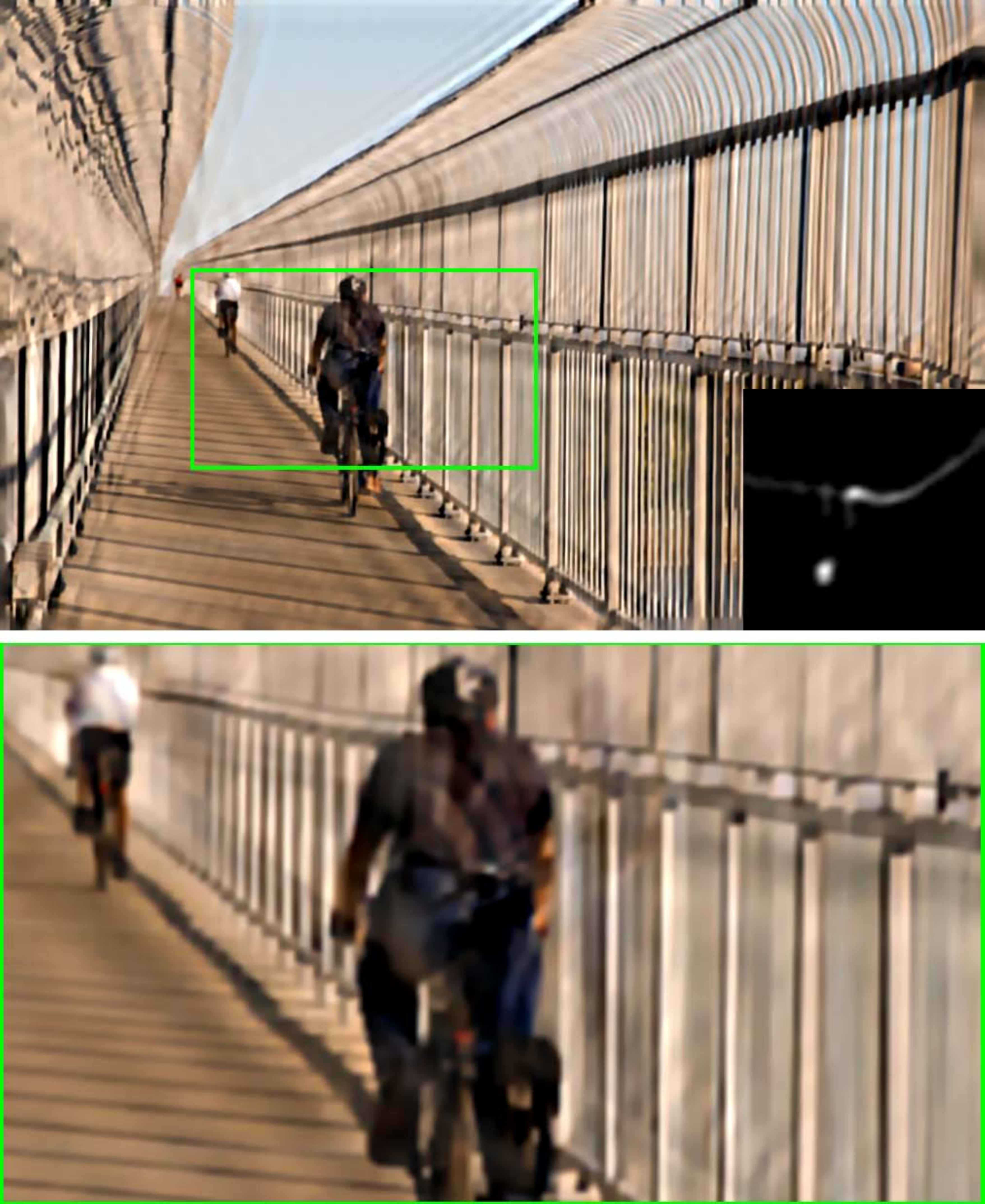}~
		&\includegraphics[width=0.18\textwidth]{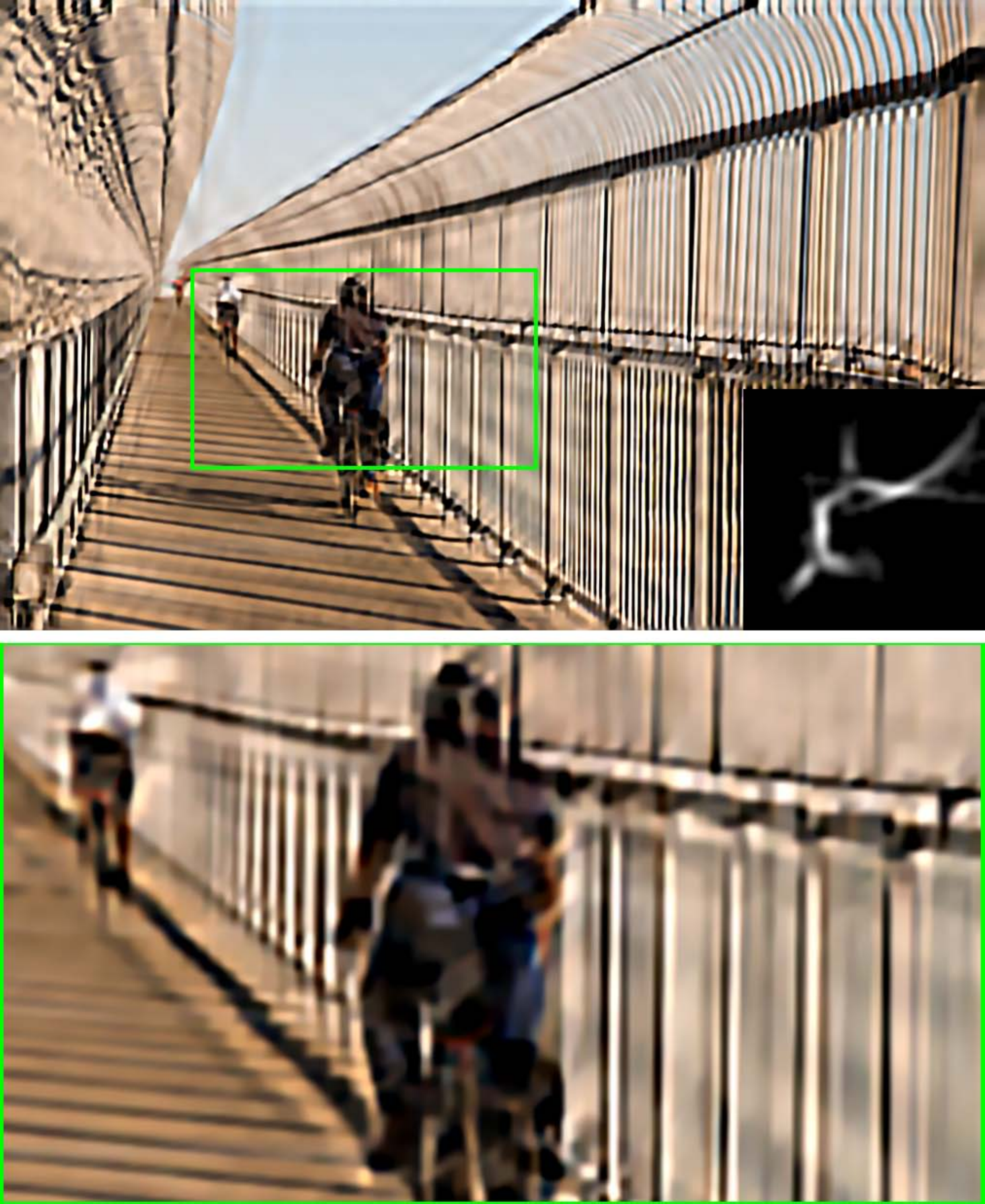}~
		&\includegraphics[width=0.18\textwidth]{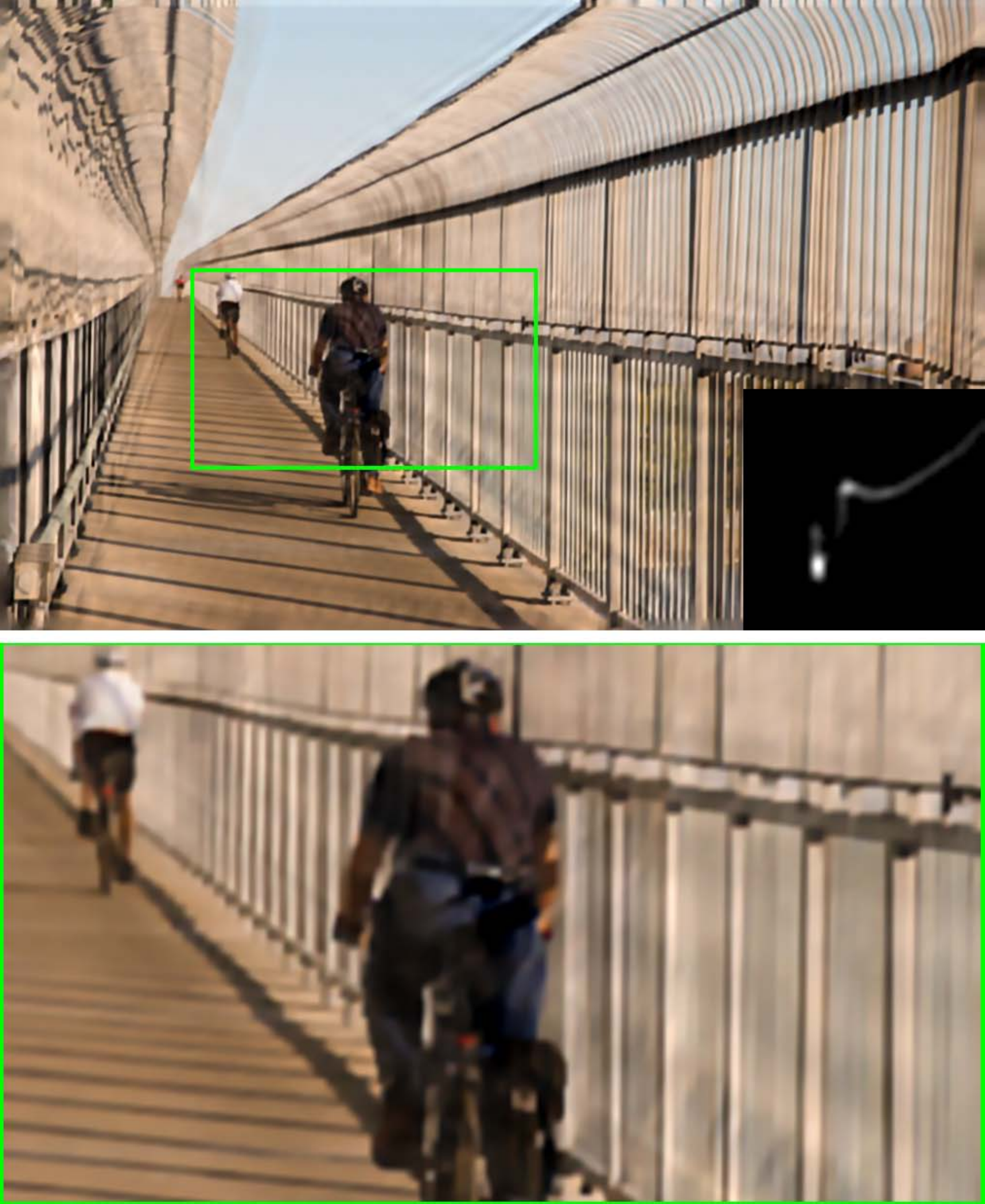}\\
		Input  &Perrone \emph{et al.} &Sun \emph{et al.} & Pan \emph{et al.} & mFIMA \\
		-&(15.96 / 0.49 / 0.80)&(17.35 / 0.60 / 0.88) &(14.39 / 0.44 / 0.54)&(18.11 / 0.58 / 0.95)\\
	\end{tabular}
	\caption{The blind image deconvolution results of mFIMA with comparisons to state-of-the-art approaches on  blurry image with 1\% Gaussian noise. The quantitative scores (i.e., PSNR / SSIM / KS) are reported below each image.}
	\label{fig:blind-manmade03-supp}
\end{figure*}

\begin{figure*}[tb]
	\centering
	\begin{tabular}{c@{\extracolsep{0.2em}}c@{\extracolsep{0.2em}}c@{\extracolsep{0.2em}}c@{\extracolsep{0.2em}}c}
		\includegraphics[width=0.18\textwidth]{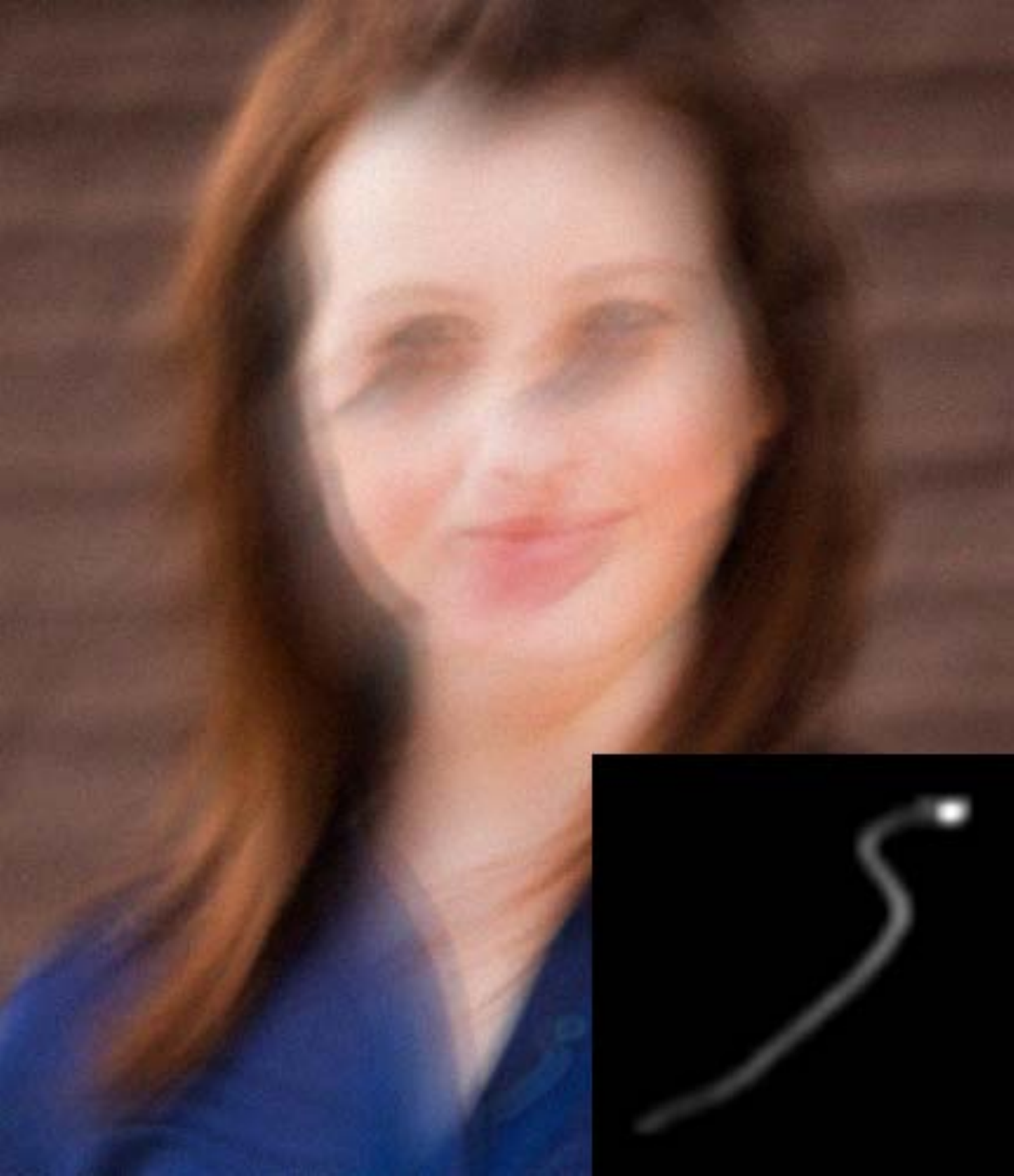}~
		&\includegraphics[width=0.18\textwidth]{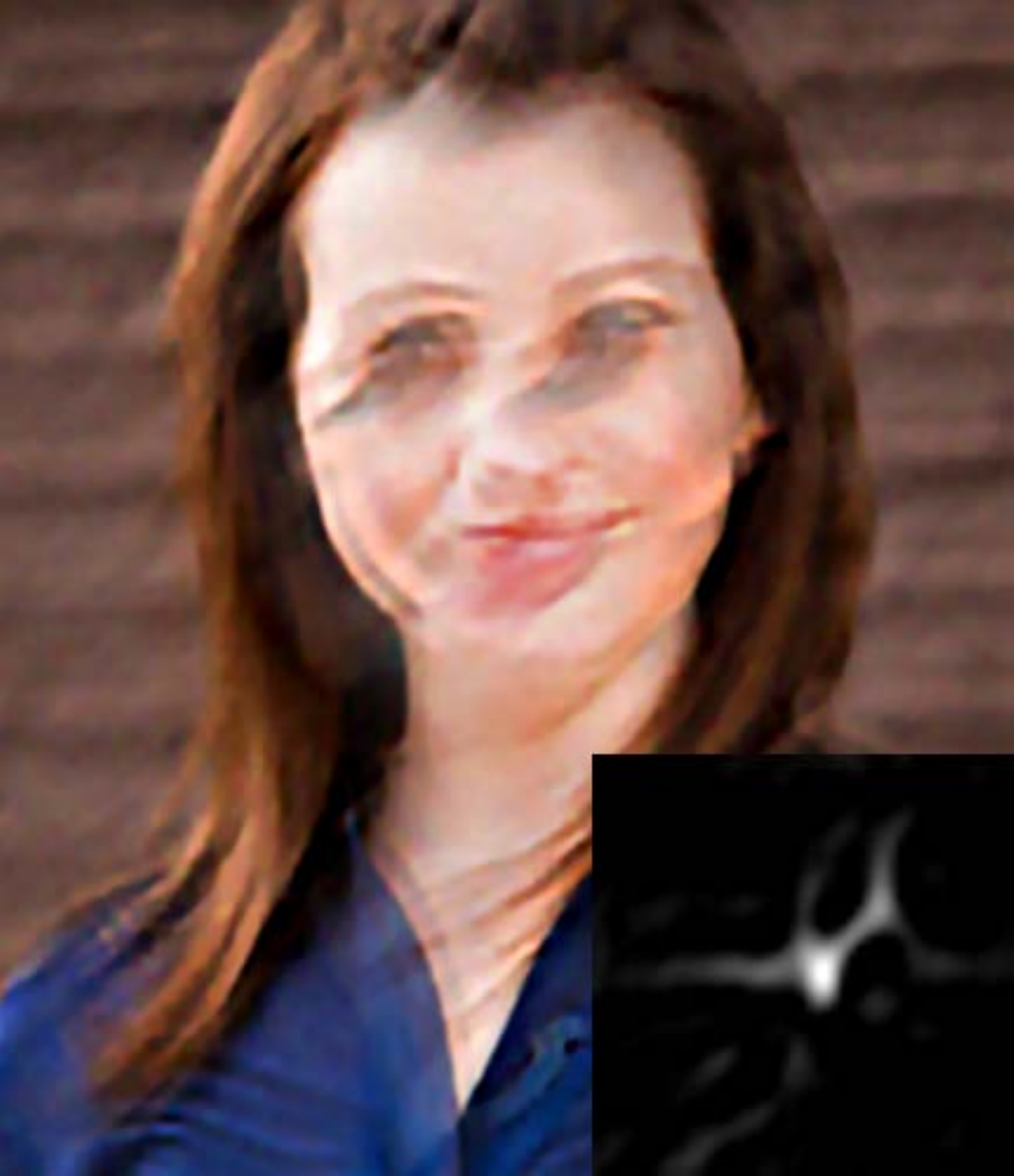}~
		&\includegraphics[width=0.18\textwidth]{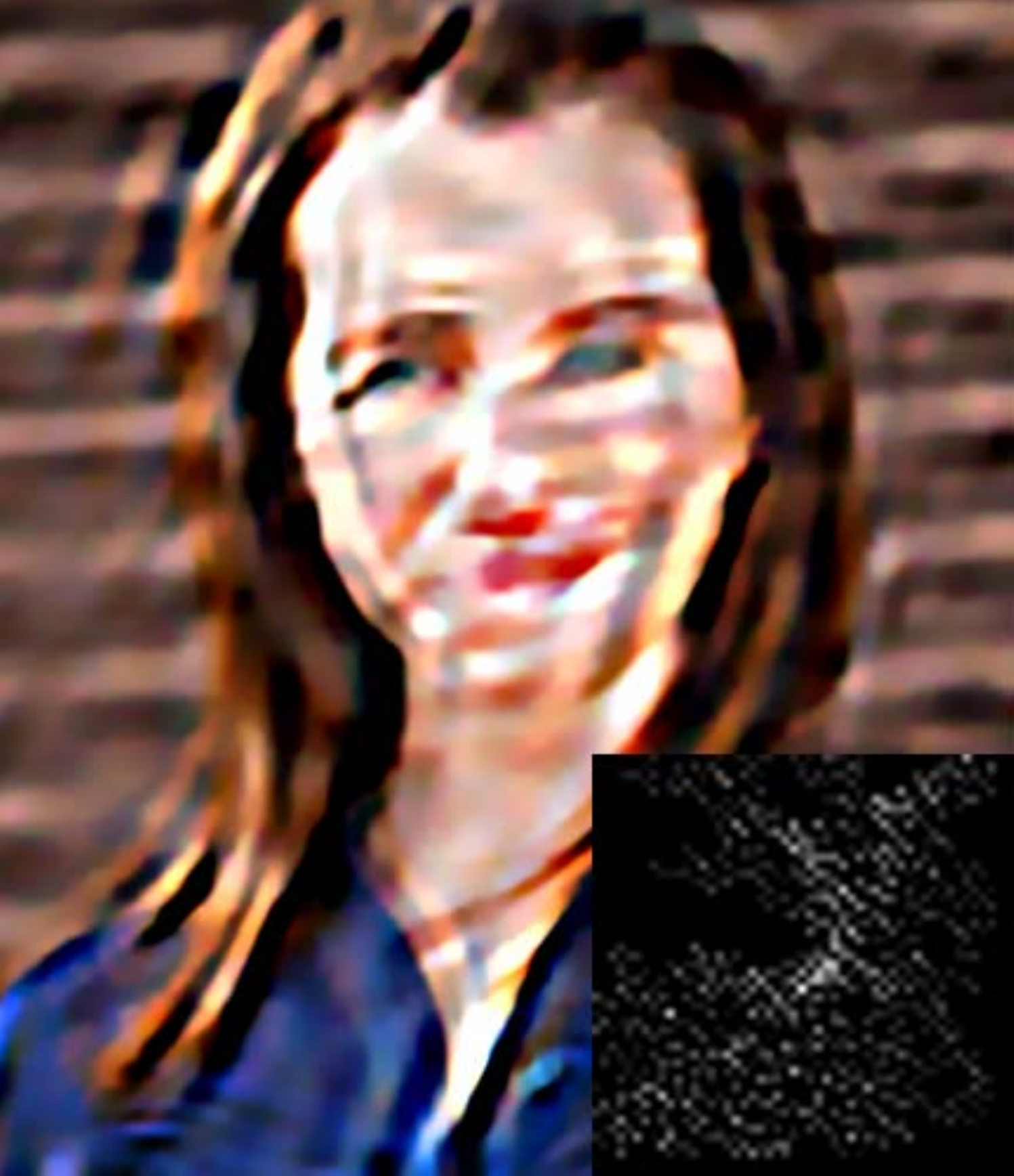}~
		&\includegraphics[width=0.18\textwidth]{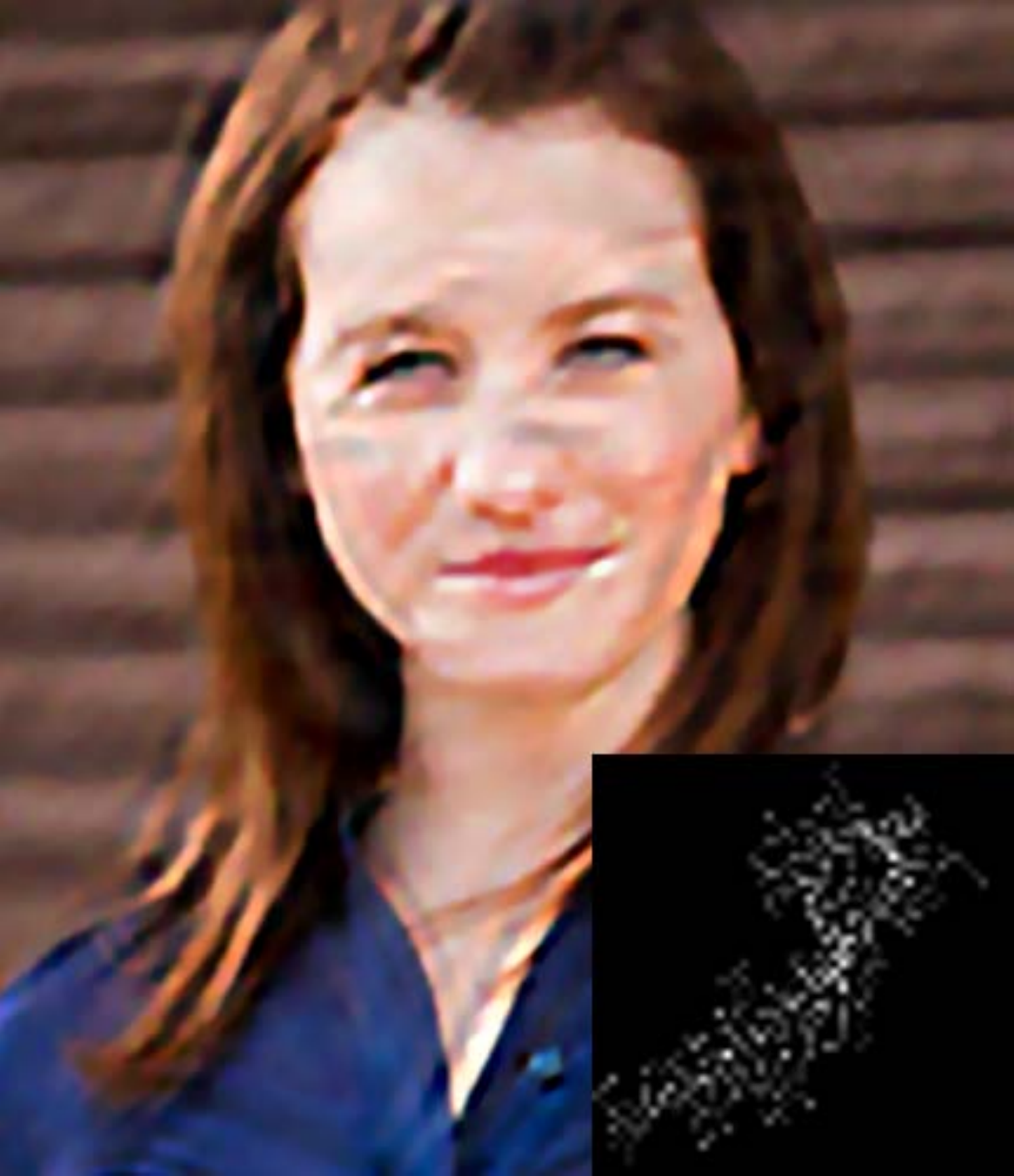}~
		&\includegraphics[width=0.18\textwidth]{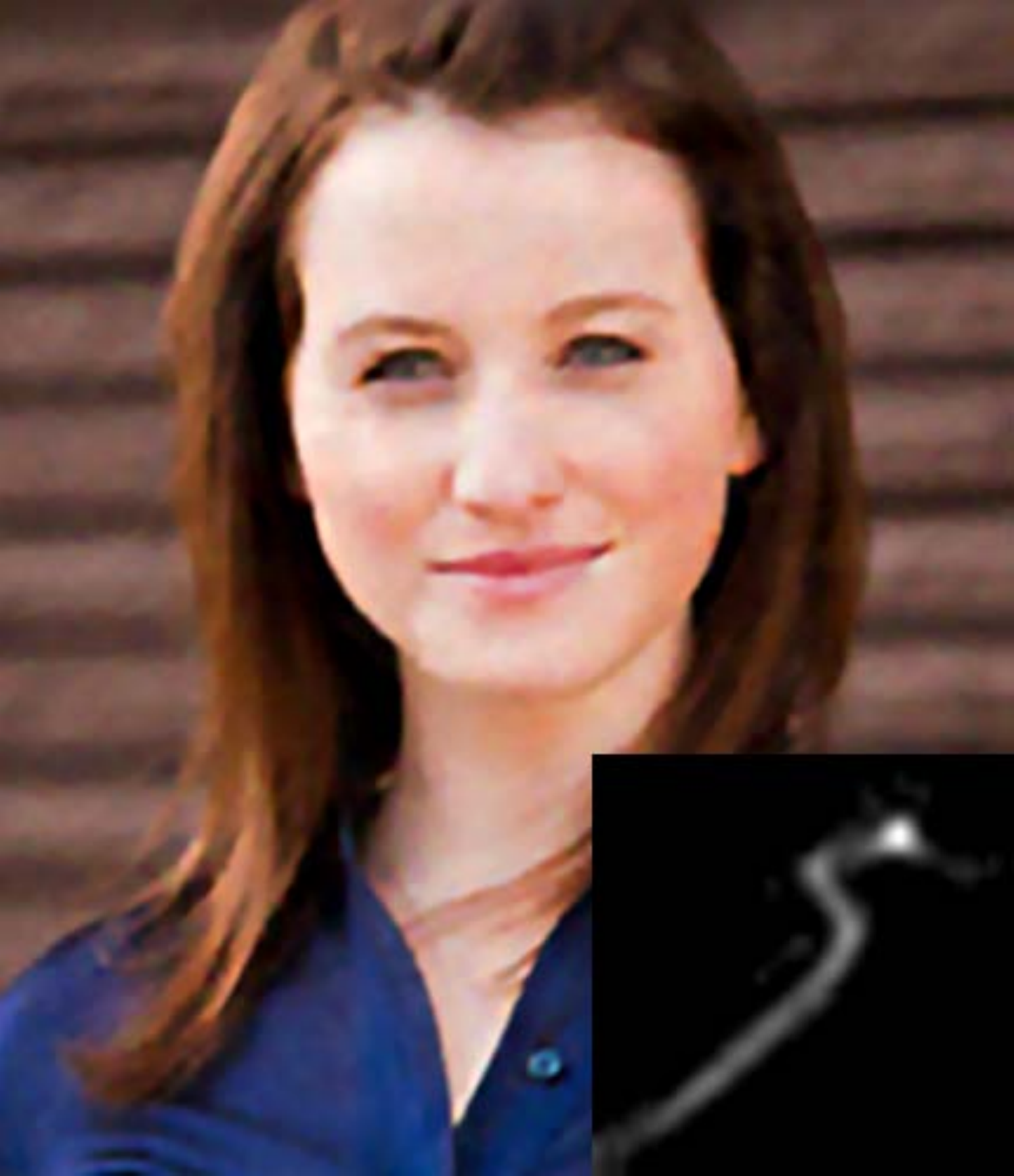}\\
		Input &Perrone \emph{et al.}&Sun \emph{et al.} &  Pan \emph{et al.} & mFIMA \\
		- & (24.76 / 0.75 / 0.48)& (20.48 / 0.56 / 0.32) &  (28.05 / 0.83 / 0.40)& (31.25 / 0.87 / 0.89)\\
	\end{tabular}
	\caption{The blind image deconvolution results of mFIMA with comparisons to state-of-the-art approaches on blurry facial image with 3\% Guassian noise. The quantitative scores (i.e., PSNR / SSIM / KS) are reported below each image.}
	\label{fig:blind-people-02-supp}
\end{figure*}

In Figs.~\ref{fig:blind-manmade03-supp} and~\ref{fig:blind-people-02-supp}, we further compare the blind image deconvolution performance of mFIMA with Perrone \emph{et al.}~\cite{perrone2014total}, Sun \emph{et al.}~\cite{patchdeblur_iccp2013}
and Pan \emph{et al.}~\cite{Pan2017Deblurring} (top 3 among all the compared methods in Tab. 2) on example images corrupted by not only unknown blur kernels, but also different levels of Gaussian noises (1\% and 3\% in Figs.~\ref{fig:blind-manmade03-supp} and~\ref{fig:blind-people-02-supp}, respectively). It can be seen that mFIMA is robust to these corruptions and outperforms all the compared state-of-the-art deblurring methods.

\section{Conclusion}
This paper provided FIMA, a framework to analyze the convergence behaviors of learning-based iterative methods for nonconvex inverse problems. We proposed two novel mechanisms to adaptively guide the trajectories of learning-based iterations and proved their strict convergence. We also showed how to apply FIMA for real-world applications, such as non-blind and blind image deconvolution.


%

\appendices

\section{Proofs}
We first give some preliminaries on variational analysis and nonconvex optimization in Sec.~\ref{sec:Preliminaries}.  Secs.~\ref{sec:Convergence-of-eFIMA}-\ref{sec:Convergence-of-mFIMA} then prove the main results in our manuscript.
\subsection{Preliminaries}
\label{sec:Preliminaries}

\begin{defi}\cite{rockafellar2009variational} The necessary function properties, including  proper, lower semi-continuous, Lipschitz smooth, and coercive are summarized as follows. Let $f: \mathbb{R}^D\to(-\infty, +\infty]$. Then we have
	\begin{itemize}
		\item Proper and lower semi-continuous: $f$ is proper if $\mathtt{dom}f:=\{\mathbf{x}\in\mathbb{R}^D: f(\mathbf{x})<+\infty\}$ is nonempty and $f(\mathbf{x})>-\infty$. $f$ is lower semi-continuous if $\liminf \limits_{\mathbf{x}\to\mathbf{y}}f(\mathbf{x})\geq f(\mathbf{y})$ at any point $\mathbf{y}\in\mathtt{dom}f$.
		\item Coercive: $f$ is said to be coercive, if $f$ is bounded from below and $f\to\infty$ if $\|\mathbf{x}\|\to\infty$, where $\|\cdot\|$ is the $\ell_2$ norm.
		\item $L$-Lipschitz smooth (i.e., $C_L^{1,1}$): $f$ is  $L$-Lipschitz smooth if $f$ is differentiable and there exists $L>0$ such that 
		$$
		\|\nabla f(\mathbf{x}) - \nabla f(\mathbf{y})\| \leq L \|\mathbf{x} - \mathbf{y}\|, \ \forall \ \mathbf{x},\mathbf{y} \in \mathbb{R}^{D}.
		$$
		If f is $L$-Lipschitz smooth, we have the following inequality
		$$
		f(\mathbf{x})\leq f(\mathbf{y}) + \langle \nabla f(\mathbf{y}), \mathbf{y}-\mathbf{x}\rangle + \frac{L}{2}\|\mathbf{x}-\mathbf{y}\|^2, \ \forall \mathbf{x}, \mathbf{y}\in\mathbb{R}^D.
		$$
	\end{itemize}
\end{defi}

\begin{defi}\cite{rockafellar2009variational,attouch2009convergence}
	Let $g:\mathbb{R}^D\to(-\infty,+\infty]$ be a proper and lower semi-continuous function. Then we have
	\begin{itemize}
		\item Sub-differential: The Frech¨¦t sub-differential (denoted as $\hat{\partial}g$) of $g$ at point $\mathbf{x} \in \mathtt{dom}(g)$ is the set of all vectors $\mathbf{z}$ which satisfies
		$$
		\liminf\limits_{\mathbf{y}\neq \mathbf{x},\mathbf{y} \to \mathbf{x}}
		\frac{g(\mathbf{y}) -g(\mathbf{x}) -\langle\mathbf{z}, \mathbf{y}-\mathbf{x}\rangle}{\|\mathbf{y}-\mathbf{x}\|} \geq 0,
		$$
		where $\langle\cdot,\cdot\rangle$ denotes the inner product. Then the limiting Frech¨¦t sub-differential (denoted as $\partial g$) at $\mathbf{x} \in \mathtt{dom} g$ is the following closure of $\hat{\partial}g$:
		$$
		\begin{array}{c}
		\{\mathbf{z}\in\mathbb{R}^n:\exists (\mathbf{x}^{k},g(\mathbf{x}^{k}))\to (\mathbf{x},g(\mathbf{x})) \},
		\end{array}
		$$
		where $\mathbf{z}^{k}\in\hat{\partial}g(\mathbf{x}^k)\to \mathbf{z}$ when $k\to\infty $.
		\item Kurdyka-{\L}ojasiewicz property: $g$ is said to have the Kurdyka-{\L}ojasiewicz property at $\bar{\mathbf{x}}\in\mathtt{dom}\partial g:=\{x\in\mathbb{R}^D: \partial g(x) \neq \emptyset\}$ if there exist $\eta\in(0,\infty]$, a neighborhood $\mathcal{U}_{\bar{\mathbf{x}}}$ of $\bar{\mathbf{x}}$ and a desingularizing function $\phi:[0,\eta)\to \mathbb{R}_+$ which satisfies (1) $\phi$ is continuous at $0$ and $\phi(0)=0$; (2) $\phi$ is concave and $C^1$ on $(0,\eta)$; (3) for all $s\in(0,\eta): \phi'(s)>0$, such that for all
		$$
		\mathbf{x}\in\mathcal{U}_{\bar{\mathbf{x}}}\cap[g(\bar{\mathbf{x}})<g(\mathbf{x})<g(\bar{\mathbf{x}})+\eta],
		$$
		the following inequality holds
		$$
		\phi'(g(\mathbf{x})-g(\bar{\mathbf{x}}))\mathtt{dist}(0,\partial g(x)) \geq 1.
		$$
		Moreover, if $g$ satisfies the K{\L} property at each point of $\mathtt{dom}\partial g$ then $g$ is called a K{\L} function.
		\item Semi-algebraic set and function: A subset $\Omega$ of $\mathbb{R}^D$ is a real semi-algebraic set if there exist a finit number of real polynomial functions $r_{ij}, h_{ij}:\mathbb{R}^D\to\mathbb{R}$ such that
		\begin{equation}
			\Omega = \bigcup\limits_{j=1}^p\bigcap\limits_{i=1}^q\left\{\mathbf{x}\in\mathbb{R}^D:
			r_{ij}(\mathbf{x})=0 \ \mbox{and} \ h_{ij}(\mathbf{x})<0\right\}.
		\end{equation}
		$g$ is called semi-algebraic if its graph $\{(\mathbf{x},z)\in\mathbb{R}^{D+1}: g(\mathbf{x})=z\}$ is a semi-algebraic subset of $\mathbb{R}^{D+1}$. 
		It is verified in \cite{attouch2009convergence} that all semi-algebraic functions satisfy the K{\L} property.
	\end{itemize}
\end{defi}

%

\subsection{Explicit Momentum FIMA (eFIMA)}\label{sec:Convergence-of-eFIMA}
\subsubsection{Proof of Theorem \ref{thm:eFIMA}}
\begin{proof}
	We first prove the inequality relationship of $\Psi\left(\mathbf{x}^{k+1}\right)$ and $\Psi\left(\mathbf{v}^k\right)$.
	According to the update rule of $\mathbf{x}^{k+1}$ (Step 8 in Alg.~\ref{alg:eFIMA}): $\mathbf{x}^{k+1} \in \mathtt{prox}_{{\gamma^k}g}\left(\mathbf{v}^{k}-\gamma^k\nabla f(\mathbf{v}^{k})\right)$ ), we have 
	\begin{equation}
		\mathbf{x}^{k+1}\in\arg\min\limits_{\mathbf{x}}g\left(\mathbf{x}\right)+\langle \nabla f\left(\mathbf{v}^k\right),\mathbf{x}-\mathbf{v}^k\rangle + \frac{1}{2\gamma^k}\|\mathbf{x}-\mathbf{v}^k\|^2,\label{eq:proximal}
	\end{equation}
	thus
	\begin{equation}
		g\left(\mathbf{x}^{k+1}\right)+
		\langle \nabla f\left(\mathbf{v}^k\right),\mathbf{x}^{k+1}-\mathbf{v}^k\rangle
		+ \frac{1}{2\gamma^k}\|\mathbf{x}^{k+1}-\mathbf{v}^k\|^2
		\leq g(\mathbf{v}^{k}).\label{eq:proximal-xk-vk}
	\end{equation}
	Since $f$ is $C_L^{1,1}$, we have 
	\begin{equation}
		f(\mathbf{x}^{k+1}) \leq f(\mathbf{v}^{k}) + \langle \nabla f\left(\mathbf{v}^k\right),\mathbf{x}^{k+1}-\mathbf{v}^k\rangle +
		\frac{L}{2}\| \mathbf{x}^{k+1}- \mathbf{v}^{k} \|^{2},\label{eq:Lipschitz-xk-vk}
	\end{equation}
	where $L$ is the Lipschitz moduli of $\nabla f$.
	Combining this with Eqs.~\eqref{eq:proximal-xk-vk} and \eqref{eq:Lipschitz-xk-vk}, we have
	\begin{equation}
		\Psi\left(\mathbf{x}^{k+1}\right)\leq\Psi\left(\mathbf{v}^k\right)-\left(\frac{1}{2\gamma^k}
		-\frac{L}{2}\right)\|\mathbf{x}^{k+1}-\mathbf{v}^k\|^2.\label{eq:psi-ineq}
	\end{equation}
	Set $\gamma^k< 1/L$ and define $\alpha^k = \frac{1}{2\gamma^k} -\frac{L}{2}$, we have $\alpha^k >0$ and $\Psi\left(\mathbf{x}^{k+1}\right)\leq \Psi\left(\mathbf{v}^{k}\right) - \alpha^k \|\mathbf{x}^{k+1}-\mathbf{v}^k\|^2$.
	
	Then we prove the boundness and convergence of $\{\mathbf{x}^k\}_{k\in\mathbb{N}}$. Based on the momentum scheduling policy in Alg.~\ref{alg:eFIMA}, we obviously have $\Psi\left(\mathbf{v}^{k}\right)\leq\Psi\left(\mathbf{x}^k\right)$. This together with the result in Eq.~\eqref{eq:psi-ineq} (i.e., $\Psi\left(\mathbf{x}^{k+1}\right) \leq \Psi\left(\mathbf{v}^{k}\right)$ with $\gamma^{k} < 1/L$)
	concludes that for any $k\in\mathbb{N}_{+}$,
	\begin{equation} \Psi\left(\mathbf{x}^{k+1}\right)\leq\Psi\left(\mathbf{v}^k\right)
		\leq\Psi\left(\mathbf{x}^k\right)\leq\Psi\left(\mathbf{v}^{k-1}\right)
		\leq \Psi\left(\mathbf{x}^0\right). \label{eq:psi-x-v}
	\end{equation}
	Since both $f$ and $g$ are proper, we also have $\Psi\left(\mathbf{v}^k\right)\geq\inf\Psi>-\infty$.
	Thus both sequences $\{\Psi\left(\mathbf{x}^k\right)\}_{k\in\mathbb{N}}$ and $\{\Psi\left(\mathbf{v}^k\right)\}_{k\in\mathbb{N}}$ are non-increasing and bounded.
	This together with the coercive of $\Psi$ concludes that both $\{\mathbf{x}^k\}_{k\in\mathbb{N}}$ and $\{\mathbf{v}^k\}_{k\in\mathbb{N}}$ are bounded and thus have accumulation points.
	
	Then we prove that all accumulation points are the critical points of $\Psi$.
	From Eq.~\eqref{eq:psi-x-v}, we actually have that the objective sequences $\{\Psi(\mathbf{x}^{k})\}_{k\in\mathbb{N}}$ and $\{\Psi(\mathbf{v}^{k})\}_{k\in\mathbb{N}}$ converge to  the same  value $\Psi^{*}$, i.e.,
	\begin{equation}
		\lim\limits_{k\to\infty}\Psi\left(\mathbf{x}^k\right)=\lim\limits_{k\to\infty}\Psi
		\left(\mathbf{v}^k\right)=\Psi^*.\label{eq:psi-accu}
	\end{equation}
	From Eqs.~\eqref{eq:psi-ineq} and \eqref{eq:psi-x-v}, we have
	\begin{equation}
		\begin{array}{l}
			\quad\left(\frac{1}{2\gamma^k}-\frac{L}{2}\right)\|\mathbf{x}^{k+1}-\mathbf{v}^{k}\|^2 \\
			\leq \Psi\left(\mathbf{v}^k\right)-\Psi\left(\mathbf{x}^{k+1}\right)
			\leq \Psi\left(\mathbf{x}^k\right)-\Psi\left(\mathbf{x}^{k+1}\right).
		\end{array}
	\end{equation}
	Summing over $k$, we further have
	\begin{equation}
		\min_{k}\left\{\frac{1}{2\gamma^k}-\frac{L}{2}\right\}\sum\limits_{k=0}^{\infty}\|\mathbf{x}^{k+1}
		-\mathbf{v}^{k}\|^2\leq\Psi\left(\mathbf{x}^0\right)-\Psi^*<\infty.
	\end{equation}
	The above inequality implies that $\|\mathbf{x}^{k+1}-\mathbf{v}^{k}\|\to 0$ and
	hence $\{\mathbf{x}^k\}_{k\in\mathbb{N}}$ and  $\{\mathbf{v}^k\}_{k\in\mathbb{N}}$
	share the same set of accumulation points (denoted as $\Omega$).
	Consider that $\mathbf{x}^{*} \in \Omega$ is any accumulation point of $\{\mathbf{x}^{k}\}_{k\in\mathbb{N}}$, i.e., $\mathbf{x}^{k_{j}} \to \mathbf{x}^{*}$ if $j \to \infty$.
	Then by Eq.~\eqref{eq:proximal}, we have
	\begin{equation}
		\begin{array}{l}
			\quad g\left(\mathbf{x}^{k+1}\right)+\langle \nabla f\left(\mathbf{v}^k\right),\mathbf{x}^{k+1}-\mathbf{v}^k\rangle + \frac{1}{2\gamma^k}\|\mathbf{x}^{k+1}-\mathbf{v}^k\|^2\\
			\leq g\left(\mathbf{x}^*\right)+\langle \nabla f\left(\mathbf{v}^k\right),\mathbf{x}^*-\mathbf{v}^k\rangle + \frac{1}{2\gamma^k}\|\mathbf{x}^*-\mathbf{v}^k\|^2.
		\end{array} \label{eq:subseq-x-v}
	\end{equation}
	Let $k_{j} = k+1$ in Eq.~\eqref{eq:subseq-x-v} and $j \to \infty$ ,
	by taking $\limsup$ on both sides of Eq.~\eqref{eq:subseq-x-v}, we have
	$
	\limsup \limits_{j\to\infty}g\left(\mathbf{x}^{k_{j}}\right)\leq g\left(\mathbf{x}^*\right)
	$.
	On the other hand, since $g$ is lower semi-continuous and $\mathbf{x}^{k_{j}}\to\mathbf{x}^*$, it follows that $\liminf\limits_{j\to\infty}g\left(\mathbf{x}^{k_{j}}\right)\geq g\left(\mathbf{x}^*\right)$. So we have
	$\lim\limits_{j\to\infty}g\left(\mathbf{x}^{k_{j}}\right)= g\left(\mathbf{x}^*\right)$.
	Note that the continuity of $f$ yields $\lim\limits_{j\to\infty}f\left(\mathbf{x}^{k_{j}}\right)= f\left(\mathbf{x}^*\right)$, so we conclude
	\begin{equation}
		\lim\limits_{j\to\infty}\Psi\left(\mathbf{x}^{k_{j}}\right)= \Psi\left(\mathbf{x}^*\right). \label{eq:psi-lim}
	\end{equation}
	Recall that $\lim\limits_{k\to\infty}\Psi\left(\mathbf{x}^{k+1}\right)= \Psi^*$ in Eq.~\eqref{eq:psi-accu}, we have
	$\lim\limits_{j\to\infty}\Psi\left(\mathbf{x}^{k_j}\right)= \Psi^*$, so
	\begin{equation}
		\Psi\left(\mathbf{x}^*\right) =\Psi^*, \ \forall \ \mathbf{x}^*\in\Omega.
	\end{equation}	
	By first-order optimality condition of Eq.~\eqref{eq:proximal} and $k_{j} = k+1$, we have
	\begin{equation}
		\begin{array}{l}
			\quad \mathbf{0}\in\partial g\left(\mathbf{x}^{k_{j}}\right) + \nabla f\left(\mathbf{v}^{k}\right) + \frac{1}{\gamma^k}\left(\mathbf{x}^{k_{j}}-\mathbf{v}^k\right). \label{eq:gradient}
		\end{array}
	\end{equation}
	Thus, we have
		\begin{equation}
	\begin{array}{l}
	\nabla f\left(\mathbf{x}^{k_{j}}\right) - \nabla f\left(\mathbf{v}^{k}\right) - \frac{1}{\gamma^k}\left(\mathbf{x}^{k_{j}}-\mathbf{v}^k\right) \in\partial \Psi\left(\mathbf{x}^{k_{j}}\right)\\
	\Rightarrow\|\nabla f\left(\mathbf{x}^{k_{j}}\right) - \nabla f\left(\mathbf{v}^{k}\right) - \frac{1}{\gamma^k}\left(\mathbf{x}^{k_{j}}-\mathbf{v}^k\right)\| \\
	\quad \leq \left(L + \frac{1}{\gamma^k}\right)\|\mathbf{x}^{k_{j}}-\mathbf{v}^k\|\to 0, \ \mbox{as} \ j\to\infty.\label{eq:gradient_deduce}
	\end{array}
	\end{equation}
	Then from the definition of sub-differential and Eqs.~\eqref{eq:psi-lim}, \eqref{eq:gradient}, and \eqref{eq:gradient_deduce}, we conclude that
	\begin{equation}
		\mathbf{0}\in\partial \Psi\left(\mathbf{x}^*\right), \ \forall \mathbf{x}^*\in\Omega.
	\end{equation}
	Therefore, we have that all accumulation points $\mathbf{x}^{*}$ are the critical points of $\Psi$.	
\end{proof}

\subsubsection{Proof of Corollary \ref{cor:eFIMA}}
\begin{proof}
	Considering the semi-algebraic (thus K{\L}) property of $\Psi(\mathbf{x})$ and defining a desingularizing function with the form  $\phi(s)=\frac{t}{\theta}s^{\theta}$, 
	we can prove \textbf{Corollary~\ref{cor:eFIMA}} by Eqs.~\eqref{eq:psi-ineq},~\eqref{eq:psi-x-v}, and~\eqref{eq:gradient} using similar methodology as that in \cite{attouch2009convergence,chouzenoux2016block}. Since these derivations are quite standard, we omit details of this proof in our Supplemental Materials.
\end{proof}

\subsection{Implicit Momentum FIMA (iFIMA)}\label{sec:Convergence-of-iFIMA}

\subsubsection{Proof of Proposition \ref{prop:x-u}}
\begin{proof}
	First, by using the same derivations as that in Eq.~\eqref{eq:psi-ineq}, we can directly obtain the inequality in Theorem~\ref{alg:eFIMA} for iFIMA. Then we show how to build the relationship between $\Psi(\tilde{\mathbf{u}}^k)$ and $\Psi(\mathbf{x}^{k})$.
	It is known that $\tilde{\mathbf{u}}^{k}$ is actually an inexact minimizer of $\Psi^{k}$. But by defining its sub-differential $\mathbf{d}^{\tilde{\mathbf{u}}^{k}}_{\Psi^k}\in\partial\Psi^k(\tilde{\mathbf{u}})$
	as that in Eq.~\eqref{eq:error}, we can also consider it as the exact solution to the following problem
	\begin{equation}
		\tilde{\mathbf{u}}^{k} \in \arg \min\limits_{\mathbf{x}}\Psi^{k}\left(\mathbf{x}\right) - \langle \mathbf{d}^{\tilde{\mathbf{u}}^{k}}_{\Psi^k}, \mathbf{x}\rangle.\label{eq:prox-error}
	\end{equation}
	Thus, we have
	\begin{equation}
		\begin{array}{l}
			\quad \Psi\left(\tilde{\mathbf{u}}^{k}\right) +  \frac{\mu^k}{2}\|\tilde{\mathbf{u}}^{k}-\mathbf{x}^{k}\|^2  - \langle \mathbf{d}^{\tilde{\mathbf{u}}^{k}}_{\Psi^k}, \tilde{\mathbf{u}}^{k}\rangle \\
			\leq \Psi\left(\mathbf{x}^{k}\right) - \langle \mathbf{d}^{\tilde{\mathbf{u}}^{k}}_{\Psi^k}, \mathbf{x}^{k}\rangle \\
			\Rightarrow \Psi\left(\tilde{\mathbf{u}}^{k}\right)
			\leq \Psi\left(\mathbf{x}^{k}\right) - \frac{\mu^k}{2}\|\tilde{\mathbf{u}}^{k}-\mathbf{x}^{k}\|^2
			+ \langle \mathbf{d}^{\tilde{\mathbf{u}}^{k}}_{\Psi^k}, \tilde{\mathbf{u}}^{k} - \mathbf{x}^{k}\rangle \\
			\leq \Psi\left(\mathbf{x}^{k}\right)- \frac{\mu^k}{2}\|\tilde{\mathbf{u}}^{k}-\mathbf{x}^{k}\|^2 + C^{k}\|\tilde{\mathbf{u}}^{k} -\mathbf{x}^{k} \|^2\\
			= \Psi\left(\mathbf{x}^{k}\right)- \left(\frac{\mu^k}{2}-C^{k}\right)\|\tilde{\mathbf{u}}^{k}-\mathbf{x}^{k}\|^2,
		\end{array} \label{eq:condition-y-x}
	\end{equation}
	in which the second inequality holds under Cauchy-Schwarz inequality and our error-control-based scheduling policy in Alg.~\ref{alg:iFIMA}. Set $ C^{k}< \frac{\mu^{k}}{2}$ and define $\beta^k = \frac{\mu^{k}}{2}-C^{k}$, we have $\beta^k >0$ and $\Psi\left(\tilde{\mathbf{u}}^{k}\right) \leq \Psi\left(\mathbf{x}^{k}\right) - \beta^k \|\tilde{\mathbf{u}}^{k}-\mathbf{x}^{k}\|^2$, which concludes the proof.
\end{proof}

\subsubsection{Proof of Theorem \ref{thm:iFIMA}}
\begin{proof}
	We first prove the boundedness of $\{\mathbf{x}^{k}\}_{k\in\mathbb{N}}$. According to \textbf{Proposition}~\ref{prop:x-u} we have $\Psi\left(\tilde{\mathbf{u}}^{k}\right) \leq \Psi\left(\mathbf{x}^{k}\right)$ when $\mu^k/2 > C^{k} $. So if the error-control criteria in Alg.~\ref{alg:iFIMA} is satisfied, we have $\mathbf{v}^{k} = \tilde{\mathbf{u}}^{k}$, $ \Psi\left(\mathbf{v}^{k}\right) = \Psi\left(\tilde{\mathbf{u}}^{k}\right) \leq \Psi\left(\mathbf{x}^{k}\right)$, otherwise, we have $\mathbf{v}^{k} = \mathbf{x}^{k}$,  $ \Psi\left(\mathbf{v}^{k}\right) = \Psi\left(\mathbf{x}^{k}\right)$.
	This together with the results in Theorem~\ref{thm:eFIMA} (i.e., $\Psi\left(\mathbf{x}^{k+1}\right) \leq \Psi\left(\mathbf{v}^{k}\right)$ with $\gamma^{k} < 1/L$)
	concludes that for any $k\in\mathbb{N}_{+}$,
	\begin{equation} \Psi\left(\mathbf{x}^{k+1}\right)\leq\Psi\left(\mathbf{v}^k\right)
		\leq\Psi\left(\mathbf{x}^k\right)\leq\Psi\left(\mathbf{v}^{k-1}\right)
		\leq \Psi\left(\mathbf{x}^0\right). \label{eq:psi-x-v-i}
	\end{equation}
	Then by using similar derivations as that in \textbf{Theorem}~\ref{thm:eFIMA}, we have that all accumulation points $\mathbf{x}^{*}$ are the critical points of $\Psi$.
	
	Now we are ready to prove that $\{\mathbf{x}^{k}\}_{k \in \mathbb{N}}$ is a Cauchy sequence. 
	Following the inequalities in \textbf{Proposition~\ref{prop:x-u}}, we have
	\begin{equation}
		\begin{array}{l}
			\quad \min\limits_{k} \left\{ \frac{1}{2\gamma^k}-\frac{L}{2}, \frac{\mu^{k}}{2} -C^{k} \right\} \sum\limits_{k=0}^{\infty} \|\mathbf{x}^{k+1} -\mathbf{x}^{k}\|^2 \\
			\leq \sum\limits_{k=0}^{\infty} \left( \left(\frac{1}{2\gamma^k}-\frac{L}{2} \right) \|\mathbf{x}^{k+1} -\mathbf{v}^{k} \|^{2} + \left(\frac{\mu^{k}}{2} -C^{k} \right) \|\mathbf{v}^{k} -\mathbf{x}^{k} \|^{2} \right)\\
			\leq \sum\limits_{k=0}^{\infty} \left(\Psi\left(\mathbf{v}^{k}\right) - \Psi\left(\mathbf{x}^{k+1}\right) + \Psi\left(\mathbf{x}^{k}\right) -\Psi\left(\mathbf{v}^{k}\right) \right) \\  
			=\Psi\left(\mathbf{x}^{0}\right)-\Psi^*<\infty.
		\end{array} \label{eq:xk1-xk}
	\end{equation}
	
	Since $\Psi$ is a semi-algebraic function, it satisfies the K{\L} property. So we have that $\sum_{k=0}^{\infty} \|\mathbf{x}^{k+1} - \mathbf{x}^{k}\| < \infty$ following~\cite{attouch2010proximal}
	and Eq.~\eqref{eq:xk1-xk}. This implies that 
	$\{\mathbf{x}^{k}\}_{k \in \mathbb{N}}$
	is a Cauchy sequence. Thus the sequence globally converges to a critical point of $\Psi(\mathbf{x})$ in Eq.~\eqref{eq:psi}.
\end{proof}

\subsection{Multi-block FIMA}\label{sec:Convergence-of-mFIMA}

\subsubsection{Definition Extension}
As for the generalized Lipschitz smooth property of $f$, we actually need $f$ satisfy that
\begin{itemize}
	\item For each $\mathbf{x}_n$ with other variables fixed, there exits $L_n>0$ such that
	\begin{equation}
	\begin{array}{l}
	\| \nabla_{n} f\left(\mathfrak{X}_{[<n]},\mathbf{x}_{n},\mathfrak{X}_{[>n]}\right) - \nabla_{n} f\left(\mathfrak{X}_{[<n]},\mathbf{y}_{n},\mathfrak{X}_{[>n]}\right)\| \\
	\leq L_{n}\left(\mathfrak{X}_{[<n]},\mathfrak{X}_{[>n]}\right)\|\mathbf{x}_{n} - \mathbf{y}_{n}\|,  \ \forall \mathbf{x}_{n},\mathbf{y}_{n}\in \mathbb{R}^{D_n},
	\end{array}
	\label{eq:lip-partial}
	\end{equation}
	where $\nabla_n$ denotes the gradient with respect to $\mathbf{x}_n$. 
	\item For each bounded subset $\Omega_{1}\times \dots \times \Omega_{N}\subseteq\mathbb{R}^{D_1}\times\cdots\times\mathbb{R}^{D_N}$, there exists $M >0$ such that
	\begin{equation}
	\begin{array}{l}
	\| \left(\nabla_{1}f(\mathfrak{X}) -\nabla_{1}f(\mathfrak{Y}), \dots, \nabla_{N}f(\mathfrak{X}) -\nabla_{N}f(\mathfrak{Y})\right)\|\\
	\leq M\|\mathfrak{X}-\mathfrak{Y}\|, \ \forall \ \mathfrak{X}, \mathfrak{Y} \in \Omega_{1}\times \dots \times \Omega_{N}.
	\end{array}\label{eq:lip-joint}
	\end{equation}
\end{itemize}

\subsubsection{Proof of Corollary \ref{cor:MFIMA-ims}}
\begin{proof}
	
	We first prove the boundedness of $\left\{ \mathfrak{X}^{k} \right\}_{k\in \mathbb{N}}$.
	Using the inequality in Theorem~\ref{thm:eFIMA} and Step 10 in Alg.~\ref{alg:MFIMA}, we have 
	\begin{equation}
	\begin{array}{l}
	\mathbf{x}^{k+1}_{n}\in\arg\min_{\mathbf{x}_{n}} g_n\left(\mathbf{x}_n\right)+ \frac{1}{2 \gamma_n^k}\| \mathbf{x}_n-\mathbf{v}^k_n \|^2\\
	\qquad \quad +\left\langle \nabla_{n} f\left(\mathfrak{X}^{k+1}_{[< n]},\mathbf{v}^{k}_{n},\mathfrak{X}^{k}_{[> n]} \right),\mathbf{x}_n-\mathbf{v}^k_n \right\rangle 
	.
	\end{array}\label{eq:multi-mc-min-x}
	\end{equation}
	This together with Eq.~\eqref{eq:lip-partial} concludes that
	\begin{equation}
		\begin{array}{l}
			\quad\Psi\left(\mathfrak{X}^{k+1}_{[< n]},\mathbf{v}^{k}_{n},\mathfrak{X}^{k}_{[> n]}\right) - \Psi\left(\mathfrak{X}^{k+1}_{[\leq n]},\mathfrak{X}^{k}_{[> n]}\right)\\
			\geq\left(\frac{1}{2\gamma_n^k}-\frac{L_{n}\left(\mathfrak{X}^{k+1}_{[< n]}, \mathfrak{X}^{k}_{[> n]} \right)}{2} \right)\|\mathbf{x}^{k+1}_{n}-\mathbf{v}^k_n\|^2.
		\end{array}\label{eq:multi-inequ-x-v}
	\end{equation}
	Define the auxiliary function $\Psi^{k}_{n}(\mathbf{x}_{n}) = f\left( \mathfrak{X}^{k+1}_{[<n]},\mathbf{x}_{n},\mathfrak{X}^{k}_{[>n]}\right) + g_{n}\left(\mathbf{x}_{n}\right) + \frac{\mu^{k}_n}{2} \| \mathbf{x}_{n} - \mathbf{x}^{k}_{n}\|^{2}$.
	Then by considering that $\tilde{\mathbf{u}}^{k}_{n}$ is an inexact solution of the auxiliary function $\Psi^{k}_{n}(\mathbf{x}_{n})$, and applying \textbf{Proposition}~\ref{prop:x-u},
	we have
	\begin{equation}
	\begin{array}{l}
		\quad \Psi\left(\mathfrak{X}^{k+1}_{[<n]},\mathbf{x}^{k}_{n},\mathfrak{X}^{k}_{[>n]}\right) - \Psi\left(\mathfrak{X}^{k+1}_{[<n]},\tilde{\mathbf{u}}^{k}_{n},\mathfrak{X}^{k}_{[>n]}\right) \\
		\geq  \left(\frac{\mu^k_n}{2}-C^{k}_{n}\right)\|\tilde{\mathbf{u}}^{k}_{n}-\mathbf{x}^{k}_{n}\|^2. 
	\end{array}\label{eq:multi-c}
	\end{equation}
	Let $\lambda_{n}^{k} =L_{n}\left(\mathfrak{X}^{k+1}_{[< n]}, \mathfrak{X}^{k}_{[> n]} \right) $, then
	consider Eq.~\eqref{eq:multi-inequ-x-v} with $\gamma_{n}^{k} < 1/\lambda_{n}^{k}$, Eq.~\eqref{eq:multi-c} with $\mu_{n}^{k} > 2C_{n}^{k}$, and our error-control updating rule, we have		
	$$
	\begin{array}{l}
	\quad \Psi\left(\mathfrak{X}^{k+1}_{[\leq n]},\mathfrak{X}^{k}_{[> n]}\right) 
	\leq \Psi\left(\mathfrak{X}^{k+1}_{[< n]},\mathbf{v}_{n}^{k},\mathfrak{X}^{k}_{[> n]}\right)\\
	\leq \Psi\left(\mathfrak{X}^{k+1}_{[<n]},\mathbf{x}^{k}_{n},\mathfrak{X}^{k}_{[>n]}\right).
	\end{array}
	$$
	It concludes that for any $k \in \mathbb{N}_{+}$ and $n \in \{1,\dots,N\}$, 
	\begin{equation}
		\begin{array}{l}
			\quad \Psi\left(\mathfrak{X}^{k+1}\right) =
			\Psi\left(\mathfrak{X}^{k+1}_{[\leq N]},\mathfrak{X}^{k}_{[> N]}\right) \\
			\leq \Psi\left(\mathfrak{X}^{k+1}_{[< n]},\mathbf{v}_{n}^{k},\mathfrak{X}^{k}_{[> n]}\right) 
			\leq \Psi\left(\mathfrak{X}^{k+1}_{[<n]},\mathbf{x}^{k}_{n},\mathfrak{X}^{k}_{[>n]}\right)  \\
			\leq \Psi\left(\mathfrak{X}^{k+1}_{[<1]},\mathbf{x}^{k}_{1},\mathfrak{X}^{k}_{[>1]}\right)
			= \Psi\left( \mathfrak{X}^{k} \right)
			\leq\cdots \leq \Psi\left(\mathfrak{X}^{0}\right).
		\end{array} \label{eq:multi-psi-x-v}
	\end{equation}		 	
	Since $f,g_{n}$ are proper, we also have
	$-\infty <\inf{\Psi} \leq \Psi\left(\mathfrak{X}^{k+1}\right)$. Thus  
	$\left\{ \Psi \left(\mathfrak{X}^{k}\right) \right\}_{k\in\mathbb{N}}$ and $\left\{ \Psi\left(\mathfrak{X}^{k+1}_{[< n]},\mathbf{v}^{k}_{n},\mathfrak{X}^{k}_{[> n]}\right)\right\}_{k\in\mathbb{N}}$
	are all non-increasing and bounded. This together with the coercive of $\Psi$ concludes that
	the sequences 
	$\{\mathfrak{X}^{k}\}_{k\in\mathbb{N}}$ and $\{\mathbf{v}_n^{k}\}_{k\in\mathbb{N}}$ ($1\leq n\leq N$)
	are bounded and thus have accumulation points.
	
	Then we prove that all accumulation points are the critical points of $\Psi$. From Eq.~\eqref{eq:multi-psi-x-v}, we have that the function value sequences
	$\left\{ \Psi \left(\mathfrak{X}^{k}\right) \right\}_{k\in\mathbb{N}}$ and $\left\{ \Psi\left(\mathfrak{X}^{k+1}_{[< n]},\mathbf{v}^{k}_{n},\mathfrak{X}^{k}_{[> n]}\right)\right\}_{k\in\mathbb{N}}$
	converge to the same value $\Psi^{*}$, i.e.,
	\begin{equation}
		\begin{array}{l}
			\lim\limits_{k\to\infty}\Psi\left(\mathfrak{X}^k\right)
			= \lim\limits_{k\to\infty}\Psi\left(\mathfrak{X}^{k+1}_{[< n]},\mathbf{v}^{k}_{n},\mathfrak{X}^{k}_{[> n]}\right)
			=\Psi^*.
		\end{array}\label{eq:multi-psi-accu}
	\end{equation}
	From Eqs.~\eqref{eq:multi-inequ-x-v} and \eqref{eq:multi-psi-x-v},
	summing over $k$ and $n$ we have 
	
	\begin{equation}
		\begin{array}{l}
			\quad \min\limits_{k,n} \left\{\frac{1}{2\gamma_{n}^{k}}-\frac{\lambda_{n}^{k}}{2}\right\}\sum\limits_{k=0}^{\infty} \sum\limits_{n=1}^{N}\|\mathbf{x}^{k+1}_{n}-\mathbf{v}^k_n\|^2 \\
			\leq \sum\limits_{k=0}^{\infty} \sum\limits_{n=1}^{N} \left( \Psi\left(\mathfrak{X}^{k+1}_{[< n]},\mathbf{v}^{k}_{n},\mathfrak{X}^{k}_{[> n]}\right) - \Psi\left(\mathfrak{X}^{k+1}_{[\leq n]},\mathfrak{X}^{k}_{[> n]}\right) \right) \\
			\leq \sum\limits_{k=0}^{\infty} \left( \Psi \left( \mathfrak{X}^{k}\right) - \Psi \left(\mathfrak{X}^{k+1}\right) \right)
			= \Psi \left( \mathfrak{X}^{0}\right) - \Psi^{*} < \infty.
		\end{array} \label{eq:multi-sum-x-v}
	\end{equation}	  
	The above inequality implies that $ \|\mathbf{x}^{k+1}_{n}-\mathbf{v}^k_n\| \to 0$, hence
	$\left\{\mathbf{x}_n^{k}\right\}_{k\in\mathbb{N}}$ and $\left\{ \mathbf{v}^{k}_{n}\right\}_{k\in\mathbb{N}}$
	share the same set of accumulation points.
	Consider that  $\mathfrak{X}^{*}=\{\mathbf{x}_1^*,\cdots,\mathbf{x}_N^*\}$ is any accumulation point of $\{\mathfrak{X}^{k}\}_{k \in \mathbb{N}}$ , there exists a subsequence $\{\mathfrak{X}^{k_j}\}_{j\in\mathbb{N}}$ such that
	\begin{equation}
		\mathbf{x}^{k_j}_{n} \to \mathbf{x}^{*}_{n}, \ \mbox{as} \ j \to \infty.
	\end{equation}
	From Eq.~\eqref{eq:multi-mc-min-x}, we have
	\begin{equation}
		\begin{array}{l}
			\quad g_n\left(\mathbf{x}^{k+1}_{n}\right)+\left\langle \nabla_{n} f\left(\mathfrak{X}^{k+1}_{[< n]},\mathbf{v}^{k}_{n},\mathfrak{X}^{k}_{[> n]}\right),\mathbf{x}^{k+1}_{n}-\mathbf{v}^k_{n} \right\rangle \\
			\quad + \frac{1}{2\gamma_{n}^k}\|\mathbf{x}^{k+1}_{n}-\mathbf{v}^k_n\|^2\\
			\leq g_n\left(\mathbf{x}^*_n\right)+\left\langle \nabla_{n} f\left(\mathfrak{X}^{k+1}_{[< n]},\mathbf{v}^{k}_{n},\mathbf{x}^{k}_{[> n]}\right),\mathbf{x}^*_n-\mathbf{v}^k_n \right\rangle \\
			\quad + \frac{1}{2\gamma_{n}^k}\|\mathbf{x}^*_n-\mathbf{v}^k_n\|^2.
		\end{array}\label{eq:xk1-xstar}
	\end{equation}
	Let $k_j = k+1 $ in Eq.~\eqref{eq:xk1-xstar} and $j \to \infty$, by taking $\limsup$ on both sides, we have
	$
	\limsup \limits_{j\to\infty}g_n\left(\mathbf{x}^{k_j}_{n}\right)\leq g_n\left(\mathbf{x}_n^*\right)
	$.
	On the other hand, since $g_n$ is lower semi-continuous and $\mathbf{x}^{k_j}_{n} \to \mathbf{x}^{*}_{n}$, 
	it follows that
	$\liminf\limits_{j\to\infty}g_n\left(\mathbf{x}^{k_j}_{n}\right)\geq g_n\left(\mathbf{x}^*_n\right)$.
	Thus we have
	$\lim\limits_{j\to\infty}g_n\left(\mathbf{x}^{k_j}_{n}\right)= g_n\left(\mathbf{x}^*_n\right)$.
	Note that the continuity of $f$ yields $\lim\limits_{j\to\infty}f\left(\mathfrak{X}^{k_j}\right)= f\left(\mathfrak{X}^*\right)$. Therefore, we conclude
	\begin{equation}
		\begin{array}{l}
			\quad \lim \limits_{j\to\infty} f\left(\mathfrak{X}^{k_j}\right) + \sum \limits_{n=1}^{N} g_{n}\left( \mathbf{x}^{k_j}_{n}\right) = f\left(\mathfrak{X}^{*}\right) + \sum \limits_{n=1}^{N} g_{n}\left( \mathbf{x}^{*}_{n}\right) \\
			\Rightarrow
			\lim\limits_{j\to\infty}\Psi\left(\mathfrak{X}^{k_j}\right)= \Psi\left(\mathfrak{X}^*\right). \end{array} \label{eq:multi-psi-lim-xstar}
	\end{equation}
	Considering $\lim\limits_{k\to\infty}\Psi\left(\mathfrak{X}^{k+1}\right) = \Psi^{*}$ in Eq.~\eqref{eq:multi-psi-accu}, we have 
	$\lim\limits_{j\to\infty}\Psi\left(\mathfrak{X}^{k_j}\right) = \Psi^{*}$, and thus
	\begin{equation}
		\Psi\left(\mathfrak{X}^{*}\right)= \Psi^{*}. \label{eq:multi-psi-lim}
	\end{equation}
	From Eqs.~\eqref{eq:multi-inequ-x-v} and ~\eqref{eq:multi-c}, we  conclude that


	\begin{equation}
		\begin{array}{l}
			\quad \min\limits_{k,n} \left\{ \frac{\mu^k_n}{2}-C^{k}_{n}, \frac{1}{2\gamma_{n}^k}-\frac{\lambda_n^{k}}{2} \right\}
			\sum\limits_{k=0}^{\infty}\| \mathfrak{X}^{k+1} - \mathfrak{X}^{k}\|^{2} \\
			\leq \min\limits_{k,n} \left\{ \frac{\mu^k_n}{2}-C^{k}_{n}, \frac{1}{2\gamma_n^k}-\frac{\lambda_n^{k}}{2} \right\}
			\sum\limits_{k=0}^{\infty}\sum\limits_{n=1}^{N} \| \mathbf{x}^{k+1}_{n} - \mathbf{x}^{k}_{n}\|^2 \\
			\leq \sum\limits_{k=0}^{\infty} \sum\limits_{n=1}^{N} \left(\frac{1}{2\gamma_n^k}-\frac{L_{n}\left(\mathfrak{X}^{k+1}_{[< n]}, \mathfrak{X}^{k}_{[> n]} \right)}{2}\right) \| \mathbf{x}^{k+1}_{n} - \mathbf{v}^{k}_{n}\|^2 \\
			\qquad + \left(\frac{\mu^k_n}{2}-C^{k}_{n}\right) \| \mathbf{v}^{k}_{n} - \mathbf{x}^{k}_{n}\|^2 \\
			\leq \sum\limits_{k=0}^{\infty} \sum\limits_{n=1}^{N}
			\Psi\left(\mathfrak{X}^{k+1}_{[< n]},\mathbf{v}^{k}_{n},\mathfrak{X}^{k}_{[> n]}\right) - \Psi\left(\mathfrak{X}^{k+1}_{[\leq n]},\mathfrak{X}^{k}_{[> n]}\right) \\
			\qquad
			+ \Psi\left(\mathfrak{X}^{k+1}_{[\leq n-1]},\mathfrak{X}^{k}_{[>n-1]}\right)  -
			\Psi\left(\mathfrak{X}^{k+1}_{[< n]},\mathbf{v}^{k}_{n},\mathfrak{X}^{k}_{[> n]}\right) \\
			\leq \sum\limits_{k=0}^{\infty}
			\Psi\left( \mathfrak{X}^{k}\right) - \Psi\left( \mathfrak{X}^{k+1}\right)
			\leq \Psi\left( \mathfrak{X}^{0}\right) - \Psi\left( \mathfrak{X}^{*}\right) < \infty,
		\end{array}
	\end{equation}
	which implies $\| \mathfrak{X}^{k+1} - \mathfrak{X}^{k}\| \to 0$ when $k \to \infty$.	
	By considering the first-order optimality condition of Eq.~\eqref{eq:multi-mc-min-x} and setting $k_j = k+1$, we have 
	
	\begin{equation}
		\begin{array}{l}
			\quad \mathbf{0} \in\partial_n g_n(\mathbf{x}^{k_j}_{n})   + \nabla_{n} f(\mathfrak{X}^{k_j}_{[< n]},\mathbf{v}_{n}^{k},\mathfrak{X}^{k}_{[> n]} )
			 + \frac{1}{\gamma_n^k}(\mathbf{x}^{k_j}_{n}-\mathbf{v}^{k}_{n}) \\
			\Leftrightarrow \nabla_{n} f\left(\mathfrak{X}^{k_j} \right) - \nabla_{n} f\left(\mathfrak{X}^{k_j}_{[< n]},\mathbf{v}_{n}^{k},\mathfrak{X}^{k}_{[> n]} \right) - \frac{1}{\gamma_n^k}\left(\mathbf{x}^{k_j}_{n}-\mathbf{v}^k_{n}\right) \\ \qquad \in\partial_n \Psi\left(\mathfrak{X}^{k_j}\right)\\
			\Rightarrow \left\|\nabla_{n} f\left(\mathfrak{X}^{k_j} \right) - \nabla_{n} f\left(\mathfrak{X}^{k_{j}}_{[< n]},\mathbf{v}_{n}^{k},\mathfrak{X}^{k}_{[> n]} \right) - \frac{1}{\gamma_n^k}\left(\mathbf{x}^{k_j}_{n}-\mathbf{v}^k_{n}\right)\right\| \\	
			\quad \leq M\left\|\mathfrak{X}^{k_{j}} - \left( \mathfrak{X}^{k_{j}}_{[< n]},\mathbf{v}_{n}^{k},\mathfrak{X}^{k}_{[> n]} \right) \right\| +  \frac{1}{\gamma_n^k}\|\mathbf{x}^{k_j}_{n}-\mathbf{v}^k_{n}\| \\
			\quad \leq M(\sum\limits_{i=n+1}^{N}\| \mathbf{x}_{i}^{k_j} - \mathbf{x}_{i}^{k}\|  + \| \mathbf{x}^{k_j}_{n}-\mathbf{v}^k_{n}\|) + \frac{1}{\gamma^{-}}\|\mathbf{x}^{k_j}_{n}-\mathbf{v}^k_{n}\| \\
			\quad \leq M\|\mathfrak{X}^{k_j}-\mathfrak{X}^k\| + ( M + \frac{1}{\gamma^{-}} )\|\mathbf{x}^{k_j}_{n}-\mathbf{v}^k_{n}\| \to 0
		\end{array}\label{eq:multi-gradient}
	\end{equation}
	when $j\to\infty$.  
	Here $\partial_n$ denotes the partial sub-differential with respect to $\mathbf{x}_n$, $\gamma^{-} = \inf\{\gamma_{n}^{k}: k \in \mathbb{N} , n = 1,\dots,N\}$ and $M$ is defined in Eq.~\eqref{eq:lip-joint}.
	Therefore, combing Eqs.~\eqref{eq:multi-psi-accu},~\eqref{eq:multi-psi-lim} and \eqref{eq:multi-gradient} with the definition of sub-differential, we finally concludes
	\begin{equation}
	\begin{array}{l}
	\quad \|\partial \Psi\left(\mathfrak{X}^{k_j}\right)\|
	\leq \sum\limits_{n=1}^{N} \|\partial_n \Psi\left(\mathfrak{X}^{k_j}\right) \|\\
	\leq NM\|\mathfrak{X}^{k_j}-\mathfrak{X}^k\| + (M+ \frac{1}{\gamma^{-}}) \sum\limits_{n=1}^{N}\|\mathbf{x}^{k_j}_{n}-\mathbf{v}^k_{n}\|
	\to 0
	\end{array}
	\label{eq:multi-partical}
	\end{equation}
	when $j\to\infty$.
	i.e.,
	\begin{equation}
		0 \in \partial \Psi\left(\mathfrak{X}^{*}\right), \ \forall \ \mathfrak{X}^{*} \in \Omega,
	\end{equation}
	where $\Omega$ denotes the set including all accumulation points of $\left\{ \mathfrak{X}^{k}\right\}_{k \in \mathbb{N}}
	$. Therefore, we have that all accumulation points $\mathfrak{X}^{*}$ are the critical points of $\Psi$.
	
	Finally, based on the K{\L} property of $\Psi\left( \mathfrak{X}\right)$ and using similar derivations as that in the proof of \textbf{Theorem~\ref{thm:iFIMA}}, we also have
	\begin{equation}
		\sum\limits_{k=0}^{\infty}\| \mathfrak{X}^{k+1} - \mathfrak{X}^{k}\| < \infty. \label{eq:multi-ineq-x}
	\end{equation}
	It is clear that Eq.~\eqref{eq:multi-ineq-x} implies that the sequence $\{\mathfrak{X}^{k}\}_{k \in \mathbb{N}}$ is a Cauchy sequence, thus is globally converged to the critical points of $\Psi(\mathfrak{X})$ in Eq.~\eqref{eq:psi-m}. 
	
	Considering $\Psi(\mathfrak{X}^{k})$ is semi-algebraic and choosing $\phi(s)=\frac{t}{\theta}s^{\theta}$ as the desingularizing function, it is also easy to conclude that mFIMA still shares the same convergence rates stated in  \textbf{Corollary~\ref{cor:eFIMA}}.	
\end{proof}

\ifCLASSOPTIONcompsoc
  \section*{Acknowledgments}
\else
  \section*{Acknowledgment}
\fi

This work is partially supported by the National Natural Science Foundation of China (Nos. 61672125, 61733002, 61572096, 61432003 and 61632019), and the Fundamental Research Funds for the Central Universities.

\ifCLASSOPTIONcaptionsoff
  \newpage
\fi



%
%
\bibliographystyle{IEEEtran}
\bibliography{egbib}

%
%

%

%






\end{document}